\documentclass[iicol,pdflatex,sn-mathphys-ay]{sn-jnl}% Math and Physical %%%% Standard Packages

\usepackage{graphicx}%
\usepackage{multirow}%
\usepackage{amsmath,amssymb,amsfonts}%
\usepackage{amsthm}%
\usepackage{mathrsfs}%
\usepackage[title]{appendix}%
\usepackage{xcolor}%
\usepackage{textcomp}%
\usepackage{manyfoot}%
\usepackage{booktabs}%
\usepackage{algorithm}%
\usepackage{algorithmicx}%
\usepackage{algpseudocode}%
\usepackage{listings}%
\usepackage{comment}
\usepackage{fontawesome5}
\usepackage{graphicx}
\usepackage{booktabs}
\usepackage{multirow}
\usepackage{stmaryrd}
\usepackage{pgfplots}
\usepackage{xfakebold}
\usepackage{colortbl}
\usepackage{tikz}
\usepackage{makecell}
\usepackage{eso-pic}

\theoremstyle{thmstyleone}%
\newcommand{\mstd}[2]{$\text{#1}_{\pm#2}$}
\newcommand*{\yes}{\checkmark}
\newcommand*{\no}{\textcolor{gray}{--}}
\newcommand{\pqs}{$\text{PQ}_{16}$}
\newcommand{\pqn}{$\text{PQ}_{19}$}
\newcommand{\ours}{MC-PanDA\texttt{++}}

\newcommand*{\blarrow}{\rotatebox[origin=c]{270}{$\Rsh$}}

\definecolor{Improved}{HTML}{2CA02C}  
\definecolor{Degraded}{HTML}{D62728}  

\definecolor{google-orange}{HTML}{E37400}
\definecolor{google-blue}{HTML}{4285F4}

\definecolor{myroad}{HTML}{804080}

\definecolor{csroad}         {HTML}{804080} % road           (128,  64, 128)
\definecolor{cssidewalk}     {HTML}{F423E8} % sidewalk       (244,  35, 232)
\definecolor{csbuilding}     {HTML}{464646} % building       ( 70,  70,  70)
\definecolor{cswall}         {HTML}{66669C} % wall           (102, 102, 156)
\definecolor{csfence}        {HTML}{BE9999} % fence          (190, 153, 153)
\definecolor{cspole}         {HTML}{999999} % pole           (153, 153, 153)
\definecolor{cstrafficlight} {HTML}{FAAA1E} % traffic light  (250, 170,  30)
\definecolor{cstrafficsign}  {HTML}{DCDC00} % traffic sign   (220, 220,   0)
\definecolor{csvegetation}   {HTML}{6B8E23} % vegetation     (107, 142,  35)
\definecolor{csterrain}      {HTML}{98FB98} % terrain        (152, 251, 152)
\definecolor{cssky}          {HTML}{4682B4} % sky            ( 70, 130, 180)
\definecolor{csperson}       {HTML}{DC143C} % person         (220,  20,  60)
\definecolor{csrider}        {HTML}{FF0000} % rider          (255,   0,   0)
\definecolor{cscar}          {HTML}{00008E} % car            (  0,   0, 142)
\definecolor{cstruck}        {HTML}{000046} % truck          (  0,   0,  70)
\definecolor{csbus}          {HTML}{003C64} % bus            (  0,  60, 100)
\definecolor{cstrain}        {HTML}{005064} % train          (  0,  80, 100)
\definecolor{csmotorcycle}   {HTML}{0000E6} % motorcycle     (  0,   0, 230)
\definecolor{csbicycle}      {HTML}{770B20} % bicycle        (119,  11,  32)
\definecolor{csterrainhard}  {HTML}{46E446}
\definecolor{csunlabeled}    {HTML}{000000}

\newcommand{\cslegendbox}[2]{%
  \tikz[baseline=(box.center)]{
    \node[inner sep=0pt, minimum width=0.25cm, minimum height=0.25cm,
          fill=#1, draw=none] (box) {};
    \node[anchor=west, font=\footnotesize\ttfamily]
         at ([xshift=-1pt]box.east) {#2};
  }%
}

\newcommand{\cityscapeslegendsynth}{%
  \scriptsize
  \begin{tabular}{l@{\,}l@{\,}l@{\,}l@{\,}l}
    \cslegendbox{csfence}{fence} &
    \cslegendbox{cswall}{wall} &
    \cslegendbox{cscar}{car} &
    \cslegendbox{csbuilding}{building} &
    \cslegendbox{cstrafficsign}{traffic sign} \\
  \end{tabular}%
}

\newcommand{\cityscapeslegendfailure}{%
  \scriptsize
  \begin{tabular}{l@{\,}l@{\,}l@{\,}l@{\,}l}
    \cslegendbox{csroad}{road} &
    \cslegendbox{cssidewalk}{sidewalk} &
    \cslegendbox{cscar}{car} &
    \cslegendbox{csbicycle}{bicycle} &
    \cslegendbox{cstrain}{train} \\
  \end{tabular}%
}

\newcommand{\cityscapeslegendfailureapp}{%
  \scriptsize
  \begin{tabular}{l@{\,}l@{\,}l@{\,}l@{\,}l}
    \cslegendbox{csroad}{road} &
    \cslegendbox{cssidewalk}{sidewalk} &
    \cslegendbox{cscar}{car} &
    \cslegendbox{csperson}{person} &
    \cslegendbox{csterrain}{terrain} \\
  \end{tabular}%
}

\newcommand{\cityscapeslegendmislabeled}{%
  \scriptsize
  \begin{tabular}{l@{\,}l@{\,}l@{\,}l@{\,}l}
    \cslegendbox{csroad}{road} &
    \cslegendbox{cssky}{sky} &
    \cslegendbox{csbuilding}{building} &
    \cslegendbox{csvegetation}{vegetation} &
    \cslegendbox{cstrafficsign}{traffic sign} \\
  \end{tabular}%
}

\newcommand{\cityscapeslegendall}{%
  \scriptsize
  \begin{tabular}{l@{\,}l@{\,}l@{\,}l@{\,}l@{\,}l@{\,}l@{\,}l@{\,}l@{\,}l}
    \cslegendbox{csroad}{road} &
    \cslegendbox{cssidewalk}{sidewalk} &
    \cslegendbox{csbuilding}{building} &
    \cslegendbox{cswall}{wall} &
    \cslegendbox{csfence}{fence} &
    \cslegendbox{cspole}{pole} &
    \cslegendbox{cstrafficlight}{traffic light} &
    \cslegendbox{cstrafficsign}{traffic sign} &
    \cslegendbox{csvegetation}{vegetation} &
    \cslegendbox{csterrain}{terrain} \\ [-2pt]
    \cslegendbox{cssky}{sky} &
    \cslegendbox{csperson}{person} &
    \cslegendbox{csrider}{rider} &
    \cslegendbox{cscar}{car} &
    \cslegendbox{cstruck}{truck} &
    \cslegendbox{csbus}{bus} &
    \cslegendbox{cstrain}{train} &
    \cslegendbox{csmotorcycle}{motorcycle} &
    \cslegendbox{csbicycle}{bicycle} &
    \cslegendbox{csunlabeled}{unlabeled} \\
  \end{tabular}%
}

\theoremstyle{thmstyletwo}%

\theoremstyle{thmstylethree}%

\definecolor{changesframe}{RGB}{0,0,0}

\AddToShipoutPictureFG*{%
  \AtPageLowerLeft{%
    \hspace*{\dimexpr 1in+\hoffset+\oddsidemargin\relax}%
    \raisebox{8mm}{\small Preprint.}%
  }%
}

\begin{document}

\title[Article Title]{\centering{\ours{}: Simpler, Stronger, and More Robust \\ Domain-Adaptive Panoptic Segmentation}}

\author*[1]{\fnm{Ivan} \sur{Martinović}}\email{ivan.martinovic@fer.hr}
\author[1,2]{\fnm{Josip} \sur{Šarić}}\email{josip.saric@fri.uni-lj.si}
%\author[1]{\fnm{Jan} \sur{Šnajder}}\email{jan.snajder@fer.hr}
\author[3]{\fnm{Yuki M.} \sur{Asano}}\email{yuki.asano@utn.de}
\author[1]{\fnm{Siniša} \sur{Šegvić}}\email{sinisa.segvic@fer.hr}

\affil[1]{\orgdiv{Faculty of Electrical Engineering and Computing}, \orgname{University of Zagreb}, \orgaddress{\street{Unska 3}, \postcode{10000} \city{Zagreb}, \country{Croatia}}}
\affil[2]{\orgdiv{Faculty of Computer and Information Science}, \orgname{University of Ljubljana}, \orgaddress{\street{Večna Pot 113}, \postcode{1000} \city{Ljubljana}, \country{Slovenia}}}
\affil[3]{\orgdiv{Fundamental AI Lab}, \orgname{University of Technology Nuremberg}, \orgaddress{\street{Dr.-Luise-Herzberg-Straße}, \postcode{90461} \city{Nürnberg}, \country{Germany}}}

\abstract{
Unsupervised domain adaptation (UDA) reduces the annotation burden in panoptic segmentation by leveraging a cost-effectively labeled source domain (e.g., synthetic) and an unlabeled target domain to bridge the distribution gap. Existing panoptic UDA methods rely on teacher-student consistency learning built upon suboptimal per-pixel segmentation architectures. In contrast, state-of-the-art mask transformers are rarely adopted due to their pronounced vulnerability to confirmation bias in consistency learning, where erroneous teacher predictions are reinforced during training. Our earlier approach, MC-PanDA~\citep{martinovic2024eccv}, mitigates this issue through fine-grained confidence estimation, which suppresses gradients from unreliable masks while sampling informative yet reliable locations for loss computation. 
However, this method entails a complex multi-stage training and requires careful hyperparameter tuning.
This work presents \ours{}, which addresses these limitations by introducing: (i) self-supervised vision encoders that provide a stronger and more robust initialization, further reducing the reliance on human annotations, (ii) per-class, self-adapting mask-wide loss scaling that stabilizes training and enables the usage of a single set of hyperparameters across domains, and (iii) a single-stage training pipeline that decreases overall conceptual complexity. Together, these improvements result in a conceptually simpler, better-performing, and more robust method for domain-adaptive panoptics. Source code: \href{https://github.com/martinovicivan/MC-PanDA}{github.com/martinovicivan/MC-PanDA}.
}
\maketitle

\section{Introduction} \label{sec:introduction}

Panoptic segmentation provides 
comprehensive scene understanding 
by assigning each pixel 
both a semantic category 
and an instance identity~\citep{kirillov2019panoptic}.
Recent advances have led to 
consistent performance improvements, 
bringing a wide range of practical applications 
within reach~\citep{cheng2022masked,yu2022k,li2023mask}.
\begin{figure*}[t]
    \centering
  \includegraphics[width=\linewidth]{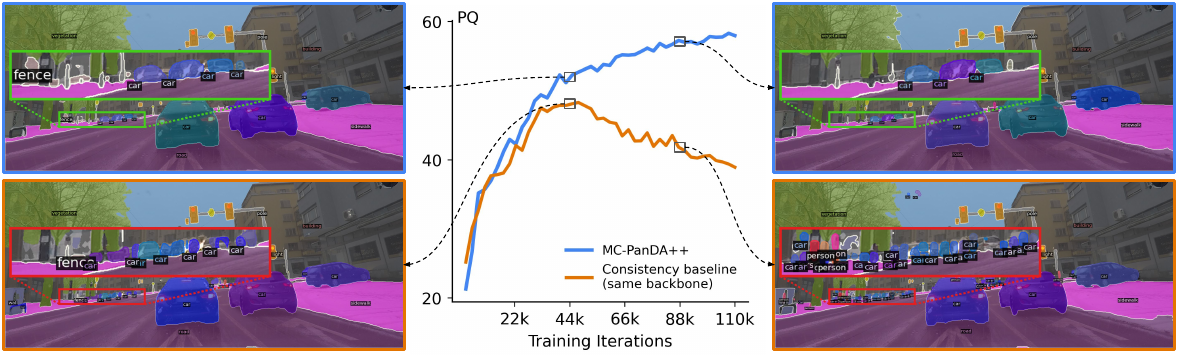}
  \caption{Validation Panoptic Quality (PQ) during domain adaptation from Cityscapes to ACDC. Baseline consistency training collapses due to confirmation bias, producing hallucinated false-positive masks, whereas our \ours{} mitigates this issue through curated consistency learning based on mask confidence estimation.}
  \label{fig:fig1}
\end{figure*}
However, many important domains
require application-specific datasets
that are very expensive to annotate~\citep{zlateski18cvpr,zendel19cvprw}, costing up to 5 hours per image for human experts~\citep{brodermann2024muses}.
Furthermore, even standard domains
such as driving scenes
suffer from the long tail of corner cases
that are not easily collected~\citep{zendel18eccv,sakaridis21iccv,uijlings22eccv,zendel22cvpr}, e.g., animals on the road at night.

Synthetic datasets offer 
a cost-effective and scalable alternative, 
allowing controlled simulation 
of corner cases 
and yielding extensive labeled training data.
Yet, the benefits of synthetic data 
can only be fully realized 
if the domain gap between 
synthetic and real images
is properly addressed~\citep{Ros_2016_CVPR,richter16eccv,kim20cvpr}.
This challenge has established unsupervised domain adaptation (UDA) as a critical research direction~\citep{bendavid10ml,ganin15icml,sun16eccvw}, as it aims to reduce the domain gap 
using only unlabeled images 
from the target domain.
Currently, the dominant UDA paradigm is
teacher–student consistency learning~\citep{meanteacher}, effective across numerous downstream tasks~\citep{hoyer23cvpr,li2022cross}, including panoptic segmentation~\citep{mansour2025wacv}.
In particular, significant progress has been achieved through various extensions of this paradigm, including auxiliary consistency terms~\citep{huang2021cross,mansour2025wacv}, loss modulation~\citep{Saha_2023_ICCV}, and teacher prediction calibration~\citep{zhang23cvpr}.

However, 
the performance of panoptic UDA methods 
remains limited by their reliance on suboptimal 
per-pixel segmentation architectures 
with separate \textit{things} and \textit{stuff} decoders,
requiring heuristic post-processing.
In contrast, supervised panoptic segmentation 
is dominated by mask transformers~\citep{cheng2022masked,jain2023oneformer,wang2023internimage}, 
which provide a unified solution 
for both semantic and instance prediction.
The limited adoption of mask transformers in UDA context 
stems from difficult 
and unstable consistency training
as observed by~\citet{zhang23cvpr}
and our baseline experiments.
We believe that the high capacity of mask transformers makes them particularly susceptible to confirmation bias, 
which commonly arises in self-training frameworks 
like consistency learning~\citep{arazo2020pseudo}.
Therefore, a novel training strategy 
is essential to unlock the potential 
of mask transformers in UDA.

To address this challenge, we propose \ours{}, 
which uses \underline{M}ask \underline{C}onfidence to guide consistency training in 
\underline{Pan}optic \underline{D}omain \underline{A}daptation.
In particular, our method associates each 
teacher-predicted mask with a mask-level confidence measure.
Additionally, for every mask, we derive 
the dense sampling affinity map
by blending fine-grained teacher  
confidence with student uncertainty. 
Together, these mask-wide confidences and sampling affinities allow us
to discourage self-learning in uncertain masks
and at locations with inappropriate pixel-level uncertainty.
This mechanism is crucial for mitigating confirmation bias,
as illustrated in Figure~\ref{fig:fig1}.
We observe that the validation panoptic quality (PQ) of a standard consistency baseline collapses mid-training, whereas our method maintains a stable learning trajectory.
The collapse arises from naive self-training amplifying noise, evident in the baseline predictions (Fig.~\ref{fig:fig1}, \textcolor{google-orange}{orange}) where the number of hallucinated false-positive masks increases drastically over the course of training. On the other hand, our method (Fig.~\ref{fig:fig1}, \textcolor{google-blue}{blue}) successfully mitigates this issue through guided self-training based on mask confidence estimation.

In summary, we contribute
two novel techniques
for domain adaptive panoptic segmentation:
mask-wide loss scaling (MLS), 
which utilizes aggregated region-wide confidence, 
and confidence-based point filtering (CBPF),
which prioritizes learning
in points with
confident teacher 
and uncertain 
student predictions.
Experiments on standard benchmarks
reveal substantial improvements 
in the generalization performance.
Our method outperforms 
the state of the art
by a large margin, advancing the viability of synthetic data for real-world applications.

\ours{} is an extension of our earlier ECCV'24 work, MC-PanDA \citep{martinovic2024eccv}, introducing several key
contributions:  
\textbf{(1)} We analyze the robustness of strong self-supervised vision encoders under
domain shift and demonstrate their suitability for domain-adaptive panoptic
segmentation.  
\textbf{(2)} We significantly streamline the preliminary three-stage MC-PanDA training
pipeline into a single-stage design by leveraging robust self-supervised initialization.  
This simplification reduces the number of hyperparameters and the overall
conceptual complexity.  
\textbf{(3)} We improve mask-wide loss scaling (MLS) by introducing class-dependent and
self-adapting confidence thresholding, enabling the use of the same
initialization-independent hyperparameters across all experiments.  
This substantially reduces variance and increases robustness, addressing a key
limitation of MC-PanDA.  
\textbf{(4)} We provide a notably extended experimental study, detailing the evolution from MC-PanDA to \ours{}. Specifically, we present: 
i) a detailed ablation study showing that our contributions remain
effective even with stronger vision encoders,  
ii) results on two new synthetic-to-real and two new
clear-to-adverse benchmarks, accompanied by corresponding ablations,  
iii) analysis of key design choices of our new class-dependent adaptive MLS, demonstrating its effectiveness in increasing stability and reducing sensitivity to the initial hyperparameter value,
iv) robustness analysis of CBPF, and
v) impact of scaling the vision 
encoder.

\section{Related work}                  \label{sec:relatedwork}
\bmhead{Panoptic segmentation}
While many real-world applications require both semantic and instance-level
segmentation, early research treated these tasks separately. The introduction of
panoptic segmentation as a unified formulation~\citep{kirillov2019panoptic} sparked
considerable interest in jointly modeling \textit{things} and \textit{stuff} classes.
Initial approaches extended Mask R-CNN~\citep{he2017mask} with an additional semantic
segmentation branch and explored various fusion strategies for combining the two
prediction streams~\citep{kirillov2019panoptic,li2018tascnet,xiong2019upsnet,kirillov2019pfpn}. 
Panoptic-DeepLab~\citep{cheng2020panoptic} advanced this direction by building on
a semantic segmentation architecture and introducing class-agnostic instance
predictions via center and offset regression.

\bmhead{Unified panoptic segmentation with Mask Transformers}
Recent work has introduced unified transformer-based architectures that represent
both \textit{things} and \textit{stuff} regions using dense sigmoidal maps called \emph{masks}~\citep{yu2022k,cheng2022masked,li2023mask}.
These masks are recovered by scoring dense features 
with the associated embeddings~\citep{cheng2021per}.
Mask embeddings are obtained 
through direct set prediction
with an appropriate transformer module~\citep{carion2020end}.
Beyond simplifying inference,
these models 
currently achieve the
state of the art on
panoptic segmentation benchmarks~\citep{zhou2017scene,lin2014microsoft}.
Moreover, recent studies highlight the strong ability of mask transformers to
estimate their own prediction uncertainty~\citep{grcic23cvprw}, utilized by several leading methods in dense anomaly detection
~\citep{rai23iccv,nayal23iccv,ackermann23bmvc,delic24bmvc}. 

\bmhead{Self-supervised representation learning} 
Self-supervised learning (SSL) is a pre-training paradigm in which models optimize a pretext task on unlabeled data. Since such data can be collected at scale and without manual labeling, self-supervised models offer robust and generalizable representations while avoiding the cost, bias, and ambiguity of human annotations. A major line of SSL research includes clustering-based methods such as DeepCluster~\citep{caron2018deep}, SeLa~\citep{asano2020self}, SwAV~\citep{caron2020unsupervised}, and DINO~\citep{caron2021emerging}. Extending DINO with masked image modeling~\citep{zhou2022image} and large-scale curated pre-training yields DINOv2~\citep{oquab2024dinov}, proven to be effective across many downstream tasks~\citep{tolan2024very,xu2024whole}. Beyond that, self-supervised foundation models showed great robustness to domain shift~\citep{hummer2024strong,wei2024stronger,yun2025soma}.
Hence, we initialize our backbone with self-supervised DINOv2, which stabilizes training and enables us to streamline the pipeline from three stages to a single stage.
\textcolor{changesframe}{
Beyond standard transformers, recent state-space models such as
Mamba~\citep{gu2024mamba} have been adapted to vision tasks~\citep{vim,jiang2025ph}.
Related progress in self-supervised and multimodal visual learning includes remote-sensing
super-resolution~\citep{superresolution} and multimodal feature fusion~\citep{aigcvideodetection},
which are complementary to our focus on domain-adaptive panoptic segmentation.
}

\bmhead{Unsupervised domain adaptation for segmentation}
Unsupervised domain adaptation (UDA) considers a labeled source domain 
$\mathcal{D}_\mathrm{src}$ and an unlabeled target domain $\mathcal{D}_\mathrm{tgt}$, assuming access to target-domain images during training.
Since the 
two
domains typically exhibit a substantial distribution
shift, models trained solely on $\mathcal{D}_{src}$ often generalize poorly to
$\mathcal{D}_{tgt}$~\citep{zendel18eccv,sakaridis21iccv}.  
UDA addresses this by optimizing a joint objective over labeled source data and
unlabeled target data~\citep{bendavid10ml,sun16eccvw}:
\begin{align}
    \mathcal{L}_\mathrm{uda} &= \mathcal{L}_\mathrm{src} + \mathcal{L}_\mathrm{tgt}. 
    \label{eq:uda}
\end{align}
The key challenge is therefore how to construct the target-domain loss
$\mathcal{L}_\mathrm{tgt}$.  
Recent approaches rely heavily on consistency learning and the Mean
Teacher framework~\citep{meanteacher}, which has shown strong performance
in segmentation UDA~\citep{hoyer23cvpr,Saha_2023_ICCV}.  
A number of methods further incorporate pixel-level uncertainty to filter or refine
pseudo-labels~\citep{zheng2021rectifying,zhou2022uncertainty}.  
In contrast, our method aggregates \emph{region-wide} uncertainties, leverages both
teacher and student uncertainty, and employs sparse point sampling
~\citep{kirillov2020pointrend} rather than per-pixel loss scaling
~\citep{zheng2021rectifying,zhou2022uncertainty}.

\bmhead{Domain adaptive panoptic segmentation}
Domain-adaptive panoptic segmentation remains far less explored than semantic
segmentation, with only a few notable methods.  
EDAPS~\citep{Saha_2023_ICCV} extends consistency learning with image-wide scaling of
the self-supervised loss~\citep{hoyer2022daformer,tranheden2021dacs,hoyer2022hrda},
which aims to stabilize training by reducing gradient magnitudes in images where
the teacher is uncertain.  
However, this strategy can be suboptimal because it uniformly down-weights all
pixels, even when uncertainty varies spatially. This issue becomes particularly
pronounced when using mixing-based augmentations that paste source content into
target images~\citep{french20bmvc,tranheden2021dacs}.  
LIDAPS~\citep{mansour2025wacv} builds upon EDAPS by additionally employing instance
mixing from target to source and incorporating CLIP textual embeddings to regularize
the semantic branch.

Our work is most closely related to UniDAformer~\citep{zhang23cvpr}, the
only prior method exploring domain adaptation with mask transformers.  
However, UniDAformer relies on handcrafted refinement of teacher predictions and reports its best results using a multi-branch per-pixel architecture, while its mask-transformer variants underperform (\textit{cf.}~Table~\ref{tab:synthetic_to_real_mean}
and~\citep{zhang23cvpr}).
In contrast, we introduce the first panoptic adaptation approach based on region-wide,
fine-grained uncertainty estimation.  
The design is motivated by recent success of mask transformers in dense anomaly
detection~\citep{rai23iccv,grcic23cvprw,nayal23iccv}. Our preliminary work, MC-PanDA~\citep{martinovic2024eccv}, confirmed this potential, 
showing that mask transformers paired with fine-grained uncertainty estimation can 
set new state-of-the-art performance in panoptic domain adaptation.

\section{MC-PanDA\texttt{++}} \label{sec:methodology}
\begin{figure*}[h!]
    \centering
    \includegraphics[width=\linewidth,trim=22 114 20 83, clip]{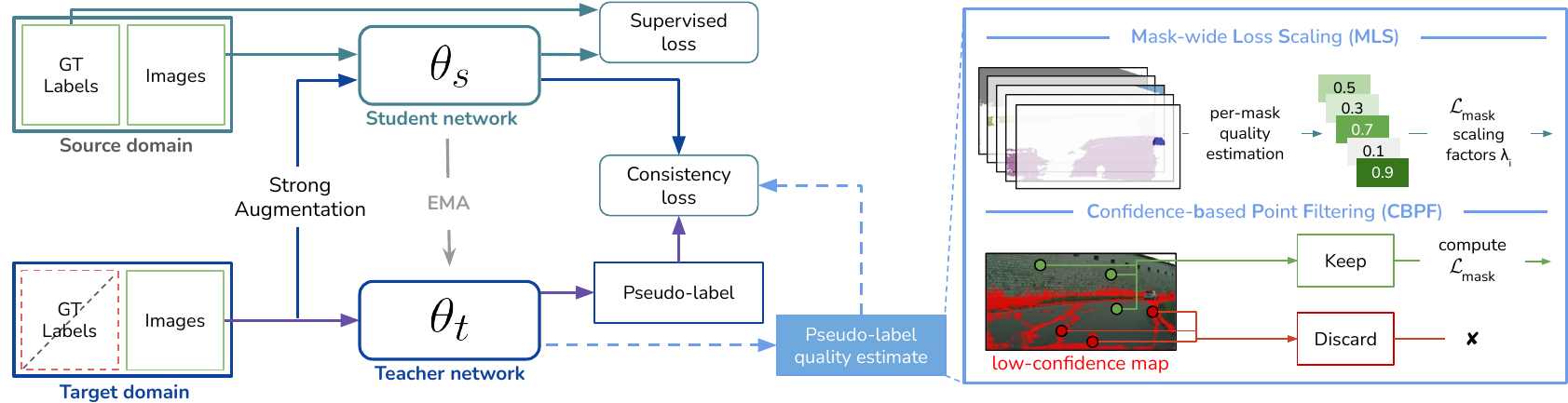}
    \caption{
    Our UDA baseline jointly optimizes the student with a supervised source-domain loss and a target-domain teacher–student consistency loss, thereby exposing it to confirmation bias. \ours{} mitigates this issue by estimating fine-grained teacher confidence, which guides mask-wide loss scaling and point sampling for loss computation.
    }
    \label{fig:mc-panda-overview}
\end{figure*}
We first recap
panoptic segmentation with mask transformers
in subsection~\ref{sec:panseg_mask}.
Then, we describe
a baseline domain adaptation 
with a mask transformer 
in~\ref{sec:uda_basics}.
Subsections~\ref{sec:mls} and~\ref{sec:cbpf}
present our contributions
based on per-mask loss modulation 
and loss subsampling
according to pixel-level confidence.
Finally, in subsection~\ref{sec:method_training}
we describe the simplified training pipeline.

\subsection{Panoptic segmentation with Mask Transformers}
\label{sec:panseg_mask}
Recent methods for scene understanding~\citep{cheng2022masked,li2023mask,yu2022k,carion2020end} 
can handle
semantic, instance or panoptic segmentation 
without changing the loss function
or the model architecture.
These models directly detect 
instances and stuff segments,
and describe them with 
distinct segmentation masks.
The dense feature extractor 
produces pixel embeddings 
$E_p \in \mathbb{R}^{H\times W \times d}$, 
where $H$ and $W$ stand for height and width.
The transformer decoder observes the features
and classifies each of the $N$ mask queries across $(C + 1)$ classes. 
The $(C+1)$-th class (no-object) indicates 
that the corresponding mask 
is unused in this particular image. 
The transformer decoder also produces 
mask embeddings $E_m \in \mathbb{R}^{N \times d}$
that identify the corresponding pixel embeddings $E_p$
through dot-product similarity. 
Combining the two embeddings 
through generalized matmul 
and sigmoid activation
produces pixel-to-mask 
assignments $\sigma \in \mathbb{R}^{N \times H \times W}$.
Thus, 
each mask
is defined with 
$\sigma_i \in \mathbb{R}^{H \times W}$
and class distribution $P_i = (p_{i}^{(1)}, p_{i}^{(2)}, ..., p_{i}^{(C+1)})$.

The training process
minimizes the difference
between the 
ground truth masks $\{(\sigma_i^\mathrm{GT}, y_i^\mathrm{GT})\}_i^{N^\mathrm{GT}}$
and the predictions $\{(\sigma_i, P_i)\}_i^N$.
The loss computation requires
bipartite matching  $\mathcal{M}$
which maps prediction mask index to 
the ground truth mask index
while minimizing the overall matching cost.
Note that the set of ground truth masks
has to be extended with empty masks 
$(\sigma_i^\mathrm{GT}=\textbf{0}, y_i^\mathrm{GT}=C + 1)$
in order to match the number of predictions.
Given $\mathcal{M}$,
we can compute the loss 
consisting of recognition
and localization terms:
\begin{align}
    \label{eq:m2f_loss}
    \mathcal{L}^\mathrm{MT} = \sum_i^N 
    \mathcal{L}_\mathrm{cls}(P_i, y^\mathrm{GT}_{\mathcal{M}(i)}) + 
    \hspace{-1em}
    \sum_{ y_{\mathcal{M}(i)}^\mathrm{GT} \neq C + 1}{
    % \llbracket y_{\mathcal{M}(i)}^{GT} \neq C + 1 \rrbracket
    \hspace{-1.5em}
    \mathcal{L}_\mathrm{mask}(\sigma_i, \sigma^\mathrm{GT}_{\mathcal{M}(i)})}
\end{align}
The recognition terms $\mathcal{L}_\mathrm{cls}$
require correct semantic classification,
while the localization terms $\mathcal{L}_\mathrm{mask}$
optimize per-pixel assignments.
Note that $\mathcal{L}_\mathrm{mask}$
is computed only for masks
matched with non-empty ground truth, and
that our source domain loss 
$\mathcal{L}_\mathrm{src}$ from~(\ref{eq:uda})
corresponds to
$\mathcal{L}^\mathrm{MT}$ from~(\ref{eq:m2f_loss}). 
During inference, 
each pixel (\underline{r}ow, \underline{c}olumn) is assigned 
the mask $M$ that maximizes 
the pixel-level confidence $\rho$ 
that is  expressed as 
a product 
of recognition and localization scores:
\begin{equation}
\label{eq:m2f_inference}
\begin{split}
M(r,c) &= \arg\max_i \rho_{i, r,c} \\
\rho_{i,r,c} &= \max_{y \neq C+1} P_i(y) \cdot \sigma_i (r, c)
\end{split}
\end{equation}

\subsection{Baseline consistency learning with Mean Teacher}
\label{sec:uda_basics}
This section presents our baseline 
panoptic domain adaptation training
of mask transformers, as illustrated
on the left panel of Figure~\ref{fig:mc-panda-overview}.
Each training iteration takes in an equal number of source and target domain images.
With ground truth available in the source domain, 
the corresponding objective $\mathcal{L}_\mathrm{src}$ follows the standard supervised mask transformer loss (Eq.~\ref{eq:m2f_loss}).
For the target domain, we employ 
Mean Teacher self-training,
since only unlabeled images are available.
We feed the teacher with clean images
and block the gradients 
through the corresponding branch.
We perturb 
the student images 
with strong augmentations consisting of
color jitter \citep{chen2020simple},
random application
of Gaussian smoothing,
and SegMix  - 
our panoptic adaptation
of ClassMix~\citep{olsson2021classmix,tranheden2021dacs}.
Instead of classes,
SegMix
samples half of the 
panoptic segments
from a random source image
and pastes them atop
the target image.
We recover the teacher predictions
and convert them 
to hard pseudo-labels
consisting of $N^\mathrm{teach}$ masks
defined with the semantic class $y_i^\mathrm{teach}$
and dense assignment map $\sigma_i^\mathrm{teach}$.
Finally, we obtain the target-domain loss
$\mathcal{L}_\mathrm{tgt}$ (Eq.~\ref{eq:uda})
as the teacher-student consistency loss, 
defined identically to the mask transformer loss $\mathcal{L}^\mathrm{MT}$ (Eq.~\ref{eq:m2f_loss}),
except that the ground truth masks
$(\sigma_i^\mathrm{GT}, y_i^\mathrm{GT})$
are replaced with the teacher pseudo-label masks
$(\sigma_i^\mathrm{teach}, y_i^\mathrm{teach})$.

We set the teacher parameters $\Theta_t$
to the temporal exponential moving average 
(EMA) of the student parameters $\Theta_s$
\citep{meanteacher}.
Nevertheless, 
our baseline students
still tend to 
deteriorate due 
to noisy pseudo-labels.
Moreover,
we notice that
our baseline teachers 
produce many 
false positive masks.
Training on such pseudo-labels 
tends to further amplify the noise
due to the Mean Teacher setup,
which introduces confirmation bias 
as the model repeatedly 
validates its own erroneous predictions.

We tackle this problem by guiding the self learning according to the estimated quality of pseudo-labels, as shown in the right panel of Figure~\ref{fig:mc-panda-overview}. Specifically, we use mask confidence to identify unreliable masks and point sampling locations for loss computation. The following subsections provide more details of our approach.

\subsection{Mask-wide loss scaling}
\label{sec:mls}
Our baseline underperforms
due to positive feedback loop 
between the student and the teacher.
We propose to alleviate this effect
by modulating the per-mask 
localization term $\mathcal{L}_\mathrm{mask}$ 
of the student loss $\mathcal{L}_\mathrm{tgt}$
with mask-wide teacher confidence $\lambda_i$. 
We scale the localization
loss as follows:
\begin{equation}
\label{eq:lambda}
\begin{split}
    \mathcal{L}_\text{loc}^{\text{MC}} &=
    \sum_{y_{\mathcal{M}(i)}^\text{teach} \neq C + 1}
    \lambda_i \cdot \mathcal{L}_{\text{mask}}
    (\sigma_i, \sigma^\text{teach}_{\mathcal{M}(i)}); \\
    \lambda_i &= 
    \frac{\sum_{(r,c) \in M^{F}_i} 
    \llbracket \rho_{i, r,c} > \tau_1 \rrbracket }
    {|M^{F}_i|}.
\end{split}
\end{equation}

Note that $M^{F}_i$ represents the foreground locations for mask i,
$\llbracket . \rrbracket$ -- Iverson indicator function, 
and $\tau_1$ a threshold. 
The pixel-level confidence $\rho$ is defined in~(\ref{eq:m2f_inference}).
The mask-wide teacher confidence $\lambda_i$ 
corresponds to the ratio 
of the foreground pixels 
where the pixel-level confidence 
is larger than the threshold. 
\textcolor{changesframe}{
Intuitively, the mask-wide coefficient $\lambda_i$ estimates the quality of a teacher
pseudo-mask. Since $\lambda_i$ is the fraction of predicted foreground pixels whose panoptic
confidence exceeds $\tau_1$, coherent and confident masks receive larger localization weights.
In contrast, incomplete or noisy masks typically contain more low-confidence foreground
regions, resulting in smaller $\lambda_i$ and a weaker contribution to the target-domain
consistency loss. Thus, MLS encourages learning from reliable pseudo-masks while downweighting
uncertain ones that could otherwise reinforce confirmation bias.
}

\subsubsection{Per-class and adaptive thresholding} \label{subsec:per_class_adaptive}

\begin{figure}[h!]
	\centering
	\includegraphics[width=\linewidth]{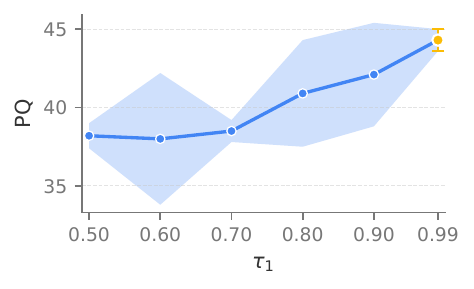}
	\caption{Panoptic performance (PQ) on Synthia$\rightarrow$Vistas as a function of the initial threshold $\tau_1$. Each point denotes the mean PQ over three random seeds, with error bars indicating the standard deviation. The orange point marks the default MC-PanDA setting with $\tau_1 = 0.99$.}
	\label{fig:instability_fixed_tau1}
\end{figure}

The previous formulation~(\ref{eq:lambda}) makes mask-wide loss scaling sensitive to the threshold $\tau_1$. A high threshold can halt learning on the target domain, while a low threshold is overly permissive, allowing noise amplification.
Fig.~\ref{fig:instability_fixed_tau1} illustrates this sensitivity by showing model performance against different values of $\tau_1$. We observe that lower $\tau_1$ values lead to substantial drops in PQ and increased variance across random seeds, compared to the default MC-PanDA setting with $\tau_1 = 0.99$. 

Furthermore, using a predetermined, fixed and uniform threshold $\tau_1$ across all classes is suboptimal for multiple reasons. First, model confidence is class-dependent.
For example, rare and visually ambiguous classes tend to exhibit lower confidence than others, even when their predictions are correct. 
Second, model confidence evolves during training, starting low and increasing over time. Third, model confidence depends on the degree of domain shift between the source and target domains. Selecting an appropriate fixed $\tau_1$ for different training and testing domains is nontrivial, making deployment across domains challenging. Ideally, the training procedure should be robust to the initial choice of the threshold $\tau_1$. However, this is difficult to guarantee if the threshold remains static throughout training.

Hence, we propose to mitigate this instability using adaptive and class-dependent thresholds that evolve during training:
\begin{equation}
\label{eq:tau_update}
\begin{split}
    \tau_{1,k}^{\,n}
    &= \alpha\,\tau_{1,k}^{\,n-1}
     + (1-\alpha)\,\delta_k^n, \\[2mm]
    \delta_{k}^n
    &= \max_{i \in \mathcal{I}_k^n}\;
       f\!\left(
            \big\{\, \rho_{i,r,c} \;\big|\; (r,c) \in M_i^{F} \,\big\}
         \right),
\end{split}
\end{equation}
where $\alpha$ is the EMA momentum controlling the smoothness of updates, 
$\mathcal{I}_k^n$ denotes the set of teacher-predicted instances belonging to class $k$ at the training iteration $n$, 
and $f\!\left(\{\rho_{i,r,c} \mid (r,c)\in M_i^{F}\}\right)$ is an instance-wise 
aggregation (e.g., \textit{mean}, \textit{median}, or N-th \textit{percentile}) of the pixel-level confidence 
over the foreground region $M_i^{F}$ of instance $i$. 
The update coefficient $\delta_k^n$ is obtained by first aggregating the panoptic score $\rho_{i,r,c}$ over the foreground pixels of each mask $M_i^{F}$ and then selecting the highest aggregated confidence among all masks of class $k$ in the batch. \textcolor{changesframe}{
If $\mathcal{I}_k^n=\emptyset$, i.e., if the teacher predicts no mask of class $k$ in the
current batch, we skip the update and keep the previous value of $\tau_{1,k}$. 
}

\textcolor{changesframe}{Formulation in Eq.~\ref{eq:tau_update}} introduces per-class thresholds $\tau_{1,k}^n$ that dynamically adapt to the evolving teacher confidence distribution, improving stability and reducing sensitivity to the
initial value. \textcolor{changesframe}{
The max-over-instances operation $\max_{i \in \mathcal{I}_k^n}$ is an empirical
choice aligned with MLS: it anchors each class threshold to the most confident teacher
masks of that class, rather than allowing many low-confidence pseudo-masks to lower the
threshold and make MLS overly permissive. Combined with the EMA update, this yields
class-specific thresholds that adapt over time while being less sensitive to noisy batch-level
fluctuations.
}

\subsection{Confidence-based point filtering}
\label{sec:cbpf}
Training dense prediction models
on high-resolution images
can be extremely memory intensive.
This problem can be alleviated by training
on a carefully chosen 
sample of $N_p$ dense predictions
instead of on 
the whole prediction tensor~\citep{kirillov2020pointrend}.
The procedure starts by sampling
a random oversized set 
of $3N_p$ floating-point locations.
The initial set
is then subsampled by choosing 
$\beta \cdot N_p$ points  
with the largest uncertainty ($\beta \in [0, 1]$),
and random $(1-\beta) \cdot N_p$ points.
However, such procedure
is inappropriate for
consistency training.
In fact, the teacher and the student
will often be uncertain
at the same locations
since they 
are presented 
with the same image 
(up to a perturbation).
Thus, blind favoring
of uncertain student points
would increase the chance
of sampling incorrect 
pseudo-labels.
On the other hand,
favouring highly confident
points
would impair the
learning process
by providing
uninformative gradients.
Thus, we propose to 
favour points with 
low student 
and high teacher confidence 
by means of
a dense sampling
affinity $A_i$:
\begin{align}
    \label{eq:unc}
    \text{A}_i\left(r,c\right) = 
    \begin{cases}
               -\infty &{\Phi}^\text{teach}_{r,c} < \tau_2,\\
       -|s_{i,r,c}| & \text{otherwise}.
    \end{cases}
\end{align}
Note that $\Phi^\text{teach}_{r,c}$ 
denotes the teacher confidence,
%that will be defined 
%before the end of this subsection,
while $s_{i,r,c}$ denotes the pre-activation 
of the student mask-assignment $\sigma$.
Thus, if the teacher confidence 
in some point (r,c)
is lower than the threshold $\tau_2$,
we prevent its sampling by setting 
the sampling affinity to $-\infty$.
When the teacher is confident, 
the sampling affinity is defined 
by the negative absolute value 
of the student pixel-to-mask pre-activation, 
allowing student-uncertain points 
to be safely prioritized.
We estimate the teacher 
confidence as follows:
\begin{align}
    \label{eq:conf_def}
    {\Phi}^\text{teach}_{r,c} = \max_i \rho^\text{teach}_{i,r,c}
\end{align}
This assigns the 
lowest confidence
to the locations
that are not 
claimed by any of the masks,
and to the locations
claimed by masks
with low classification confidence.
This formulation of point sampling
is conservative as it prevents
any mask from training on
low confidence teacher predictions.

\subsection{Simplifying the training pipeline}
~\label{sec:method_training}
\begin{figure}[h!]
	\centering
	\includegraphics[width=\linewidth,trim=200 125 200 120, clip]{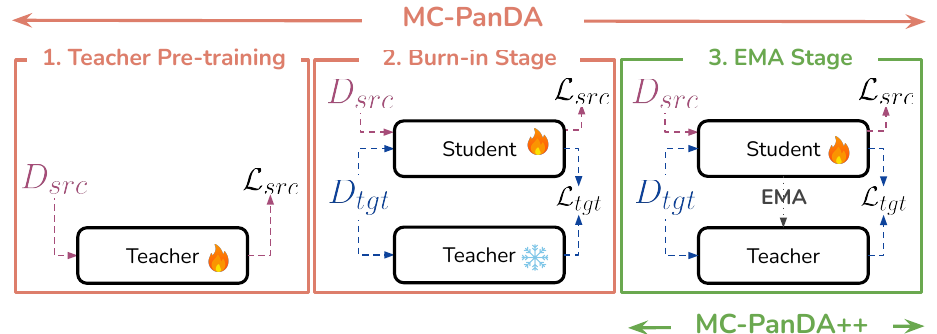}
	\caption{\ours{} simplifies the MC-PanDA training pipeline by removing source-domain pre-training and the teacher burn-in stage.}
	\label{fig:reducing_conceptual_complexity}
\end{figure}

Our preliminary method MC-PanDA relied on a three-stage training pipeline~\citep{berrada2024guided}, as illustrated in Figure~\ref{fig:reducing_conceptual_complexity}. 
The first stage initialized the backbone with supervised ImageNet weights, and pre-trained the teacher on the labeled source domain. The second stage applied consistency learning with a fixed teacher (burn-in stage)~\citep{berrada2024guided}, while the third stage employed consistency learning with a Mean Teacher~\citep{meanteacher}.
Beyond increasing conceptual complexity, the three-stage training pipeline 
introduces several stage-specific hyperparameters, such as the duration of teacher 
pre-training and the length of the burn-in phase. 
These hyperparameters vary across benchmarks,
which makes the method less robust.

Hence,  in \ours{}, we simplify this setup
to a single-stage training pipeline
based on consistency learning with Mean Teacher.
To maintain a stable learning curve similar to the three-stage setup, we introduce a simple modification.
We initialize the backbone with large-scale self-supervised weights~\citep{oquab2024dinov}, 
which provide greater robustness 
than supervised ImageNet pretraining
from MC-PanDA.
We argue this provides
sufficient stability to 
bypass the original
stabilization phases,
whose main purpose
was to mitigate early 
pseudo-label noise. 
Moreover, self-supervised backbone 
initialization further reduces the reliance on
human annotations.
In addition to the initialization, 
when using a single-stage pipeline, 
we observe that doubling the weight of the supervised $\mathcal{L}_\mathrm{mask}$ loss
yields a small performance improvement.
Overall, single-stage training in \ours{} 
eliminates both the conceptual overhead of a staged procedure 
and the need to tune burn-in hyperparameters, 
while maintaining stable adaptation performance. 

\section{Experiments}\label{sec:experiments}
We organize our experimental study as follows. 
Section~\ref{subsec:implementation_details} provides
full implementation details, followed by a comparison of \ours{} against the state of the art
on standard benchmarks in Section~\ref{subsec:sota_comparison}.
Subsequently, Section~\ref{subsec:ablating_mcpanda}
consolidates the contributions of our
preliminary method MC-PanDA
and revisits the key design choices.
This analysis 
sets the groundwork for the step-by-step transition to the simpler, stronger and more robust \ours{} detailed in Section~\ref{sec:mcpandapp_components}. 
Specifically, we:
(i) study backbone initialization (\ref{subsec:on_the_backbone_choice}) and demonstrate the effectiveness of MLS and CBPF under stronger self-supervised initialization (\ref{subsec:ablating_self-sup_init}–\ref{subsec:ablating_adverse});
(ii) simplify the original three-stage pipeline to a single-stage (\ref{subsec:streamlining});
(iii) analyze the effects of per-class adaptive mask-wide loss scaling (\ref{subsec:adaptive_mask_confidences});
and
(iv) validate the robustness of CBPF (\ref{subsec:tau2_robustness}). Finally, Section~\ref{subsec:mcpandapp} reports the complete results of \ours{} and evaluates the effect of scaling the encoder.

\subsection{Implementation details} \label{subsec:implementation_details}
\bmhead{Datasets}
We perform experiments on two standard real-world datasets:
Cityscapes~\citep{Cordts_2016_CVPR} and
Mapillary Vistas~\citep{Neuhold_2017_ICCV}.
Cityscapes comprises 2975 training and 500
validation images of European urban scenes
captured in fair weather and $1024\times2048$ resolution.
Vistas contains 18{,}000 training and 2000
validation images of worldwide scenes
under diverse weather and illumination conditions,
with a mean resolution of 8.4~MPix. 
Additionally, we evaluate \ours{} on clear-to-adverse
domain adaptation benchmarks by employing two datasets
captured in challenging conditions:
ACDC~\citep{sakaridis21iccv} and MUSES~\citep{brodermann2024muses}.
Both datasets are divided into four conditions:
\textit{fog}, \textit{nighttime}, \textit{rain}, and \textit{snow}.
The ACDC training subset comprises $1600$ images,
while the MUSES training subset includes $1500$ images.
We report performance on the corresponding validation subsets,
which consist of $406$ and $250$ images, respectively.
Following~\citep{martinovic2024eccv,Saha_2023_ICCV},
for synthetic data we consider
Synthia~\citep{Ros_2016_CVPR} with 9400 images
at $1280\times760$ resolution, and
Foggy Cityscapes~\citep{sakaridis2018semantic} with
2975 training and 500 validation images
at $1024\times2048$ resolution
and attenuation factor $0.02$.
Finally, we conduct additional experiments
on the recently introduced synthetic dataset
UrbanSyn~\citep{gomez2025urbansyn}, which comprises
7539 training images and serves
as an additional source domain. 
During inspection of UrbanSyn~\citep{gomez2025urbansyn},
we identified a subset of images
with corrupted ground-truth labels, 
such as regions assigned
to clearly incorrect classes. 
In total, we detected 57 such images
and \textcolor{changesframe}{removed them during training.}
We provide the details
of the detection procedure,
together with the full list
of affected samples in the Appendix~\ref{appendix:corrupted_urbansyn}.
\textcolor{changesframe}{
All experiments that use UrbanSyn as the source domain are conducted on this filtered split,
including the source-only baselines and all \ours{} variants. In
Appendix~\ref{appendix:corrupted_urbansyn}, we also quantify the impact of this filtering step
on the final performance.
}

\begin{table*}[t]
    \caption{\textcolor{changesframe}{Comparison with the state of the art on Synthia$\rightarrow$Cityscapes and Synthia$\rightarrow$Vistas. $\dagger$ indicates our reimplementation of EDAPS using the same Swin-B backbone as in MC-PanDA. $\ddagger$ indicates our reimplementation with the same DINOv2-B backbone as in \ours{}. All experiments for MC-PanDA and \ours{} are averaged over three random seeds.}
    }

    \label{tab:synthetic_to_real_mean}
    \centering
    \footnotesize
    \begin{tabular}{lcccccc}
        \toprule 
        & \multicolumn{3}{c}{\text{Synthia$\rightarrow$City}} & \multicolumn{3}{c}{\text{Synthia$\rightarrow$Vistas}} \\
        Method & $\textbf{PQ}_{16}$ & $\text{RQ}_{16}$ & $\text{SQ}_{16}$ & $\textbf{PQ}_{16}$ & $\text{RQ}_{16}$ & $\text{SQ}_{16}$ \\
        \midrule
        CVRN~\citep{huang2021cross}              
            & 32.1 & 40.9 & 66.6 & 21.3 & 28.1 & 65.3 \\
        UniDAF-DETR~\citep{zhang23cvpr}    
            & 33.0 & 42.2 & 64.7 & \no & \no & \no \\
        UniDAF-PSN~\citep{zhang23cvpr,kirillov2019panoptic}   
            & 34.2 & 44.3 & 66.9 & \no & \no & \no \\
        \midrule
        EDAPS~\citep{Saha_2023_ICCV}             
            & {41.2} & {53.6} & {72.7} & {36.6} & {46.1} & {71.7} \\
        EDAPS$^\dagger$~\citep{Saha_2023_ICCV}             
            & {39.3} & {51.4} & {73.1} & \no & \no & \no \\
        LIDAPS~\citep{mansour2025wacv} 
            & \text{44.8} & \text{57.6} & \text{74.4} & \text{38.0} & \text{47.7} & 73.9 \\
        MC-PanDA~\citep{martinovic2024eccv} 
            & \underline{47.4} & \underline{59.3} & \underline{76.7} & \text{38.7} & \text{49.8} & {71.0} \\
        \midrule
         % EDAPS$^\ddagger$~\citep{Saha_2023_ICCV} & \text{42.0} & \text{54.7} & \text{73.4} & \underline{40.5} & \underline{51.7} & \underline{74.7} \\
        \textcolor{changesframe}{EDAPS$^\ddagger$}~\citep{Saha_2023_ICCV} & \textcolor{changesframe}{\text{42.0}} & \textcolor{changesframe}{\text{54.7}} & \textcolor{changesframe}{\text{73.4}} & \textcolor{changesframe}{\underline{40.5}} & \textcolor{changesframe}{\underline{51.7}} & \textcolor{changesframe}{\underline{74.7}} \\
        % LIDAPS$^\ddagger$~\citep{mansour2025wacv} 
            % & {44.7} & {57.1} & {74.4} & 37.0 & 48.2 & 73.8 \\
        \textcolor{changesframe}{LIDAPS$^\ddagger$}~\citep{mansour2025wacv} & \textcolor{changesframe}{44.7} & \textcolor{changesframe}{57.1} & \textcolor{changesframe}{74.4} & \textcolor{changesframe}{37.0} & \textcolor{changesframe}{48.2} & \textcolor{changesframe}{73.8} \\
        \rowcolor{gray!10} 
        \ours{}  
            & \textbf{49.6} & \textbf{62.2} & \textbf{77.0} 
            & \textbf{44.8} & \textbf{56.6} & \textbf{76.3} \\ 
        \bottomrule
    \end{tabular}
\end{table*}

\bmhead{Architecture} 
For both MC-PanDA~\citep{martinovic2024eccv} and \ours{}, we employ the panoptic 
Mask2Former~\citep{cheng2022masked} framework with 
multi-scale deformable attention~\citep{zhu2020deformable}
in the pixel decoder.
The default backbone in MC-PanDA is Swin-B~\citep{liu2021swin}. 
In \ours{}, we adopt DINOv2-B as the primary backbone initialization
for the majority of experiments,
and DINOv2-Large to assess the effectiveness
of scaling the encoder.
We follow ViTDet~\citep{li2022exploring} to construct a ViT feature pyramid, and, following~\citet{kerssies2024benchmark}, resize the patch-embedding kernels to 
16×16 for all ViT-based experiments. \\
\bmhead{Training}
We train our models for 110k iterations
using batches composed of
2 source and 2 target-domain images. 
The teacher is updated as the exponential moving average~\citep{meanteacher}
of the student with a decay factor $\alpha = 0.999$, 
and the same $\alpha$ is used for updating the adaptive threshold $\tau_1$
(Eq.~\ref{eq:tau_update}).
In contrast to MC-PanDA, \ours{} uses a single shared set of hyperparameters 
across all experiments, further demonstrating the robustness of our method.
We use AdamW~\citep{loshchilov2018decoupled} with an initial learning rate 
0.0001 and weight decay 0.05.

\begin{table*}[h!]
	\caption{Performance evaluation on Cityscapes$\rightarrow$Foggy Cityscapes and Cityscapes$\rightarrow$Vistas. All experiments for MC-PanDA and \ours{} are averaged over three random seeds.}
	\label{tab:city_to_other_mean}
	\centering
	\footnotesize
	\begin{tabular}{lcccccccc}
		\toprule 
		& \multicolumn{4}{c}{\text{Cityscapes$\rightarrow$Foggy}} 
		& \multicolumn{4}{c}{\text{Cityscapes$\rightarrow$Vistas}} \\
		Method 
		& $\textbf{PQ}_{16}$ & $\text{RQ}_{16}$ & $\text{SQ}_{16}$ & $\textbf{PQ}_{19}$ 
		& $\textbf{PQ}_{16}$ & $\text{RQ}_{16}$ & $\text{SQ}_{16}$ & $\textbf{PQ}_{19}$ \\
		\midrule
		CVRN \citep{huang2021cross} 
		& 35.7 & 46.7 & 72.7 & \no 
		& 33.5 & 42.8 & 73.8 & \no \\
		UniDAF \citep{zhang23cvpr} 
		& 37.6 & 49.5 & 72.9 & \no 
		& \no & \no & \no & \no \\
		\midrule
		EDAPS \citep{Saha_2023_ICCV} 
		& 56.7 & 70.5 & 79.2 & \no
		& 41.2 & 53.4 & 75.9 & \no \\
		LIDAPS~\citep{mansour2025wacv} & \text{59.6} & \text{73.2} & \text{80.2} & \text{} & \text{42.6} & \text{54.9} & \text{76.6} & \text{}\\
		MC-PanDA~\citep{martinovic2024eccv} & \underline{63.7} & \underline{76.3} & \underline{82.5} & \underline{62.0} & \underline{53.8} & \underline{66.6} & \underline{79.3} &\underline{51.7} \\ 
		\rowcolor{gray!10} 
		\ours{}  
		& \textbf{66.6} & \textbf{79.2} & \textbf{83.2} & \textbf{64.7}
		& \textbf{55.9} & \textbf{68.3} & \textbf{80.3} & \textbf{54.3} \\
		\bottomrule
	\end{tabular}
\end{table*}

We apply random scaling, horizontal flipping, color jittering,
and random $512\times1024$ cropping in both domains. 
The target-domain branch in the student receives additional strong augmentations 
as described in section~\ref{sec:uda_basics}. 
The teacher generates pseudo-labels for the target domain via default 
panoptic inference~\citep{cheng2022masked},
and the student is trained with a consistency objective modulated 
by our mask-wide loss scaling and points sampling procedure. 
See Appendix~\ref{sec:implementation_details_appendix} for additional 
training and augmentation details.

All experiments with the base model variant are conducted on a single 
A100--40GB GPU, whereas the large variant requires two GPUs. 
All results are reported as the mean over three runs with different random seeds.\\
\bmhead{Evaluation}
We evaluate our models according to
panoptic quality (PQ)~\citep{kirillov2019panoptic}
that can be factored into segmentation quality (SQ) 
and recognition quality (RQ).
We report PQ for each category (Appendix~\ref{appendix:per_class_results}),
as well as the mean PQ, of the final student model.
Synthia~\citep{Ros_2016_CVPR} comprises
16 annotated classes that correspond
to a subset of the Cityscapes taxonomy.
Consequently, the experiments on Synthia
report $\text{PQ}_{16}$ as the mean
over the 16 Synthia classes.

\subsection{Comparison with the SotA}\label{subsec:sota_comparison}
We first compare the results of \ours{} with the state of the art on the four benchmarks commonly used in the domain-adaptive panoptics. 
Table~\ref{tab:synthetic_to_real_mean} summarizes the results on standard synthetic-to-real benchmarks.
Consistent with our conference findings~\citep{martinovic2024eccv}, the original MC-PanDA formulation already surpasses EDAPS~\citep{Saha_2023_ICCV} when equipped with either the MiT-B5~\citep{NEURIPS2021_segformer} or the comparable Swin-B~\citep{liu2021swin} backbone. 
MC-PanDA also exceeds the performance of LIDAPS~\citep{mansour2025wacv}, despite LIDAPS leveraging MiT-B5 combined with CLIP-based~\citep{radford2021learning} textual embeddings. 
\textcolor{changesframe}{To further assess the effect of the backbone, Table~\ref{tab:synthetic_to_real_mean} also includes our EDAPS and LIDAPS reimplementations with the DINOv2-B backbone, the same encoder as in \ours{}. This improves some baseline results, but \ours{} remains ahead on both benchmarks. Additional details required for these reimplementations are provided in Appendix~\ref{app:dinov2_edaps_lidaps}.}
Most importantly, our extended \ours{} improves upon the original MC-PanDA by \text{2.2} \pqs{} points on Synthia$\rightarrow$Cityscapes and achieves a substantial \text{6.1} \pqs{} points gain on the more challenging Synthia$\rightarrow$Vistas benchmark.

\textcolor{changesframe}{
We provide a detailed compute, memory, and inference-time comparison with LIDAPS in
Appendix~\ref{app:compute_memory_comparison}. In summary, LIDAPS uses fewer training
iterations and therefore requires fewer total GPU-hours, while \ours{} is faster at inference
(3.4 FPS vs. 2.1--2.4 FPS), uses less inference memory (4.0 GiB vs. 6.6--8.2 GiB), and
achieves higher final performance.
}

Table~\ref{tab:city_to_other_mean} further compares \ours{} with the state of the art on real-to-adverse (Cityscapes$\rightarrow$Foggy Cityscapes) and real-to-real (Cityscapes$\rightarrow$Vistas) benchmarks. Consistent with the observations in Table~\ref{tab:synthetic_to_real_mean}, MC-PanDA surpasses both EDAPS and LIDAPS on these two settings. Finally, \ours{} delivers additional gains over MC-PanDA, improving performance by approximately 2.5 \pqn{} points on both benchmarks. We provide per-class
comparison with
the state of the art on both
synthetic-to-real and
clear-to-adverse benchmarks
in the Appendix~\ref{appendix:per_class_results}.

\subsection{Ablating MC-PanDA}\label{subsec:ablating_mcpanda}
Before introducing \ours{},
we ablate MC-PanDA~\citep{martinovic2024eccv}
and examine alternative formulations
of its key components,
laying the foundation for 
the improvements developed
in the following sections.

\begin{table}[b!]
\centering
\caption{Ablation study on Synthia$\rightarrow$Cityscapes and Synthia$\rightarrow$Vistas with the Swin-B backbone, as originally reported for MC-PanDA~\citep{martinovic2024eccv}.
Top row corresponds to supervised training on Synthia.
We report mean$_{\pm\text{std}}$ over three random seeds. BC: baseline consistency (section~\ref{sec:uda_basics}).}
\label{tab:main_ablation_synthia_swin}
\footnotesize
\setlength{\tabcolsep}{3.5pt}
\begin{tabular}{cccc@{\hskip1.5pt}lcc@{\hskip1.5pt}l}
\toprule
& & & \multicolumn{2}{c}{\text{SYN$\rightarrow$City}} & & \multicolumn{2}{c}{\text{SYN$\rightarrow$Vistas}} \\
\cmidrule(lr){4-5} \cmidrule(lr){7-8}
BC & $\text{MLS}_{.99}$ & $\text{CBPF}$ & \multicolumn{2}{c}{$\text{PQ}_{16}$}
& &  \multicolumn{2}{c}{$\text{PQ}_{16}$} \\
\midrule
\no  & \no & \no  & 30.8 & \blarrow  
& &  25.2 & \blarrow \\

\yes & \no & \no  & \mstd{39.6}{1.5} & \textcolor{Improved}{+8.8} &  & \mstd{32.2}{0.9} & \textcolor{Improved}{+7.0}  \\

\yes & \yes & \no  & \mstd{44.6}{0.7} & \textcolor{Improved}{+13.8} &  & \mstd{37.1}{1.0} & \textcolor{Improved}{+11.9} \\

\yes & \no & \yes & \mstd{44.1}{0.2} & \textcolor{Improved}{+13.3}  &  & \mstd{35.1}{1.3} & \textcolor{Improved}{+9.9} \\

\yes & \yes & \yes & \textbf{\mstd{47.4}{0.8}} & \textcolor{Improved}{\textbf{+16.6}} &  & \textbf{\mstd{38.7}{1.0}} & \textcolor{Improved}{\textbf{+13.5}} \\
\bottomrule
 \end{tabular}
\end{table}

Table~\ref{tab:main_ablation_synthia_swin} quantifies 
the contributions of the proposed
\underline{M}ask-wide \underline{L}oss \underline{S}caling (MLS)
and \underline{C}onfidence-\underline{b}ased \underline{P}oint \underline{F}iltering (CBPF) 
to the panoptic performance (\pqs{})
of adapted models on two synthetic-to-real benchmarks: 
Synthia$\rightarrow$Cityscapes and Synthia$\rightarrow$Vistas. 
The first row reports
the domain generalization
performance of a supervised model 
trained on Synthia. 
Our consistency baseline
(see~\ref{sec:uda_basics}) improves supervised baseline 
by 8.8 and 7.0 \pqs{} points on Synthia$\rightarrow$Cityscapes and Synthia$\rightarrow$Vistas, respectively. 
Self-training with MLS provides an additional gain of roughly 5 points on both benchmarks, 
amounting to improvements of approximately 14.0 and 12.0 \pqs{} points over the supervised baseline. 
Similarly, CBPF yields gains of about 5.0 and 3.0 \pqs{} points over the consistency baseline. 
Combining both components, our complete method reaches 47.4 and 38.7 \pqs{}, 
corresponding to improvements of 16.6 and 13.5 \pqs{} points over the supervised, 
and 7.8 and 6.5 \pqs{} points over the consistency baseline (BC).

We next investigate
alternative formulations
of our proposed contributions.
Table~\ref{tab:confidence_calc_ablation} 
compares our mask-wide loss scaling (MLS)
with image-wide loss scaling (ILS),
which assigns all masks in an image 
a single global scaling factor~\citep{Saha_2023_ICCV}.
The image-wide loss weight is defined as: 
\begin{align}
    \lambda_i=\lambda^\text{ILS} = \frac{\sum_{r,c} \llbracket \Phi^\text{teach}_{r,c} > \tau_{\text{ILS}} \rrbracket}{HW}.
\end{align}
We evaluate three ILS thresholds 
$\tau_\text{ILS} \in \{0.95, 0.968, 0.99\}$ 
and report the best-performing configuration 
$\tau_\text{ILS}=0.968$~\citep{Saha_2023_ICCV} as an optimistic baseline. 
The top section of the table compares ILS and MLS 
without confidence-based point filtering (CBPF), 
while the bottom section includes CBPF. 
We observe a clear advantage
of mask-wide loss scaling in both cases.
We argue that this happens since
ILS down-scales the loss gradients
for many valid masks.

\begin{table}[h!]
\centering
\caption{
Comparison of image-wide (ILS) and mask-wide loss scaling (MLS) on Synthia$\rightarrow$Cityscapes, 
as originally reported for MC-PanDA~\citep{martinovic2024eccv}.
We report an optimistic ILS performance as the maximum across three different $\tau_\text{ILS}$ thresholds.
}
\label{tab:confidence_calc_ablation}
\footnotesize
\begin{tabular}{cccccc}
\toprule
ILS & MLS & CBPF & SQ$_{16}$ & RQ$_{16}$ & \textbf{PQ}$_{16}$ \\
\midrule
\yes & \no & \no & \mstd{73.7}{0.5} & \mstd{50.5}{1.1} & \mstd{39.7}{0.8} \\
\no & \yes & \no &\mstd{75.3}{0.2} & \mstd{56.5}{1.0} & \mstd{44.7}{0.7} \\
\midrule
\yes & \no & \yes &\mstd{74.8}{0.4} & \mstd{53.4}{0.2} & \mstd{42.1}{0.5} \\
\rowcolor{gray!10}\no & \yes & \yes &\mstd{76.7}{0.4} & \mstd{59.3}{1.0} & \mstd{47.4}{0.8} \\
\bottomrule
\end{tabular}
\end{table}

Table~\ref{tab:table_ubpf_variant_ablation} validates our proposed loss 
subsampling strategy based on CBPF (\textit{cf}.~\ref{sec:cbpf}). 
The baseline point-sampling approach~\citep{kirillov2020pointrend} prioritizes points 
with high student uncertainty, which increases the likelihood of selecting 
incorrect pseudo-labels. 
CBPF mitigates this issue by suppressing training on pixels for which the teacher 
exhibits low confidence. 
We further evaluate
an alternative formulation of
teacher confidence,
${\Phi}^\text{teach}_{i,r,c}=\sigma(|s^\text{teach}_i|)_{r,c}$,
where $s^\text{teach}_i$ denotes the teacher pre-activation for mask $i$.
This variant produces a per-mask confidence estimate,
in contrast to our default all-mask formulation (see Eq.~\ref{eq:conf_def}).
We also include a random point-sampling
baseline (random sampling, $\beta = 0.0$).
Our formulation, which aggregates confidence over all masks, 
outperforms the per-mask variant by 2.7 PQ points.  
We believe that per-mask underperformance occurs due to 
over-confident negative mask assignments where many
pixels get incorrectly predicted as not belonging to the corresponding mask.
See Appendix~\ref{ref:appendix_allmask_permask}
for a visual comparison of the \textit{all-mask} and \textit{per-mask} strategies.

\begin{table}[h!]
	\centering
	\caption{
		Validation of loss subsampling with confidence-based point filtering (CBPF) 
		on Synthia$\rightarrow$Cityscapes, as originally reported for MC-PanDA~\citep{martinovic2024eccv}.
		The random sampling baseline applies the loss to a random subset of image points.
	}
	\label{tab:table_ubpf_variant_ablation}
	\scriptsize
	\footnotesize
	\begin{tabular}{lccc}
		\toprule
		CBPF ($\tau_2=0.8)$ \\
		filtering method & $\text{SQ}_{16}$ & $\text{RQ}_{16}$ & $\textbf{PQ}_{16}$ \\
		\midrule
		$\text{random sampling}$ & \mstd{74.9}{0.1} & \mstd{56.1}{1.4} & \mstd{44.2}{1.1} \\
		$\text{per-mask}$ & \mstd{75.2}{0.4} & \mstd{56.5}{0.1} & \mstd{44.7}{0.1} \\
		\rowcolor{gray!10}$\text{all-masks}~(\text{eq.}\ \ref{eq:conf_def})$ & \mstd{76.7}{0.4} & \mstd{59.3}{1.0} & \mstd{47.4}{0.8} \\
		\bottomrule
	\end{tabular}   
\end{table}

\subsection{From MC-PanDA to \ours{}} \label{sec:mcpandapp_components}
This section presents the evolution from the preliminary MC-PanDA to the improved \ours{}. We begin by studying backbone initialization and demonstrating that MLS and CBPF remain effective under strong self-supervised encoders. We then extend the evaluation to additional synthetic-to-real and clear-to-adverse benchmarks, further validating the effectiveness of these components. Next, we streamline the original three-stage pipeline into a single-stage design and introduce per-class adaptive mask-wide loss scaling, which stabilizes training and removes sensitivity to the initial threshold. Finally, we demonstrate the robustness of CBPF, completing the transition to the final \ours{} framework.
 
\subsubsection{On the backbone choice} \label{subsec:on_the_backbone_choice}
\begin{figure}[t!]
\centering
\includegraphics[width=\columnwidth,trim=0 5 0 5, clip]{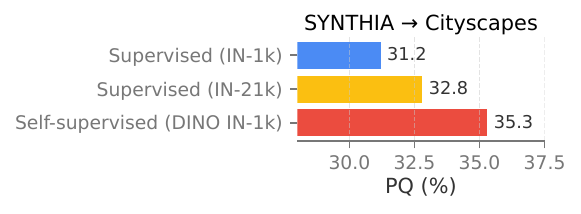}
\caption{Comparison of initialization strategies. MC-PanDA with a self-supervised ViT-B/16 (DINO, IN-1k) yields the best PQ, outperforming supervised ViT-B/16 models pre-trained on ImageNet-1k and even ImageNet-21k. Results are averaged over three seeds.}
\label{fig:self-sup-vit-motivation}
\end{figure}

We first validate the hypothesis
that self-supervised initialization
offers greater robustness to
domain shift than supervised pre-training.
To this end,
we evaluate the original MC-PanDA~\citep{martinovic2024eccv} using a ViT-B-16~\citep{dosovitskiy2021an}
backbone with three distinct
initialization strategies:
supervised pre-training on ImageNet-1k (IN-1k) and ImageNet-21k (IN-21k),
and self-supervised DINO~\citep{caron2021emerging} trained on ImageNet-1k.
Figure \ref{fig:self-sup-vit-motivation} reports the results on the Synthia$\rightarrow$Cityscapes benchmark.
The self-supervised DINO
initialization outperforms the supervised counterparts,
exceeding the IN-1k baseline by 4.2 PQ and even the IN-21k
baseline by 2.5 PQ.
\\

\begin{figure}[h!]
\includegraphics[width=\linewidth]{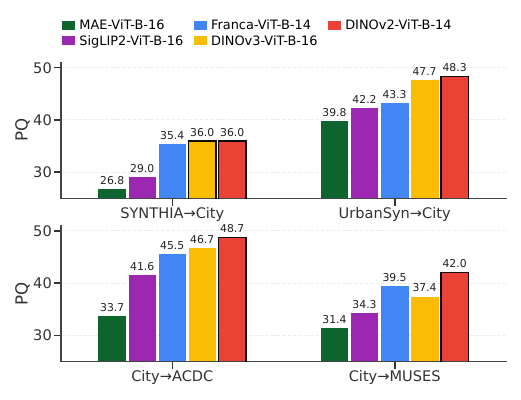}
\color{changesframe}\caption{
Selection of the pretrained backbone. We compare different pretraining strategies in a
domain-generalization setting across four benchmarks: reconstruction-based self-supervised
pretraining (MAE), vision-language pretraining (SigLIP2), and distillation-based
self-supervised pretraining (Franca, DINOv2, and DINOv3).}
\label{fig:choosing_selfsup_backbone}
\end{figure}
After establishing that self-supervised initialization benefits UDA, we evaluate several
large-scale pretrained encoders to identify the most suitable initialization.
\textcolor{changesframe}{
To better contrast different pretraining strategies, we consider three families:
reconstruction-based self-supervised pretraining, represented by MAE~\citep{he2022masked};
vision-language pretraining, represented by SigLIP2~\citep{tschannen2025siglip}; and
distillation-based self-supervised pretraining, represented by
Franca-B~\citep{venkataramanan2025franca}, DINOv2-B~\citep{oquab2024dinov}, and
DINOv3-B~\citep{simeoni2025dinov3}.}
To assess their intrinsic robustness, we compare all encoders in a domain-generalization
setting, where models train only on the source domain without access to target-domain images.
All experiments use Mask2Former~\citep{cheng2022masked} with identical hyperparameters to
ensure a fair comparison.
Figure~\ref{fig:choosing_selfsup_backbone} summarizes results across four benchmarks:
\textcolor{changesframe}{
MAE is consistently weaker, suggesting that reconstruction-based pretraining alone provides
less robust cross-domain features in our setting. SigLIP2 achieves moderate performance,
indicating that vision-language pretraining improves generalization but does not match the
strongest self-supervised distillation models. Among the distillation-based methods, DINOv2-B
provides the strongest and most consistent domain-generalization performance.}
We therefore adopt DINOv2-B as the default backbone for all subsequent experiments.

\subsubsection{MC-PanDA with DINOv2} \label{subsec:ablating_self-sup_init}
From this section onwards, all experiments
use self-supervised DINOv2~\citep{oquab2024dinov} 
as the backbone. 
We begin by directly
replacing the Swin-B in MC-PanDA
with DINOv2-B 
and evaluate whether our contributions,
MLS and CBPF,
remain effective under
this substantially 
stronger initialization.
Table~\ref{tab:main_ablation_synthia_dinov2} reports the ablation
results on two 
synthetic-to-real benchmarks: Synthia$\rightarrow$Cityscapes and Synthia$\rightarrow$Vistas.
The first row provides the domain-generalization performance of a model trained 
only on Synthia.
Compared to the supervised Swin-B initialization in 
Table~\ref{tab:main_ablation_synthia_swin}, 
self-supervised DINOv2-B improves performance by 5.2 and 8.5 \pqs{} points 
on the two benchmarks, respectively.

%%%%%%%%%%%%%%%%%%%% TABLE %%%%%%%%%%%%%%%%%%%
% Ablation study w/ Dinov2, Synthia->* 
%%%%%%%%%%%%%%%%%%%% TABLE %%%%%%%%%%%%%%%%%%%
\begin{table}[b!]
\centering
\caption{Ablation study on Synthia$\rightarrow$Cityscapes and Synthia$\rightarrow$Vistas with the self-supervised DINOv2 initialization.
Top row corresponds to supervised training on Synthia.
We report mean$_{\pm\text{std}}$ over three random seeds.}
\label{tab:main_ablation_synthia_dinov2}
\footnotesize
\setlength{\tabcolsep}{3.5pt}
\begin{tabular}{cccc@{\hskip1.5pt}cc@{\hskip1.5pt}c@{\hskip1.5pt}c}
\toprule
& & & \multicolumn{2}{c}{\text{Syn$\rightarrow$City}} & & \multicolumn{2}{c}{\text{Syn$\rightarrow$Vistas}} \\
\cmidrule(lr){4-5} \cmidrule(lr){7-8}
BC & $\text{MLS}_{.99}$ & $\text{CBPF}_{0.8}$ & \multicolumn{2}{c}{$\textbf{PQ}_{16}$} & &  \multicolumn{2}{c}{$\textbf{PQ}_{16}$} \\
\midrule
\no  & \no & \no  & 36.0 & \blarrow 
     & &  33.7 & \blarrow \\

\yes & \no & \no  & \mstd{39.7}{0.9} & \textcolor{Improved}{\text{+3.7}} 
                  & & \mstd{33.6}{1.6} & \textcolor{Degraded}{-0.1} \\

\yes & \yes & \no  & \mstd{44.5}{0.6} & \textcolor{Improved}{\text{+8.5}} 
                   & &  \mstd{42.8}{0.4} & \textcolor{Improved}{+9.1} \\

\yes & \yes & \yes & \mstd{48.5}{0.4} & \textcolor{Improved}{\textbf{+12.5}} 
     & &  \mstd{44.2}{0.1} & \textcolor{Improved}{\textbf{+10.5}} \\
\bottomrule
\end{tabular}
\end{table}

The consistency baseline (BC, row 2) further adds 3.7 \pqs{} points on 
Synthia $\rightarrow$ Cityscapes and maintains a similar performance level 
to the supervised baseline on Synthia $\rightarrow$ Vistas. 
This behavior on Vistas aligns with the self-confirmation and pseudo-label 
amplification effects discussed in our conference version and illustrated in 
Fig.~\ref{fig:fig1}, which persist even with a stronger backbone.

However, once MLS is introduced, we observe substantial performance gains 
on both benchmarks, including nearly 10 \pqs{} points on Synthia $\rightarrow$ Vistas. 
This demonstrates that MLS provides complementary improvements that remain effective 
even under a state-of-the-art self-supervised initialization. 
Finally, adding CBPF on top of MLS (last row) yields additional gains of 
4.0 and 1.4 \pqs{} points on Synthia$\rightarrow$Cityscapes and 
Synthia $\rightarrow$ Vistas, respectively, confirming that both MLS and CBPF 
continue to contribute despite the significantly stronger backbone.

\subsubsection{Improving the synthetic source domain} \label{subsec:ablating_urbansyn}
\begin{figure}[b!]
	\centering
	\setlength{\tabcolsep}{0.5pt}
	\renewcommand{\arraystretch}{1.0}
	
	\begin{tabular}{cccc}
		\rotatebox{90}{\tiny Synthia} &
		\includegraphics[width=0.29\linewidth]{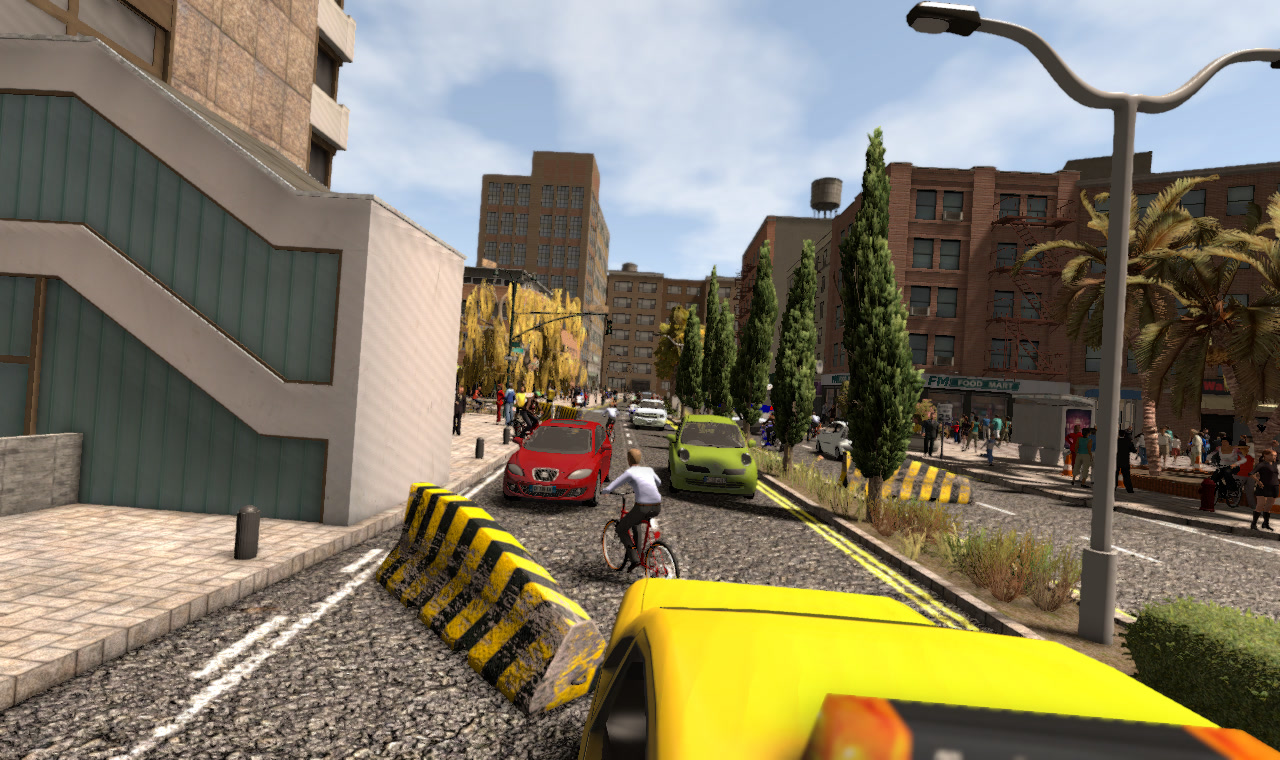} &
		\includegraphics[width=0.29\linewidth]{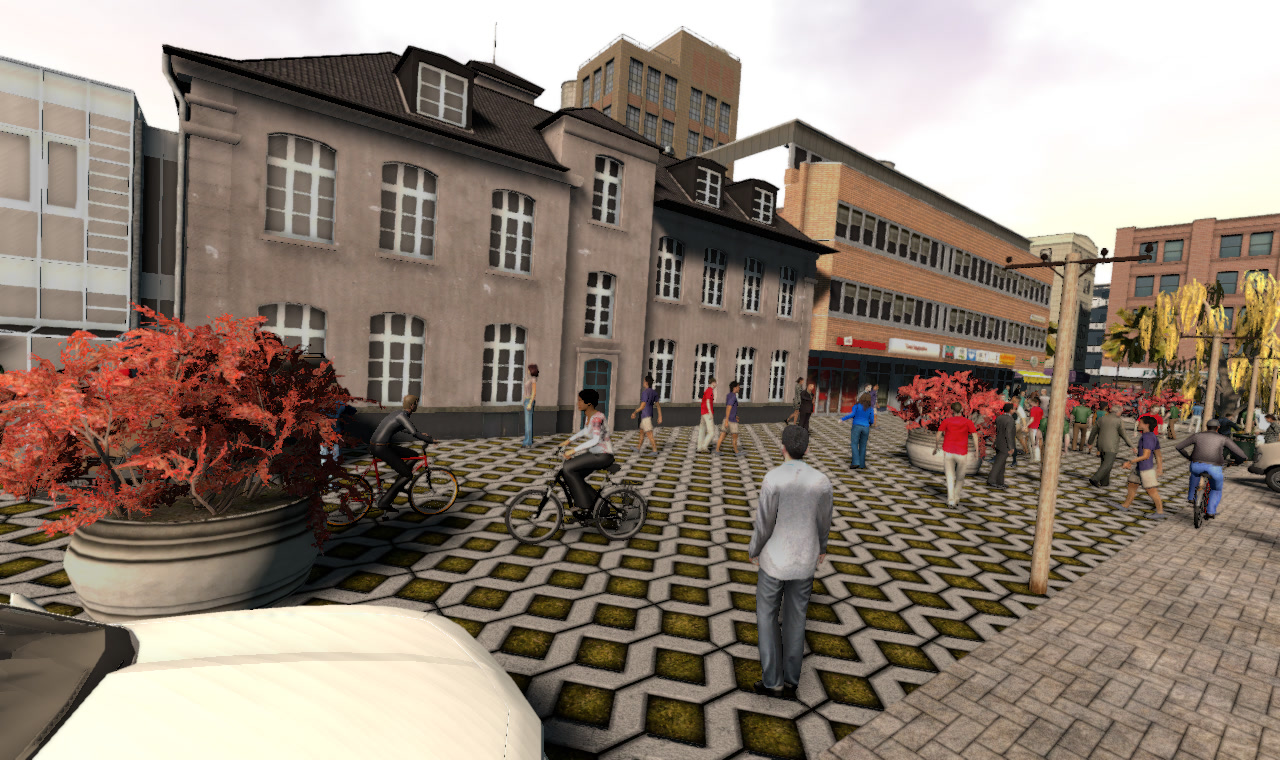} &
		\includegraphics[width=0.29\linewidth]{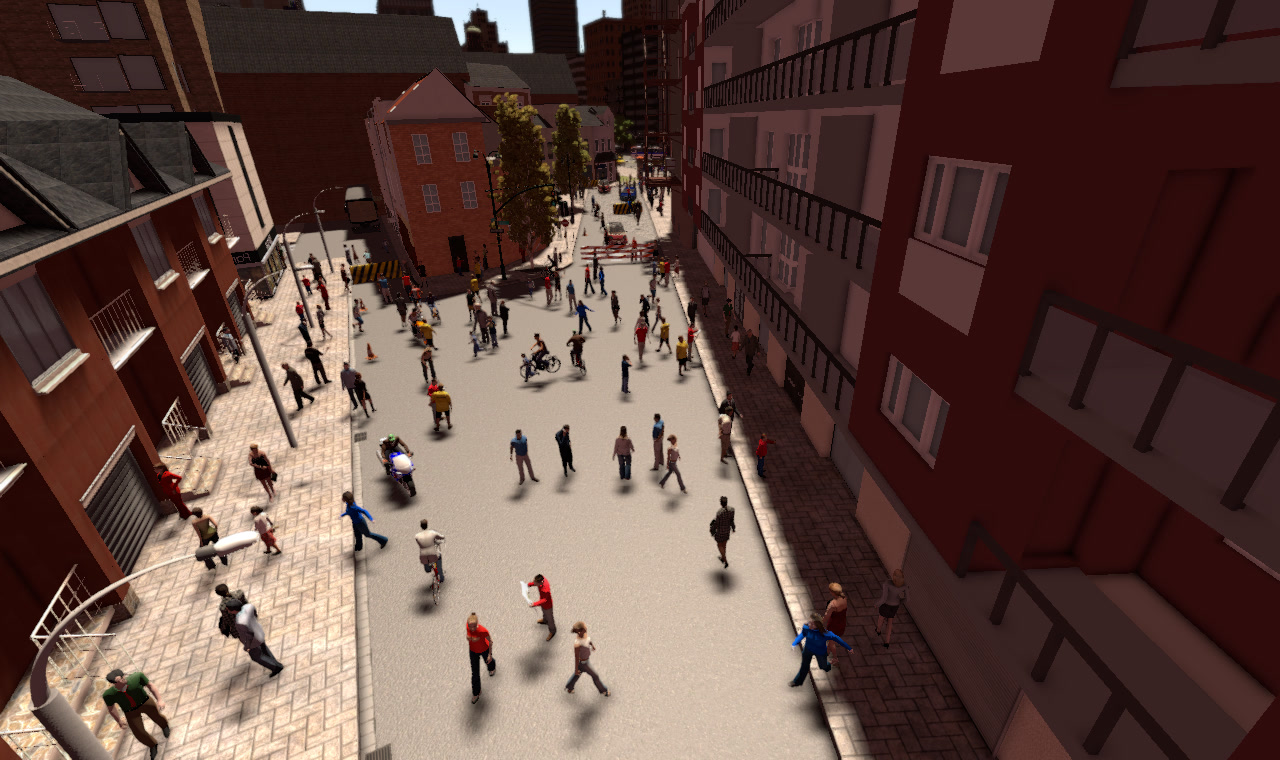} \\
		
		\rotatebox{90}{\tiny UrbanSyn} &
		\includegraphics[width=0.29\linewidth]{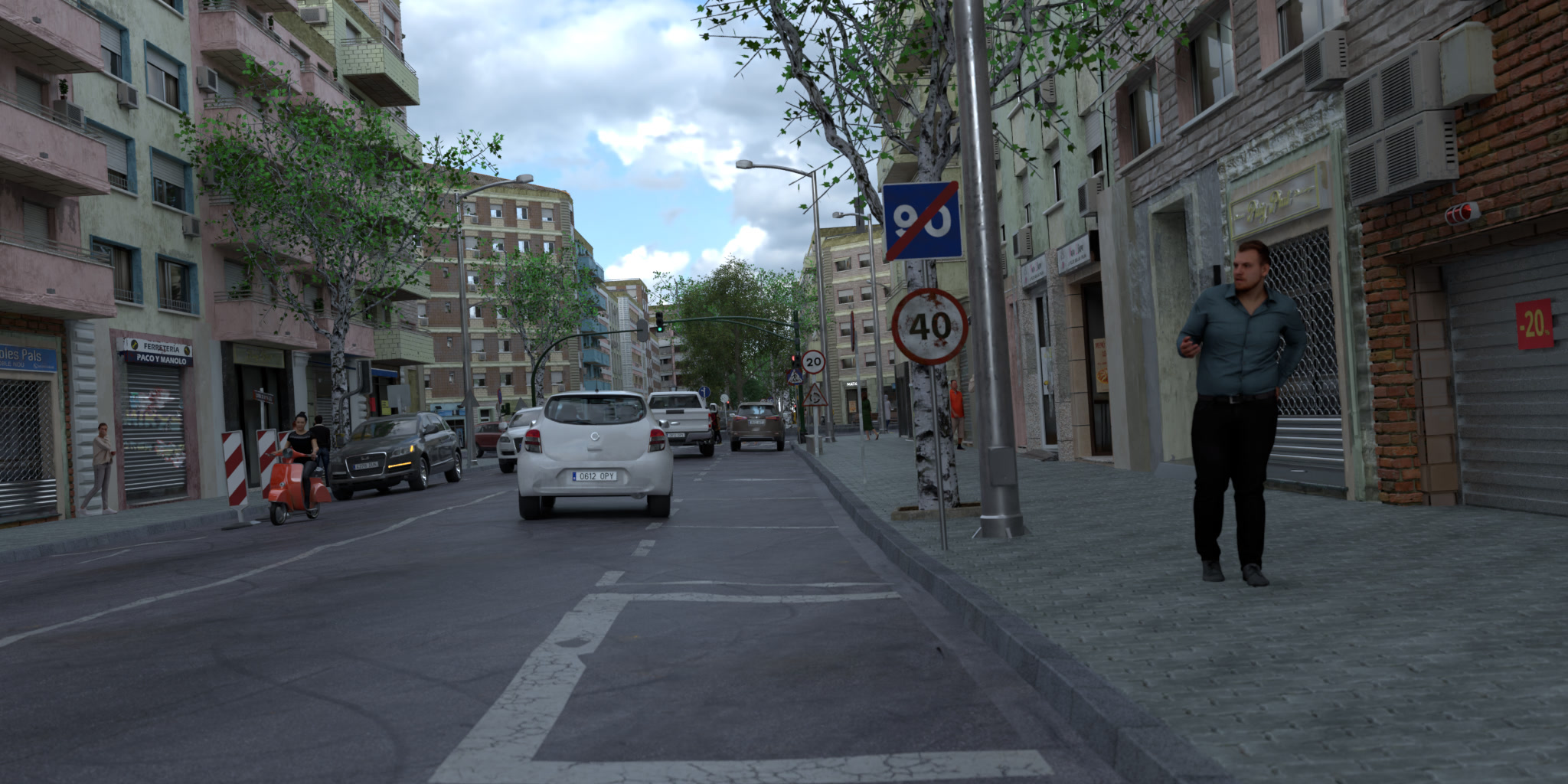} &
		\includegraphics[width=0.29\linewidth]{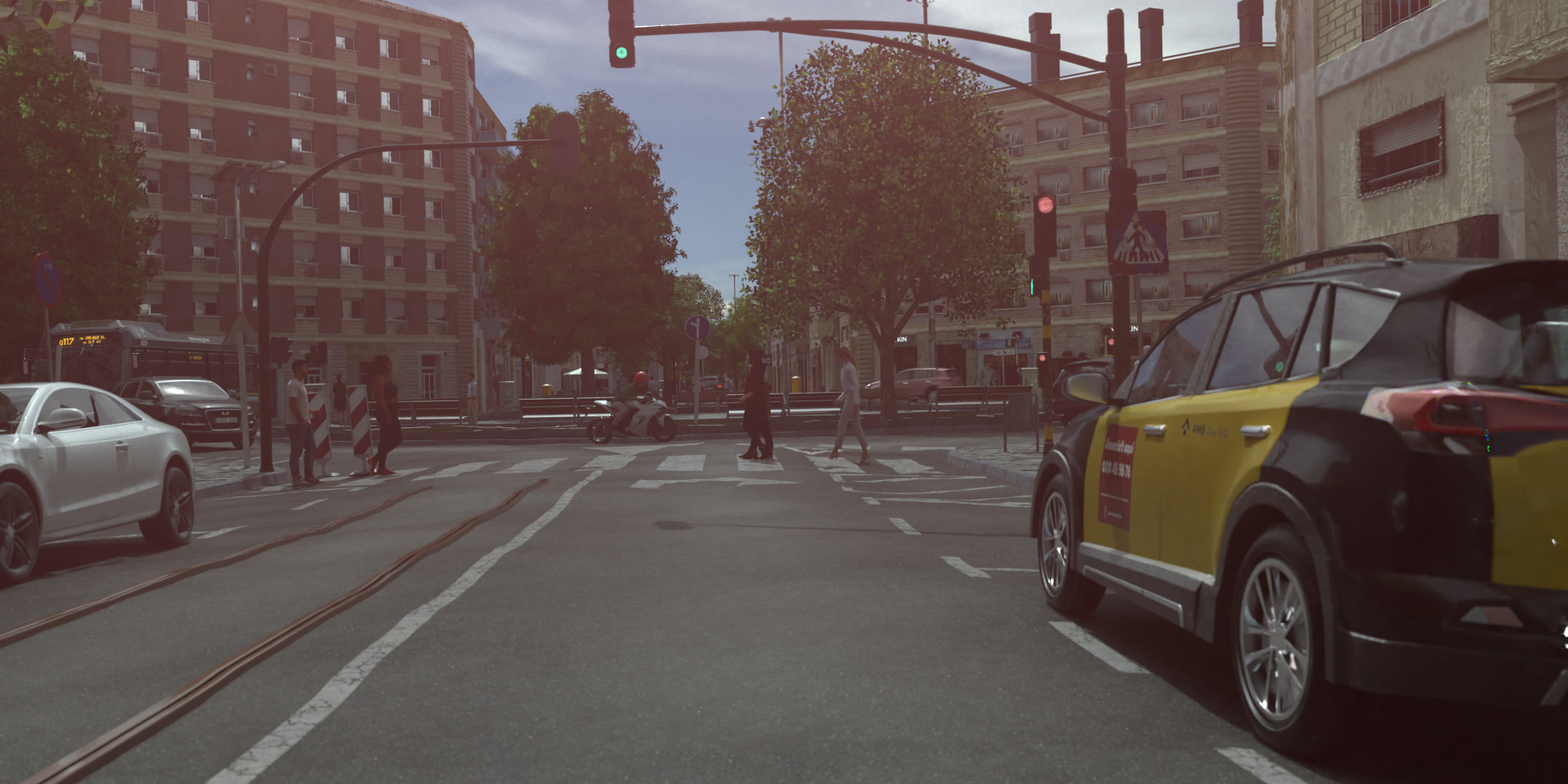} &
		\includegraphics[width=0.29\linewidth]{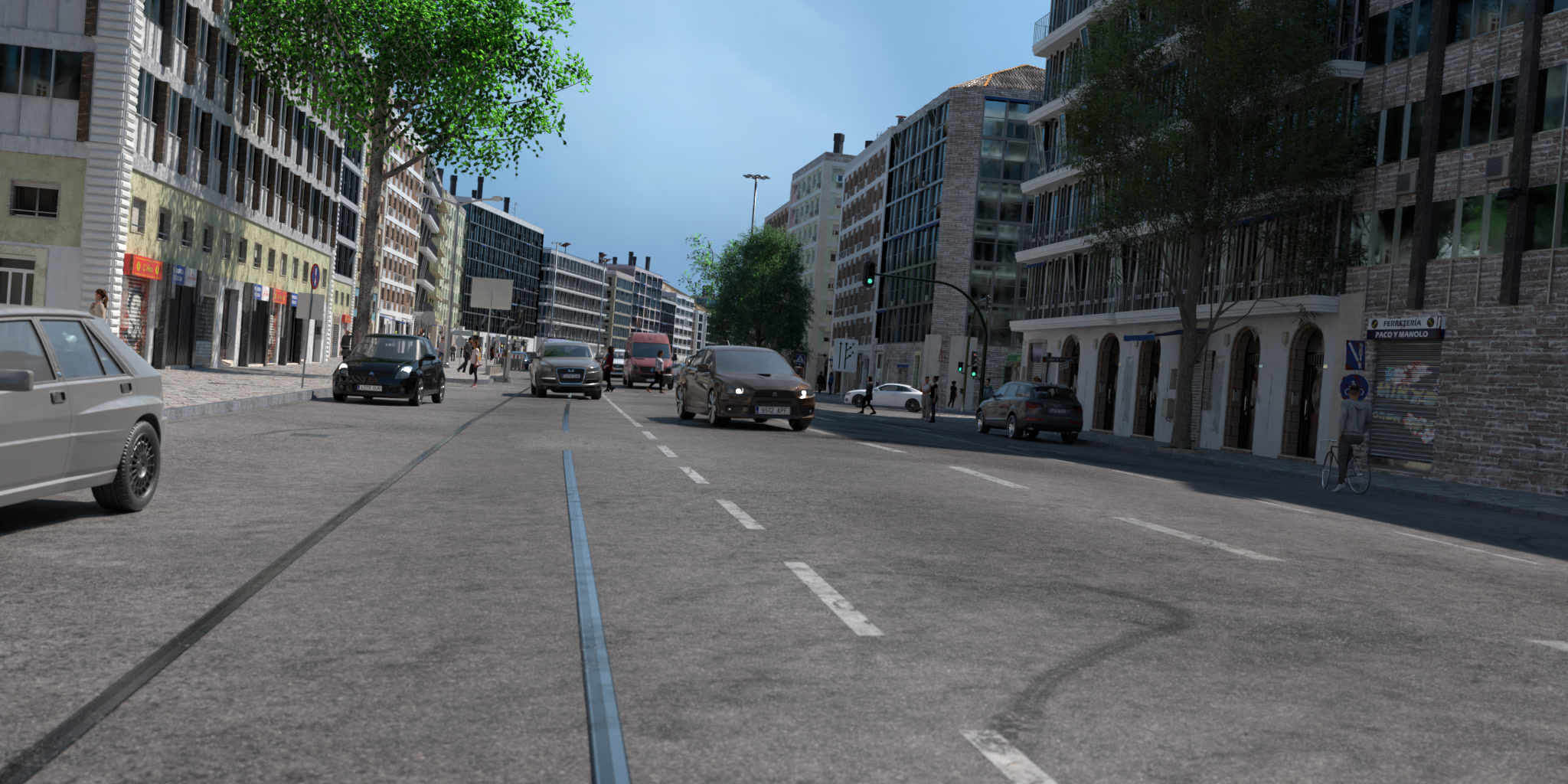}
	\end{tabular}
	
	\caption{
		Visual comparison between the synthetic datasets 
		\text{Synthia}~\citep{Ros_2016_CVPR} (top row) and 
		\text{UrbanSyn}~\citep{gomez2025urbansyn} (bottom row).
	}
	\label{fig:synthia_urbansyn_quality}
\end{figure}

We further validate
the effectiveness of our contributions
on benchmarks with a less extreme domain shift.
Synthia~\citep{Ros_2016_CVPR} has long served as the standard synthetic source domain in domain adaptive panoptic segmentation, as it was for many years the only dataset providing both semantic and panoptic labels.
However, Synthia exhibits two notable limitations: 
i) its visual appearance is often far from realistic 
(Fig.~\ref{fig:synthia_urbansyn_quality}, top), and 
ii) it provides annotations for only 16 semantic classes, 
which do not fully match the standard 19-class Cityscapes taxonomy. 
Adapting from Synthia to real-world datasets therefore remains a strong indicator of a method’s robustness due to the substantial domain gap.
However, a more realistic scenario would likely use synthetic images better aligned with the target domain.

To address such a scenario, we evaluate the recently introduced 
UrbanSyn dataset~\citep{gomez2025urbansyn} as an alternative source domain. 
As illustrated in Fig.~\ref{fig:synthia_urbansyn_quality}, UrbanSyn exhibits substantially higher visual realism. Moreover, UrbanSyn follows the standard 19-class Cityscapes taxonomy, which makes it a natural candidate for an improved synthetic-to-real benchmark.

Table~\ref{tab:main_ablation_urbansyn_dinov2} presents the ablation study of 
MC-PanDA with DINOv2 on the newly introduced benchmarks: 
UrbanSyn$\rightarrow$Cityscapes and UrbanSyn$\rightarrow$Vistas. 
From the table, we observe that MLS and CBPF together improve performance by 
8.5 and 6.6 \pqn{} points over the supervised baseline on 
UrbanSyn$\rightarrow$Cityscapes and UrbanSyn$\rightarrow$Vistas, respectively. 
Compared to the consistency baseline (BC), the combined contributions yield an 
additional 3.4 \pqn{} points on both benchmarks. 
These results indicate that even with a more realistic and better aligned synthetic 
source domain, our proposed components remain effective and provide consistent gains.
\begin{table}[h!]
\centering
\caption{Ablation study on UrbanSyn$\rightarrow$Cityscapes and UrbanSyn$\rightarrow$Vistas with the self-supervised DINOv2 initialization.
Top row corresponds to supervised training on UrbanSyn.
We report mean$_{\pm\text{std}}$ over three random seeds. USyn: UrbanSyn.}
\label{tab:main_ablation_urbansyn_dinov2}
\footnotesize
\setlength{\tabcolsep}{3.5pt}
\begin{tabular}{cccc@{\hskip1.5pt}cc@{\hskip1.5pt}c@{\hskip1.5pt}c}
\toprule
& & & \multicolumn{2}{c}{\text{USyn$\rightarrow$City}} & & \multicolumn{2}{c}{\text{USyn$\rightarrow$Vistas}} \\
\cmidrule(lr){4-5} \cmidrule(lr){7-8}
BC & $\text{MLS}_{.99}$ & $\text{CBPF}_{0.8}$ 
& \multicolumn{2}{c}{$\textbf{PQ}_{19}$} 
& & \multicolumn{2}{c}{$\textbf{PQ}_{19}$} \\
\midrule
\no  & \no & \no  & 48.3 & \blarrow 
     & & 42.1 & \blarrow \\

\yes & \no & \no  & \mstd{53.4}{0.8} & \textcolor{Improved}{\text{+5.1}} 
     & & \mstd{45.3}{1.2} & \textcolor{Improved}{+3.2} \\

\yes & \yes & \no  & \mstd{54.5}{0.4} & \textcolor{Improved}{\text{+6.2}} 
     &  & \mstd{47.3}{0.1} &  \textcolor{Improved}{+5.2}\\

\yes & \yes & \yes & \mstd{56.8}{0.7} & \textcolor{Improved}{\textbf{+8.5}} 
     & & \mstd{48.7}{0.6} & \textcolor{Improved}{\textbf{+6.6}} \\
\bottomrule
\end{tabular}
\end{table}

\begin{table}[b!]
	\centering
	\caption{Comparison of MC-PanDA with DINOv2 when using Synthia versus UrbanSyn 
		as the synthetic source domain. Results are reported in \pqs{} for the 
		16 classes shared between both datasets.}
	\label{tab:improving_synthethic_domain_comparison_pqs}
	\footnotesize
	\begin{tabular}{lcccc}
		\toprule
		Src.~domain  & \multicolumn{2}{c}{\textbf{\pqs{}} on City} & \multicolumn{2}{c}{\textbf{\pqs{}} on Vistas} \\
		\midrule
		Synthia        &\mstd{48.5}{0.4}&\blarrow&\mstd{44.2}{0.1}& \blarrow \\
		UrbanSyn       &\mstd{59.8}{0.4}& \textcolor{Improved}{\textbf{+11.3}}&\mstd{52.2}{0.3}&\textcolor{Improved}{\textbf{+8.0}} \\
		\bottomrule
	\end{tabular}
\end{table}

Complementary to improving the adaptation method itself, 
one can also improve the 
underlying synthetic source domain. 
Table~\ref{tab:improving_synthethic_domain_comparison_pqs} examines this effect by comparing the performance of MC-PanDA with DINOv2 when using Synthia or UrbanSyn as the source domain 
for adaptation to Cityscapes and Vistas. 
The comparison is reported in \pqs{} over the 16 classes available in Synthia, which 
form a subset of the 19 classes in UrbanSyn. 
We observe gains of 11.3 and 8.0 \pqs{} points on 
Cityscapes and Vistas, respectively, when switching from Synthia to UrbanSyn. 
These results suggest that, in addition to advancing domain-adaptation methods 
themselves, improving the quality and realism of the synthetic source domain can 
yield significant benefits.

\subsubsection{Clear-to-adverse benchmarks}     \label{subsec:ablating_adverse}
\begin{figure}[h!]
    \centering
    \setlength{\tabcolsep}{0.5pt}
    \renewcommand{\arraystretch}{1.0}

    \begin{tabular}{cccc}
        \rotatebox{90}{\tiny ACDC} &
        \includegraphics[width=0.31\linewidth]{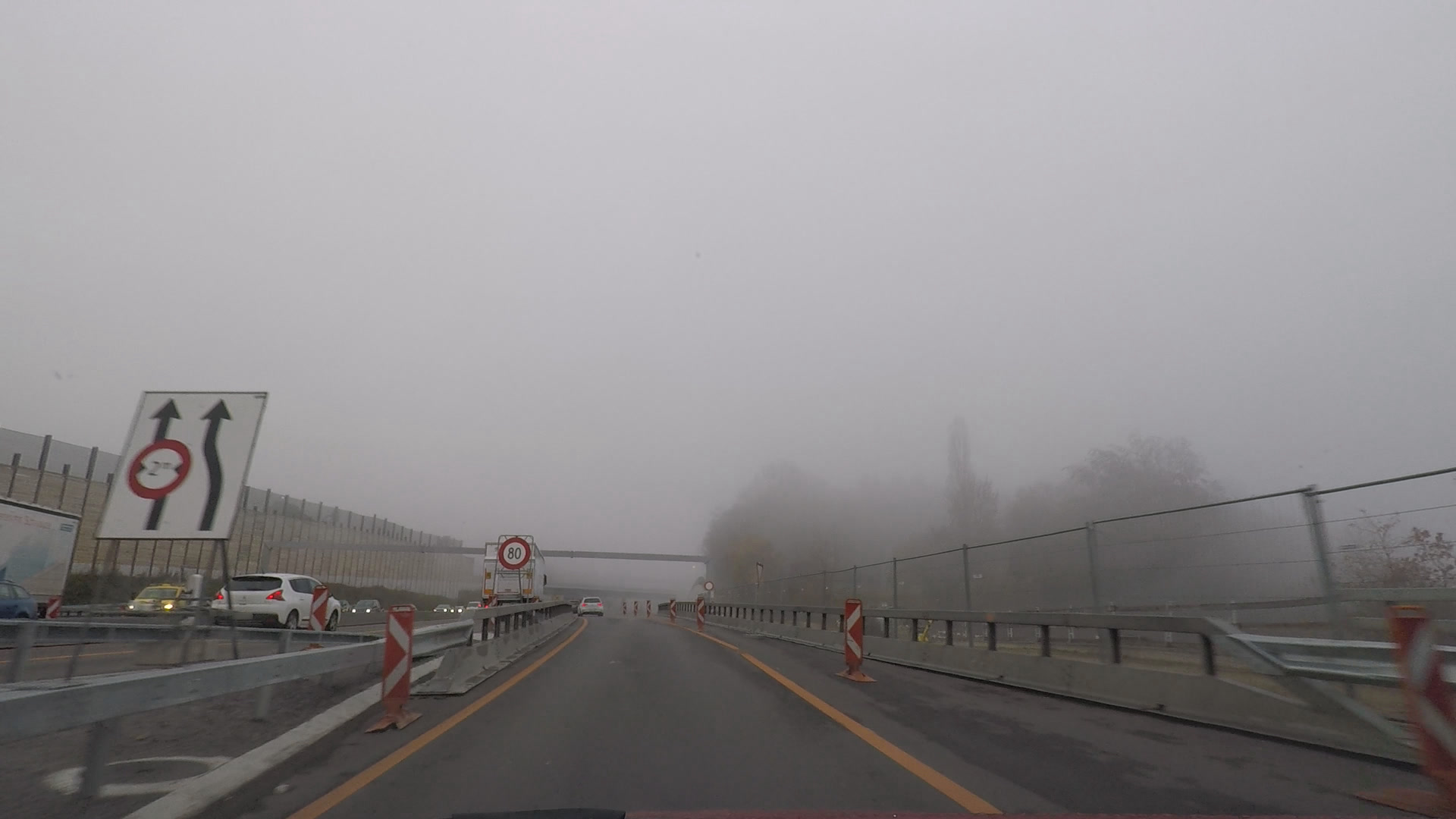} &
        \includegraphics[width=0.31\linewidth]{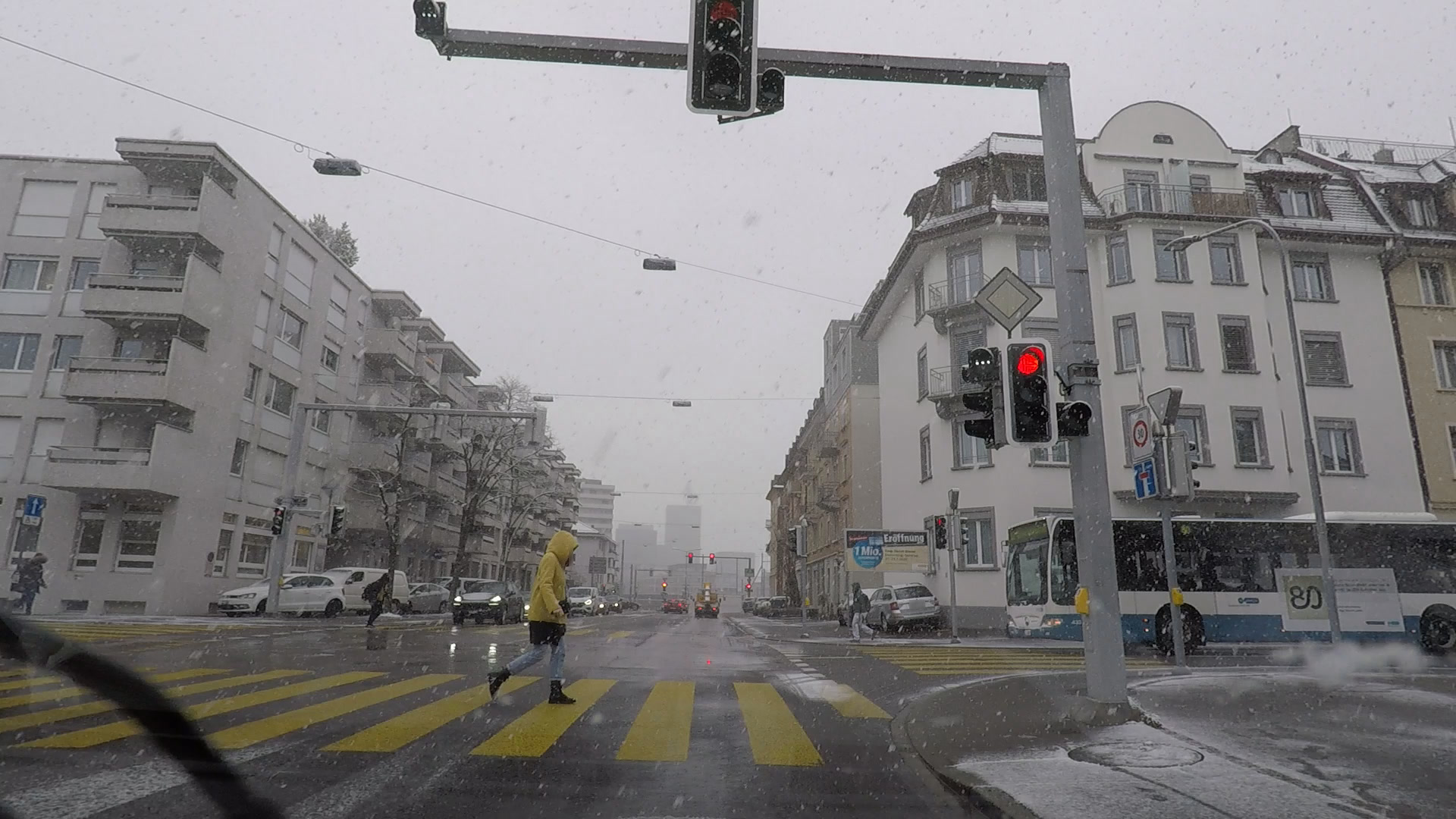} &
        \includegraphics[width=0.31\linewidth]{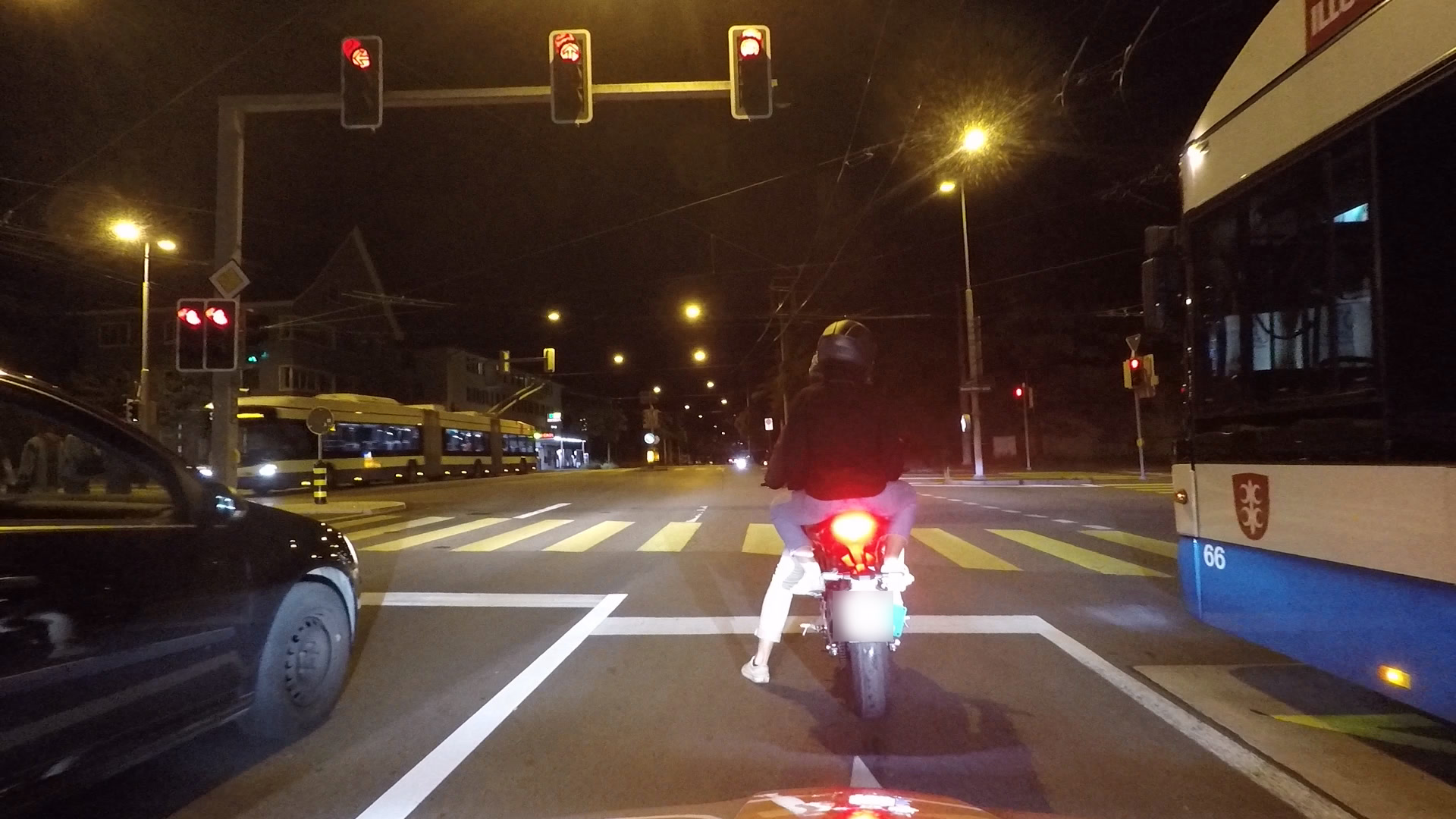} \\
        
        \rotatebox{90}{\tiny MUSES} &
        \includegraphics[width=0.31\linewidth]{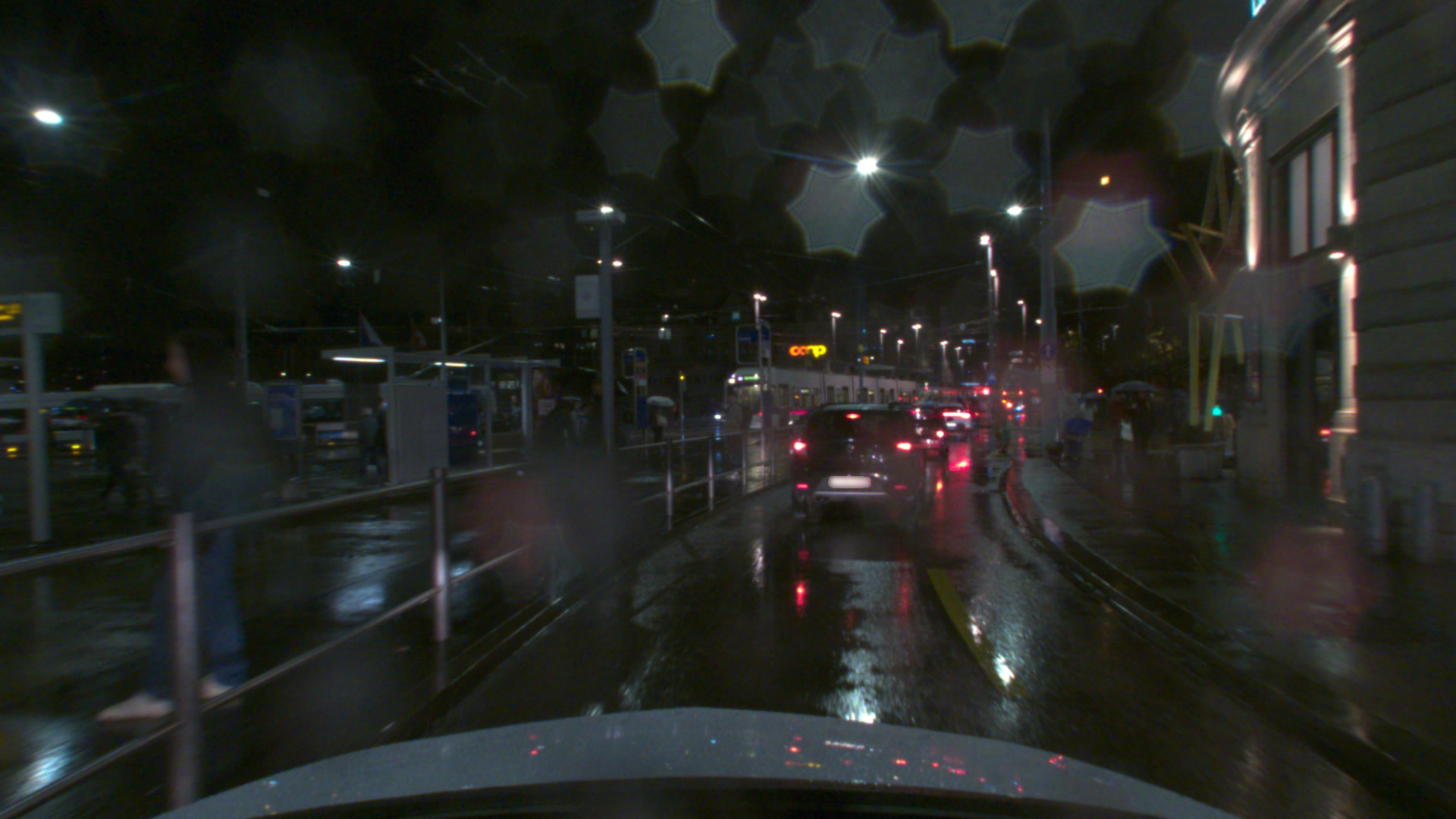} &
        \includegraphics[width=0.31\linewidth]{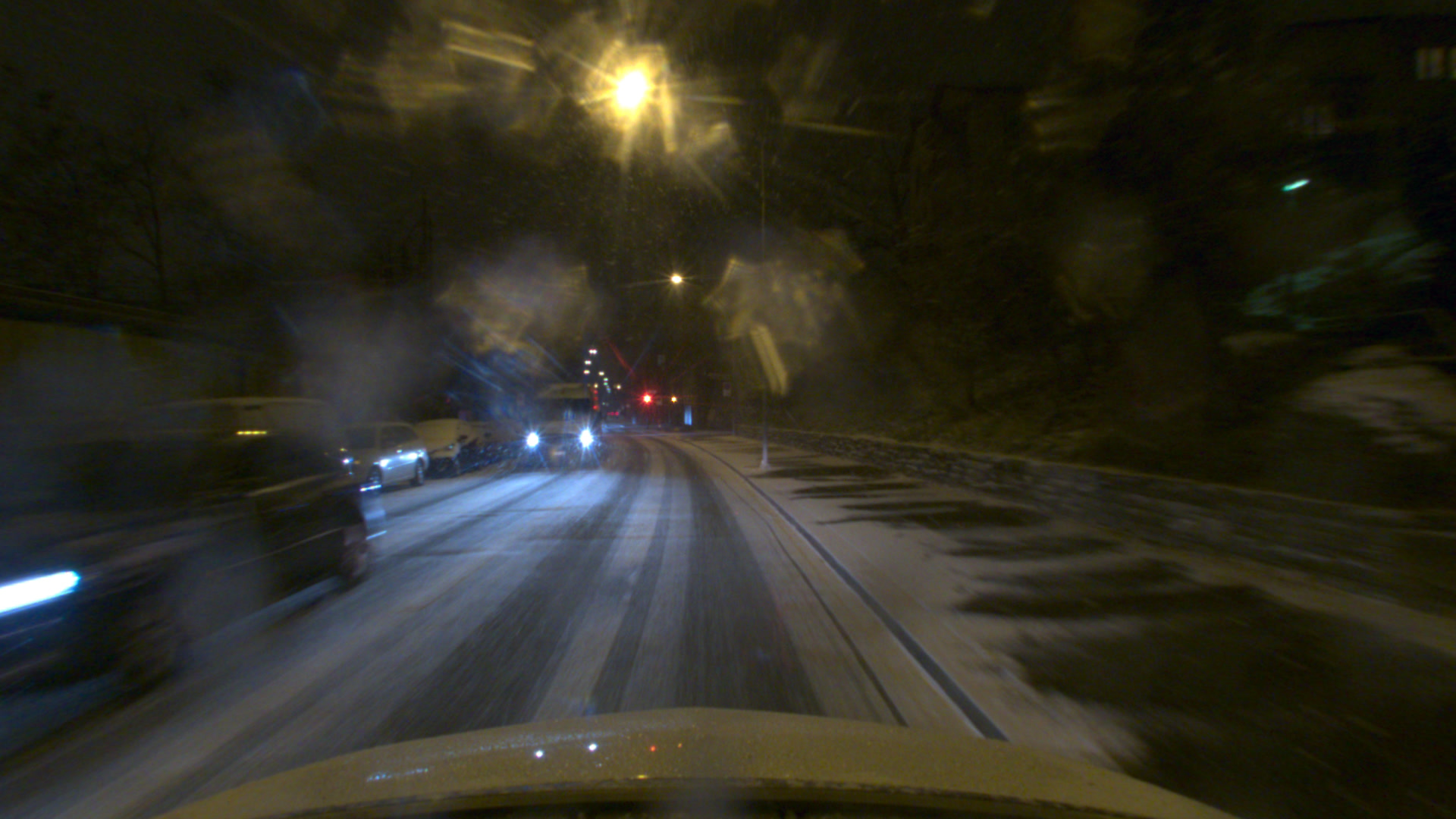} &
        \includegraphics[width=0.31\linewidth]{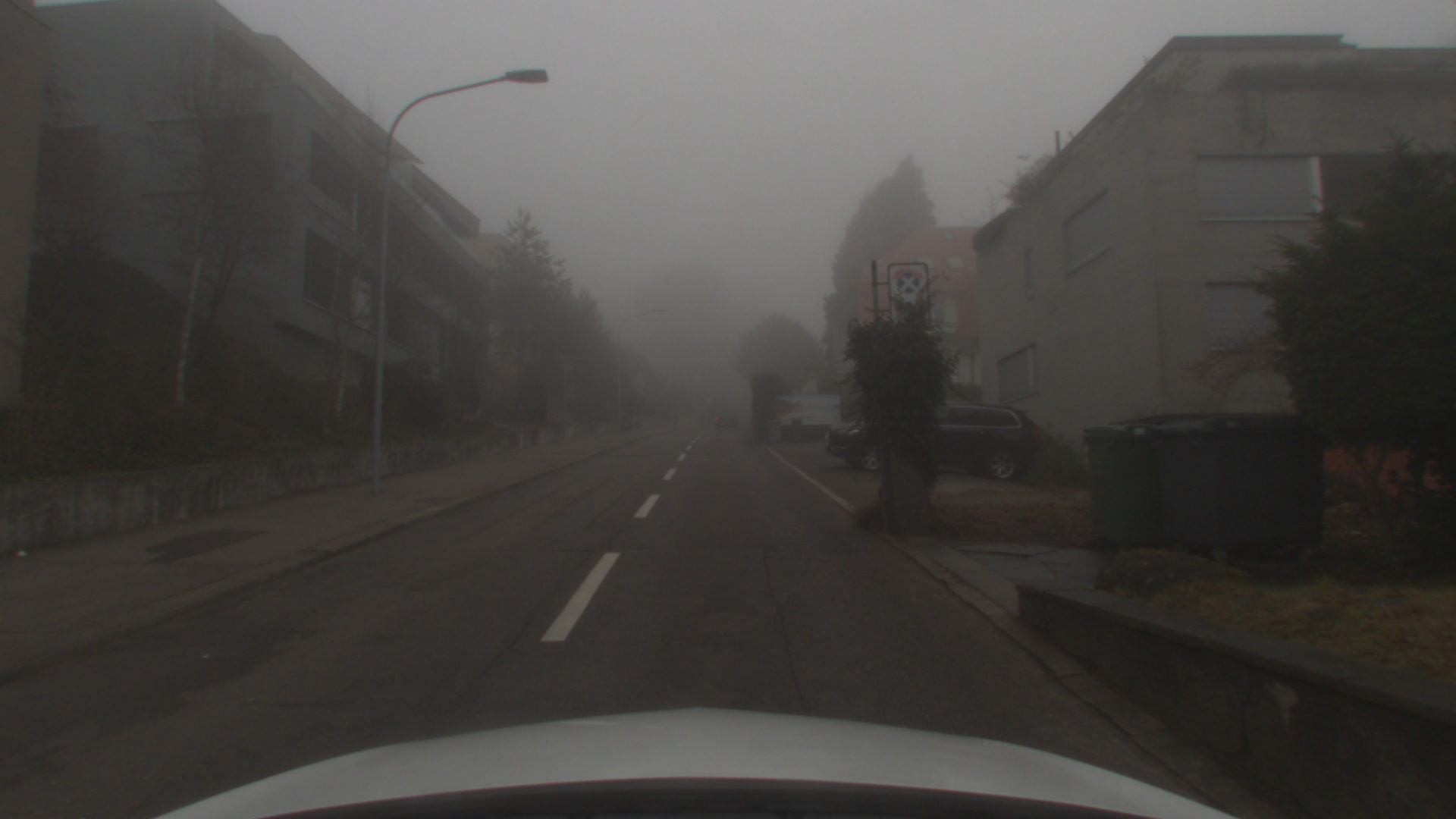}
    \end{tabular}

    \caption{
Example images from the adverse-condition datasets: 
ACDC~\citep{sakaridis21iccv} (top row) and 
MUSES~\citep{brodermann2024muses} (bottom row).
}
\label{fig:examples_acdc_muses}
\end{figure}

In previous domain-adaptive panoptic segmentation work,  
Cityscapes$\rightarrow$Foggy Cityscapes has been the only benchmark used for 
clear-to-adverse adaptation. Foggy Cityscapes is generated by synthetically 
adding fog to Cityscapes images, which limits the diversity of adverse conditions 
being evaluated. To expand this setting, we introduce two new clear-to-adverse 
benchmarks: Cityscapes$\rightarrow$ACDC and Cityscapes$\rightarrow$MUSES. 
Figure~\ref{fig:examples_acdc_muses} shows example images from 
ACDC~\citep{sakaridis21iccv} and MUSES~\citep{brodermann2024muses},  
while additional details are provided in 
subsection~\ref{subsec:implementation_details}.

\begin{table}[b!]
\centering
\caption{ 
Ablation study on Cityscapes$\rightarrow$ACDC and Cityscapes$\rightarrow$MUSES the self-supervised DINOv2 initialization. 
Top row corresponds to supervised training on Cityscapes. 
We report mean$_{\pm\text{std}}$ over three random seeds.}
\label{tab:main_ablation_dinov2_clear_adverse}
\footnotesize
\setlength{\tabcolsep}{3.5pt}
\begin{tabular}{cccc@{\hskip1.5pt}cc@{\hskip1.5pt}c@{\hskip1.5pt}c}
\toprule
& & & \multicolumn{2}{c}{\text{City$\rightarrow$ACDC}} & & \multicolumn{2}{c}{\text{City$\rightarrow$MUSES}} \\
\cmidrule(lr){4-5} \cmidrule(lr){7-8}
BC & $\text{MLS}_{.99}$  & $\text{CBPF}_{0.8}$ 
&  \multicolumn{2}{c}{$\textbf{PQ}_{19}$} 
& &  \multicolumn{2}{c}{$\textbf{PQ}_{19}$} \\
\midrule
\no  & \no & \no  &  48.7 & \blarrow 
     & &  42.0 & \blarrow \\

\yes & \no & \no  & \mstd{43.3}{1.2} & \textcolor{Degraded}{\text{-5.5}} 
     & &  \mstd{49.3}{0.5} & \textcolor{Improved}{\text{+7.3}} \\

\yes & \yes & \no  & \mstd{53.5}{1.1} & \textcolor{Improved}{\text{+4.8}} 
     & &  \mstd{50.6}{0.7} & \textcolor{Improved}{\text{+8.7}} \\

\yes & \yes & \yes & \mstd{55.5}{1.0} & \textcolor{Improved}{\textbf{+6.8}} 
     & &  \mstd{52.6}{0.2} & \textcolor{Improved}{\textbf{+10.6}} \\
\bottomrule
\end{tabular}
\end{table}

We further ablate MC-PanDA with DINOv2-B on the two new clear-to-adverse benchmarks and present results in Tab.~\ref{tab:main_ablation_dinov2_clear_adverse}. 
The consistency baseline (BC) performs 5.5 \pqn{} points below the supervised model on ACDC, 
quantitatively confirming the noise amplification effect shown in Fig.~\ref{fig:fig1}. 
In contrast, combining MLS and CBPF yields substantial gains: 
6.8 and 10.6 \pqn{} points over the supervised baseline, and 
12.3 and 3.3 \pqn{} points over consistency baseline (BC) on ACDC and MUSES, respectively.
 
\subsubsection{Simplifying the training pipeline} \label{subsec:streamlining}
After validation of our contributions on multiple benchmarks with the new DINOv2 initialization, 
we consider reducing the conceptual complexity of our method.
As described in Sec.~\ref{sec:method_training},
we hypothesize that DINOv2 is robust enough
to make the preliminary training stabilization phases obsolete.
Table~\ref{tab:removing_burn_in_stage} explores this
by comparing MC-PanDA with DINOv2 using the 
three-stage versus
the simplified single-stage pipeline, both 
evaluated with a constant MLS threshold $\tau_1$. 
The two pipelines achieve comparable performance,
while the single-stage version 
substantially reduces the training complexity.
On the other hand, our preliminary experiments
with the single-stage MC-PanDA and a Swin backbone failed to converge.
These results confirm the stabilizing effect of DINOv2 backbone,
which enables a single-stage training
pipeline, adopted in all subsequent experiments.
% adopt the
% the single-stage training pipeline.
\begin{table}[h!]
    \centering
    \footnotesize
    \caption{Comparison of the original three-stage MC-PanDA pipeline and the simplified single-stage \ours{} pipeline using a constant $\tau_1$.}
    \label{tab:removing_burn_in_stage}
    \setlength{\tabcolsep}{4pt}
    \begin{tabular}{lcccc}
         \toprule
         \multirow{2}{*}{\# stages}& \multicolumn{2}{c}{\text{Synthia$\rightarrow$Vistas}} & \multicolumn{2}{c}{\text{City$\rightarrow$ACDC}} \\
         & $\tau_1=0.9$ & $\tau_1=0.99$ & $\tau_1=0.9$ & $\tau_1=0.99$ \\
         \midrule
         three & \mstd{42.1}{3.3} & \mstd{44.2}{0.1} & \mstd{52.9}{1.2} & \textbf{\mstd{55.5}{1.0}} \\
         \rowcolor{gray!10} single  & \textbf{\mstd{42.3}{1.7}} & \textbf{\mstd{44.4}{1.2}} & \textbf{\mstd{54.4}{1.1}} & \mstd{54.1}{1.5} \\
         \bottomrule
    \end{tabular}
\end{table}

\noindent\textcolor{changesframe}{
To further compare convergence, Table~\ref{tab:single_stage_convergence} reports intermediate
checkpoints on Synthia$\rightarrow$Vistas using DINOv2 initialization and a fixed
$\tau_1=0.99$. The single-stage variant follows a very similar convergence trajectory and
slightly improves the final checkpoint, while removing the separate source-domain
pre-training and fixed-teacher burn-in stages. Since both variants train the same
Mask2Former-style architecture, their per-iteration memory cost is comparable. The main
benefit of the single-stage pipeline is therefore reduced conceptual complexity and fewer
stage-specific hyperparameters under a similar total iteration budget.
}

\begin{table}[h!]
\centering
\begingroup
\color{changesframe}
\renewcommand{\thetable}{N1}
\renewcommand{\theHtable}{N1} % important if using hyperref

\caption{
Convergence comparison on Synthia$\rightarrow$Vistas. Both variants use DINOv2
and a fixed $\tau_1=0.99$, isolating the effect of the training pipeline. We report
target-domain PQ at different training checkpoints, averaged over three random seeds.
}
\label{tab:single_stage_convergence}

\footnotesize
\setlength{\tabcolsep}{10pt}
\vspace{0.5em}
\begin{tabular}{lcc}
\toprule
\multirow{2}{*}{Checkpoint} 
& \multirow{2}{*}{\makecell[c]{MC-PanDA \\ three-stage}} 
& \multirow{2}{*}{\makecell[c]{MC-PanDA \\ single-stage}} \\
&& \\
\midrule
50k  & 40.8 & \textbf{41.5} \\
70k  & 42.8 & \textbf{43.0} \\
90k  & 43.7 & \textbf{43.9} \\
110k & 44.2 & \textbf{44.4} \\
\bottomrule
\end{tabular}
\endgroup
\end{table}
\addtocounter{table}{-1}

\subsubsection{Per-class adaptive mask-wide loss scaling} \label{subsec:adaptive_mask_confidences}
\begin{table*}[t!]
	\centering
	\caption{
		Comparison of constant and per-class adaptive $\tau_1$ for different starting values on 
		Synthia$\rightarrow$Vistas (top) and Cityscapes$\rightarrow$ACDC (bottom), using single-stage 
		MC-PanDA with DINOv2-Base. For each starting value, we report mean$_{\pm\text{std}}$ panoptic quality (PQ) over three 
		random seeds. The last column shows the average across all starting values.}
	\label{tab:constant_vs_adaptive_synthia_vistas}
	\setlength{\tabcolsep}{6pt}
	\footnotesize
	\begin{tabular}{lc c c c c c cc@{\hskip 4.7pt}c}
		\toprule
		&\multicolumn{1}{c}{$\tau_1=0.0$}
		&\multicolumn{1}{c}{$\tau_1=0.5$}
		&\multicolumn{1}{c}{$\tau_1=0.6$}
		&\multicolumn{1}{c}{$\tau_1=0.7$}
		&\multicolumn{1}{c}{$\tau_1=0.8$}
		&\multicolumn{1}{c}{$\tau_1=0.9$}
		&\multicolumn{1}{c}{$\tau_1=0.99$} 
		&\multicolumn{2}{c}{\textbf{Avg}} \\
		\midrule
		&\multicolumn{9}{c}{Synthia$\rightarrow$Vistas} \\
		\cmidrule{2-10}
		constant 
		& \no  
		& \mstd{40.3}{2.8}  
		& \mstd{42.2}{1.6} 
		& \mstd{42.4}{4.3} 
		& \mstd{37.2}{2.6} 
		& \mstd{42.3}{1.7} 
		& \mstd{44.4}{1.2} 
		& \mstd{41.5}{2.4} &\blarrow \\
		\rowcolor{gray!10} per-class adaptive 
		&\mstd{44.8}{0.1}
		& \mstd{43.6}{\textbf{0.4}}  
		& \mstd{43.8}{\textbf{0.8}} 
		& \mstd{44.6}{\textbf{1.0}} 
		& \mstd{44.7}{\textbf{1.2}} 
		& \mstd{44.2}{\textbf{0.6}} 
		& \mstd{44.5}{\textbf{0.2}}  
		& \mstd{44.3}{\textbf{0.4}} &\textcolor{Improved}{\text{+2.8}} \\
		%& \mstd{44.2}{\textbf{0.4}} &\textcolor{Improved}{\text{+2.7}} \\
		\midrule
		&\multicolumn{9}{c}{Cityscapes$\rightarrow$ACDC} \\
		\cmidrule{2-10}
		%\midrule
		constant 
		& \no        
		& \mstd{53.6}{2.6} 
		& \mstd{53.5}{2.0}
		& \mstd{53.4}{\textbf{0.5}} 
		& \mstd{51.7}{3.1} 
		& \mstd{54.4}{1.1} 
		& \mstd{54.1}{1.5}  
		& \mstd{53.5}{0.9} &\blarrow \\
		\rowcolor{gray!10} per-class adaptive 
		& \mstd{56.3}{0.6}  
		& \mstd{56.6}{\textbf{1.0}}  
		& \mstd{56.6}{\textbf{0.6}} 
		& \mstd{56.5}{\text{1.1}} 
		& \mstd{56.2}{\textbf{0.3}} 
		& \mstd{56.1}{\textbf{0.7}} 
		& \mstd{57.0}{\textbf{0.6}}  
		& \mstd{56.5}{\textbf{0.3}} &\textcolor{Improved}{\text{+3.0}} \\
		\bottomrule
	\end{tabular}
\end{table*}
\begin{figure*}[h!]
	\centering
	\includegraphics[width=\linewidth,trim=0 126 0 106, clip]{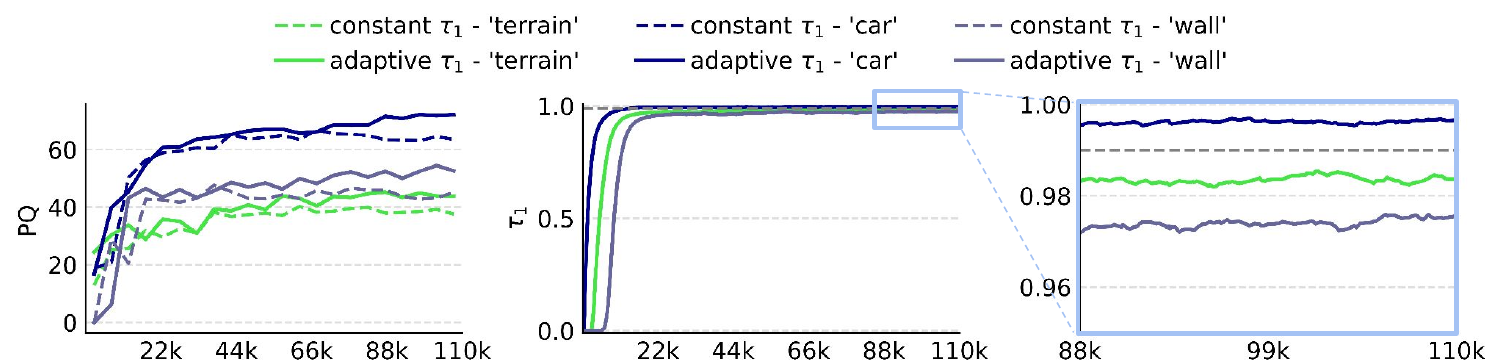}
	\caption{
		Per-class PQ performance (left) and evolution of the adaptive confidence threshold $\tau_1$ during training, alongside the constant baseline $\tau_1 = 0.99$ (middle), for Cityscapes$\rightarrow$ACDC. 
		Per-class adaptive $\tau_1$ can accelerate learning for rare classes such as \textcolor{black}{\texttt{terrain}} and \textcolor{black}{\texttt{wall}}, while also enabling additional late-stage improvements for frequent classes such as \textcolor{black}{\texttt{car}}.}
	\label{fig:per_class_adaptive_tau1_gains}
\end{figure*}

This section presents a detailed analysis of our main methodological contribution to 
the preliminary MC-PanDA: introducing per-class and adaptive mask-wide loss scaling, as 
described in~\ref{subsec:per_class_adaptive}. 
Intuitively, model confidence is not uniform across classes due to factors such as 
class imbalance, varying frequency, and differences in visual complexity. 
Therefore, requiring $\tau_1$ to be per-class constitutes a natural improvement over 
the global, constant threshold used in the preliminary MC-PanDA. 
However, selecting an optimal $\tau_1$ 
separately for each 
class is non-trivial and 
quickly becomes intractable. 
To address this,
we make $\tau_1$ both, per-class and self-adapting,
allowing it to be 
automatically updated
during training and
independent of its initial value.

\bmhead{On the robustness of $\tau_1$}  
We first evaluate the
robustness of the
mask-confidence threshold 
by comparing 
single-stage MC-PanDA with 
a constant $\tau_1$
against our new per-class adaptive 
$\tau_1$ (Table~\ref{tab:constant_vs_adaptive_synthia_vistas}) on two benchmarks: 
Synthia$\rightarrow$Vistas and Cityscapes$\rightarrow$ACDC. 
In this experiment, we vary the
initial value of $\tau_1$ and measure the final 
target-domain performance.

Two main weaknesses of a constant $\tau_1$ emerge:  
(i) performance fluctuates considerably as $\tau_1$ changes, and  
(ii) for a fixed $\tau_1$, the standard deviation across runs is large, reaching up to 4~PQ points.  
These instabilities highlight the sensitivity of MC-PanDA to manual threshold selection.
In contrast, the per-class adaptive $\tau_1$ exhibits:  
(i) stable performance across different starting values, and  
(ii) substantially reduced variance across runs.  
Beyond its stability benefits, the adaptive formulation also improves
downstream performance by 2.8~\pqs{} on Synthia$\rightarrow$Vistas and 
3.0~\pqn{} on Cityscapes$\rightarrow$ACDC, on average.  
Thus, per-class self-adapting thresholding not
only removes the need to tune $\tau_1$ but can also provide 
further performance gains. 
\textcolor{changesframe}{
Additional experiments on robustness of per-class adaptive $\tau_1$ can be found in Table~\ref{tab:tau_2_robustness}, and in an 
UrbanSyn$\rightarrow$Vistas analysis reported in Appendix~\ref{app:tau1_robustness}.
}

Figure~\ref{fig:per_class_adaptive_tau1_gains} illustrates the effect of using a 
per-class adaptive $\tau_1$ (starting value equal 0.0) compared to a fixed $\tau_1 = 0.99$ (as in the original 
MC-PanDA) when training the single-stage model. 
The left panel shows the panoptic quality evolution for three classes: \texttt{terrain}, \texttt{wall} and \texttt{car}. Adaptive thresholding accelerates learning for rare classes such as \texttt{terrain} 
and \texttt{wall}, and also provides additional late-stage improvements for frequent \texttt{car} class. 
The middle panel depicts the evolution of adaptive $\tau_1$ over training, accompanied by the 
constant baseline, while the right panel provides a zoomed-in view. 
Overall, while a fixed $\tau_1 = 0.99$ was a reasonable choice in MC-PanDA, making 
$\tau_1$ both per-class and self-adapting can yield further stability and performance gains.
\textcolor{changesframe}{
We provide an additional rare/frequent-class analysis in Appendix~\ref{app:rare_frequent_tau1},
where adaptive thresholding improves the average per-class PQ of the five rarest classes by
3.1 points on Cityscapes$\rightarrow$ACDC.
}

\bmhead{On the choice of aggregation function $f$} 
For the instance-wise aggregation function $f$ (cf.\ Eq.~\ref{eq:tau_update}), we use 
the \textit{third quartile} by default. 
Table~\ref{tab:different_aggregation_functions} further shows that MLS remains robust 
when employing alternative aggregation functions, such as the \textit{mean} or 
\textit{median}.

\begin{table}[h!]
    \centering
    \footnotesize
\caption{Different aggregation functions $f$ for per-class adaptive 
$\tau_1$. We report mean$_{\pm\text{std}}$ panoptic quality (PQ) over three 
random seeds.}
    \label{tab:different_aggregation_functions}
    \begin{tabular}{l@{\quad}cc}
         \toprule
         \multirow{1}{*}{$f$}& \multicolumn{1}{c}{\text{Synthia$\rightarrow$Vistas}} & \multicolumn{1}{c}{\text{City$\rightarrow$ACDC}} \\
         \midrule
         \textit{mean}          & \mstd{44.4}{0.3} & \mstd{54.3}{1.7}  \\
         \textit{median}        & \mstd{44.8}{1.3} & \mstd{55.2}{0.7}  \\
         \rowcolor{gray!10} \textit{3. quartile}   & \textbf{\mstd{44.8}{0.1}} & \textbf{\mstd{56.3}{0.6}}  \\
         \bottomrule
    \end{tabular}
\end{table}

\textcolor{changesframe}{
\noindent\textbf{On the choice of cross-instance aggregation.}
Eq.~\ref{eq:tau_update} first aggregates pixel-level confidences within each mask using
$f$, and then aggregates across teacher-predicted masks of the same class through the outer
$\max_{i \in \mathcal{I}_k^n}$ operation. The fixed-threshold analysis in
Table~\ref{tab:constant_vs_adaptive_synthia_vistas} suggests that MLS benefits from selective thresholding:
lower values of $\tau_1$ make the loss scaling more permissive and lead to less stable
performance, while high thresholds perform best. This motivates the max-over-instances update,
which anchors the adaptive threshold of class $k$ to the most confident teacher masks of that
class, rather than allowing many low-confidence pseudo-masks to lower the threshold. To
validate this design choice, we compare the default maximum with mean and third-quartile
aggregation across instances of the same class, while keeping the instance-wise aggregation
$f$ fixed to the third quartile (Table~\ref{tab:cross_instance_aggregation}). The maximum performs best or tied across all three benchmarks, while mean aggregation is
consistently worse. This supports max-over-instances as a simple and effective default for the
adaptive MLS threshold update.
}
\begin{table}[h!]
    \centering
    \footnotesize
    \begingroup
    \color{changesframe}
    \renewcommand{\thetable}{N2}
    \renewcommand{\theHtable}{N2}
    \caption{
    Effect of the cross-instance aggregation over teacher-predicted masks of the same class in
    Eq.~(\ref{eq:tau_update}). We keep the instance-wise aggregation $f$ fixed to the third
    quartile and vary the outer aggregation over instances. We report mean$\pm$std PQ over
    three random seeds.
    }
    \label{tab:cross_instance_aggregation}
    \setlength{\tabcolsep}{3pt}
    \begin{tabular}{lccc}
         \toprule
         \multirow{2}{*}{\makecell{Outer \\ aggregation}}
         & \multirow{2}{*}{Syn$\rightarrow$Vistas}
         & \multirow{2}{*}{City$\rightarrow$ACDC}
         & \multirow{2}{*}{City$\rightarrow$MUSES} \\
         &&& \\
         \midrule
         \textit{mean}
         & \mstd{43.5}{1.7}
         & \mstd{54.6}{0.1}
         & \mstd{52.0}{0.6} \\
         \textit{3. quartile}
         & \mstd{43.9}{0.0}
         & \mstd{56.3}{0.7}
         & \mstd{52.0}{1.2} \\
         \rowcolor{gray!10}
         \textit{max}
         & \textbf{\mstd{44.8}{0.1}}
         & \textbf{\mstd{56.3}{0.6}}
         & \textbf{\mstd{52.4}{0.4}} \\
         \bottomrule
    \end{tabular}
    \endgroup
\end{table}
\addtocounter{table}{-1}

\bmhead{Global vs.\ per-class adaptation}
As an alternative to the per-class formulation in Eq.~\ref{eq:tau_update}, 
$\tau_1$ can also be adapted globally, i.e., using a single threshold shared across 
all classes. In this case, a single update term $\delta^n$ is computed over all instances:
\begin{equation}
\label{eq:tau_update_global}
    \delta^n
    =
    \max_{i \in \mathcal{I}^n}
    \; f\!\left(
        \big\{\, \rho_{i,r,c} \;\big|\; (r,c)\in M_i^{F} \,\big\}
      \right),
\end{equation}
where $\mathcal{I}^n$ denotes the set of all teacher-predicted instances in the batch at the training iteration $n$. 
This is different from the per-class formulation, which computes the maximum only over instances of class $k$ (i.e., $i \in \mathcal{I}_k^n$). Table~\ref{tab:global_vs_per_class_adaptive_tau1} compares the global-adaptive and 
per-class adaptive formulations  
on the Cityscapes$\rightarrow$ACDC benchmark
for several initial $\tau_1$ values. 
Per-class adaptation consistently yields higher 
performance, confirming it as a preferred choice for \ours{}.

\begin{table}[h!]
    \centering
    \caption{Comparison of global-adaptive and per-class adaptive $\tau_1$ on 
Cityscapes$\rightarrow$ACDC for different initial threshold values. We report mean$_{\pm\text{std}}$ panoptic quality (PQ) over three 
random seeds.}
    \label{tab:global_vs_per_class_adaptive_tau1}
    \footnotesize
    \begin{tabular}{lccc}
         \toprule
         \multirow{2}{*}{$\tau_1$}&  \multicolumn{3}{c}{\text{Cityscapes$\rightarrow$ACDC~(\tiny\pqn{})}} \\
         & $\tau_1=0.0$ & $\tau_1=0.5$ & $\tau_1=0.99$ \\
         \midrule
         global adaptive     & \mstd{56.1}{0.7} & \mstd{55.3}{0.5}& \mstd{55.4}{1.0}  \\
         \rowcolor{gray!10} per-class adaptive  &\mstd{56.3}{0.6} & \mstd{56.6}{1.0} &\mstd{57.0}{0.6}  \\
         \bottomrule
    \end{tabular}
\end{table}
\vspace{-1.6em}

\subsubsection{Robustness of confidence-based point filtering} \label{subsec:tau2_robustness}
Table~\ref{tab:tau_2_robustness} evaluates the robustness of confidence-based point 
filtering (CBPF) with respect to the threshold $\tau_2$ across three benchmarks: 
Synthia$\rightarrow$Vistas and Cityscapes$\rightarrow$\{ACDC, MUSES\}. 
For each setting, we vary the initial value of the per-class adaptive $\tau_1$ and 
report mean$_{\pm\text{std}}$ over three random seeds. 
Across all benchmarks and $\tau_1$ initializations, performance remains remarkably 
stable for $\tau_2 \in \{0.7,0.8,0.9\}$, with variations typically below 1 PQ pp. 
This indicates that CBPF is mostly insensitive to the precise choice of $\tau_2$, 
and that $\tau_2 = 0.8$, default value used in MC-PanDA, continues to be 
a reliable choice in \ours{}. Moreover, given the small differences observed across values, tuning $\tau_2$ offers negligible additional benefit. 

\begin{table}[h!]
\centering
\caption{
Robustness of confidence-based point filtering (CBPF) with respect to the threshold 
$\tau_2$ on: Synthia$\rightarrow$Vistas, 
Cityscapes$\rightarrow$ACDC, and Cityscapes$\rightarrow$MUSES, 
evaluated across different initial values of the per-class adaptive $\tau_1$.
$\dagger$ marks the default setting used in \ours{}, $\tau_2 = 0.8$.
}
\label{tab:tau_2_robustness}
\footnotesize
\setlength{\tabcolsep}{2.5pt}
\begin{tabular}{lccccccc@{\hskip10pt}c}
\toprule
& \multicolumn{7}{c}{\text{Synthia$\rightarrow$Vistas~(\tiny\pqs{})}} & \multirow{2}{*}{\textbf{Avg}} \\
\cmidrule{2-8}
\multicolumn{1}{r}{$\tau_1{=}$} & 0.0 & 0.5 & 0.6 & 0.7 & 0.8 & 0.9 & 0.99 & \\
\midrule
$\tau_2{=}0.7$& 45.2 & 44.5 & 44.5 & 45.1 & 45.3 & 44.8 & 45.2 & \mstd{44.9}{0.3}\\
\rowcolor{gray!10} $\tau_2{=}0.8^\dagger$& 44.8 & 43.6 & 43.8 & 44.6 & 44.7 & 44.2 & 44.5 & \mstd{44.3}{0.4}\\
$\tau_2{=}0.9$ & 44.4 & 45.2 & 45.0 & 44.5 & 44.8 & 44.8 & 43.8 & \mstd{44.6}{0.4}\\
\midrule
\midrule
& \multicolumn{7}{c}{\text{Cityscapes$\rightarrow$ACDC~(\tiny\pqn{})}} & \multirow{2}{*}{\textbf{Avg}} \\
\cmidrule{2-8}
\multicolumn{1}{r}{$\tau_1{=}$} & 0.0 & 0.5 & 0.6 & 0.7 & 0.8 & 0.9 & 0.99 & \\
\midrule
$\tau_2{=}0.7$& 56.3 & 56.3& 56.7 & 55.4 & 55.7 & 56.6 &56.3 & \mstd{56.2}{0.4} \\
\rowcolor{gray!10} $\tau_2{=}0.8^\dagger$& 56.3 & 56.6& 56.6& 56.6& 56.2& 56.1& 57.0& \mstd{56.5}{0.3} \\
$\tau_2{=}0.9$ & 57.0 & 56.2& 56.7& 56.6& 56.7& 57.6& 56.8 & \mstd{56.8}{0.4} \\
\midrule
\midrule
& \multicolumn{7}{c}{\text{Cityscapes$\rightarrow$MUSES~(\tiny\pqn{})}} & \multirow{2}{*}{\textbf{Avg}} \\
\cmidrule{2-8}
\multicolumn{1}{r}{$\tau_1{=}$} & 0.0 & 0.5 & 0.6 & 0.7 & 0.8 & 0.9 & 0.99 & \\
\midrule
$\tau_2{=}0.7$ & 51.9 &  51.8&52.2 &52.4 & 52.0 & 51.5 & 52.2 & \mstd{52.0}{0.3} \\
\rowcolor{gray!10} $\tau_2{=}0.8^\dagger$& 52.4 & 52.1& 52.1& 52.4& 52.7& 52.0 & 52.6&   \mstd{52.2}{0.3} \\
$\tau_2{=}0.9$ & 52.8& 53.1& 52.8& 52.6& 52.1& 52.7& 52.9&    \mstd{52.7}{0.3} \\
\bottomrule
\end{tabular}
\end{table}
%\vspace{-1.6em}

\subsection{MC-PanDA\texttt{++}} \label{subsec:mcpandapp}
\begin{table*}[t!]
	\centering
	\footnotesize
	\caption{
		Overview of the final \ours{} results across six benchmarks. 
		Improvements over the source-only fine-tuning baseline are shown for reference. 
		All \ours{} results are reported as mean$_{\pm\text{std}}$ over three random seeds.
	}
	
	\label{tab:final_mcpandapp_results}
	\setlength{\tabcolsep}{4.2pt}
	\begin{tabular}{l l clcl c clcl c clcl}
		\toprule
		\multirow{2}{*}{Method} & \multirow{2}{*}{Backbone} & \multicolumn{4}{c}{Synthia~(\tiny \pqs{})} & & \multicolumn{4}{c}{UrbanSyn~(\tiny \pqn{})} & & \multicolumn{4}{c}{Cityscapes~(\tiny \pqn{})} \\
		\cmidrule{3-6} \cmidrule{8-11} \cmidrule{13-16}
		& &
		\multicolumn{2}{c}{\text{Cityscapes}} &
		\multicolumn{2}{c}{\text{Vistas}} & &
		\multicolumn{2}{c}{\text{Cityscapes}} &
		\multicolumn{2}{c}{\text{Vistas}} & &
		\multicolumn{2}{c}{\text{ACDC}} &
		\multicolumn{2}{c}{\text{MUSES}} \\
		\midrule
		Source-only & DINOv2-Base  &
		\text{36.0} & \blarrow &
		\text{33.7} & \blarrow & &
		\text{48.3} & \blarrow &
		\text{42.1} & \blarrow & &
		\text{48.7} & \blarrow &
		\text{42.0} & \blarrow \\
		\rowcolor{gray!10} \ours{} & DINOv2-Base &
		\text{49.6} & \textcolor{Improved}{\textbf{+13.6}} &
		\text{44.8} & \textcolor{Improved}{\textbf{+11.1}} & &
		\text{57.2} & \textcolor{Improved}{\textbf{+8.9}} &
		\text{49.5} & \textcolor{Improved}{\textbf{+7.4}} & & 
		\text{56.3} & \textcolor{Improved}{\textbf{+7.6}} &
		\text{52.4} & \textcolor{Improved}{\textbf{+10.4}} \\
		\midrule
		Source-only & DINOv2-Large  &
		\text{39.9} & \blarrow &
		\text{37.9} & \blarrow & &
		\text{51.1} & \blarrow &
		\text{45.2} & \blarrow & &
		\text{52.3} & \blarrow & 
		\text{46.4} & \blarrow  \\ 
		\rowcolor{gray!10} \ours{} & DINOv2-Large &
		\text{54.2} & \textcolor{Improved}{\textbf{+14.3}} &
		\text{47.6} & \textcolor{Improved}{\textbf{+9.7}} & &
		\text{59.1} & \textcolor{Improved}{\textbf{+8.0}} &
		\text{52.7} & \textcolor{Improved}{\textbf{+7.5}} & & 
		\text{60.1} & \textcolor{Improved}{\textbf{+7.8}} &
		\text{55.6} & \textcolor{Improved}{\textbf{+9.2}} \\
		\bottomrule
	\end{tabular}
\end{table*}

All components presented in the previous sections culminate to our final method,
\ours{}. It uses a strong self-supervised DINOv2, adopts a simplified
single-stage training pipeline, and employs per-class adaptive mask-wide loss
scaling. The latter substantially improves stability and removes the sensitivity to
the initial choice of $\tau_1$, which was a key limitation of the preliminary
MC-PanDA~\citep{martinovic2024eccv}. We initialize $\tau_1$ at 0.0 and let it evolve throughout training.

We report the final performance of
\ours{} in Table~\ref{tab:final_mcpandapp_results}, compare it against the
supervised source-only baselines, and evaluate the effect of scaling the encoder
from DINOv2-B to DINOv2-L. The consistent improvements observed with the
larger encoder further confirm the effectiveness of \ours{}.

\section{Qualitative analysis} \label{sec:qualitative}
Figure~\ref{fig:mls_factor_over_training} shows the \textcolor{csroad}{road} and
\textcolor{cstrafficsign}{traffic sign} masks at two training checkpoints together with their
corresponding mask-wide loss-scaling factors $\lambda_i$. A clear positive correlation emerges
between mask quality and the value of $\lambda_i$, indicating that MLS effectively suppresses 
gradients for low-quality masks. This behavior is particularly important during 
early training, when many pixels are incorrectly excluded from the corresponding mask. 
\textcolor{changesframe}{
We further quantify this qualitative observation in Appendix~\ref{app:mls_iou_correlation}
by measuring the Spearman correlation between $\lambda_i$ and the ground-truth IoU of matched
teacher pseudo-masks on Cityscapes$\rightarrow$ACDC. The resulting class-averaged correlation
is positive for both stuff and thing classes, with an average of $\rho=0.61$ over all
19 classes. This confirms that MLS reliably reflects the true pseudo-mask quality across all classes, supporting its role in mitigating confirmation bias.
}

\newcommand{\mywt}{0.48\linewidth}
\begin{figure}[h]
    \centering
    \begin{tabular}{c@{\,}c@{\,}c}
         \footnotesize 20k & 
         \footnotesize 40k \\
         \includegraphics[width=\mywt]{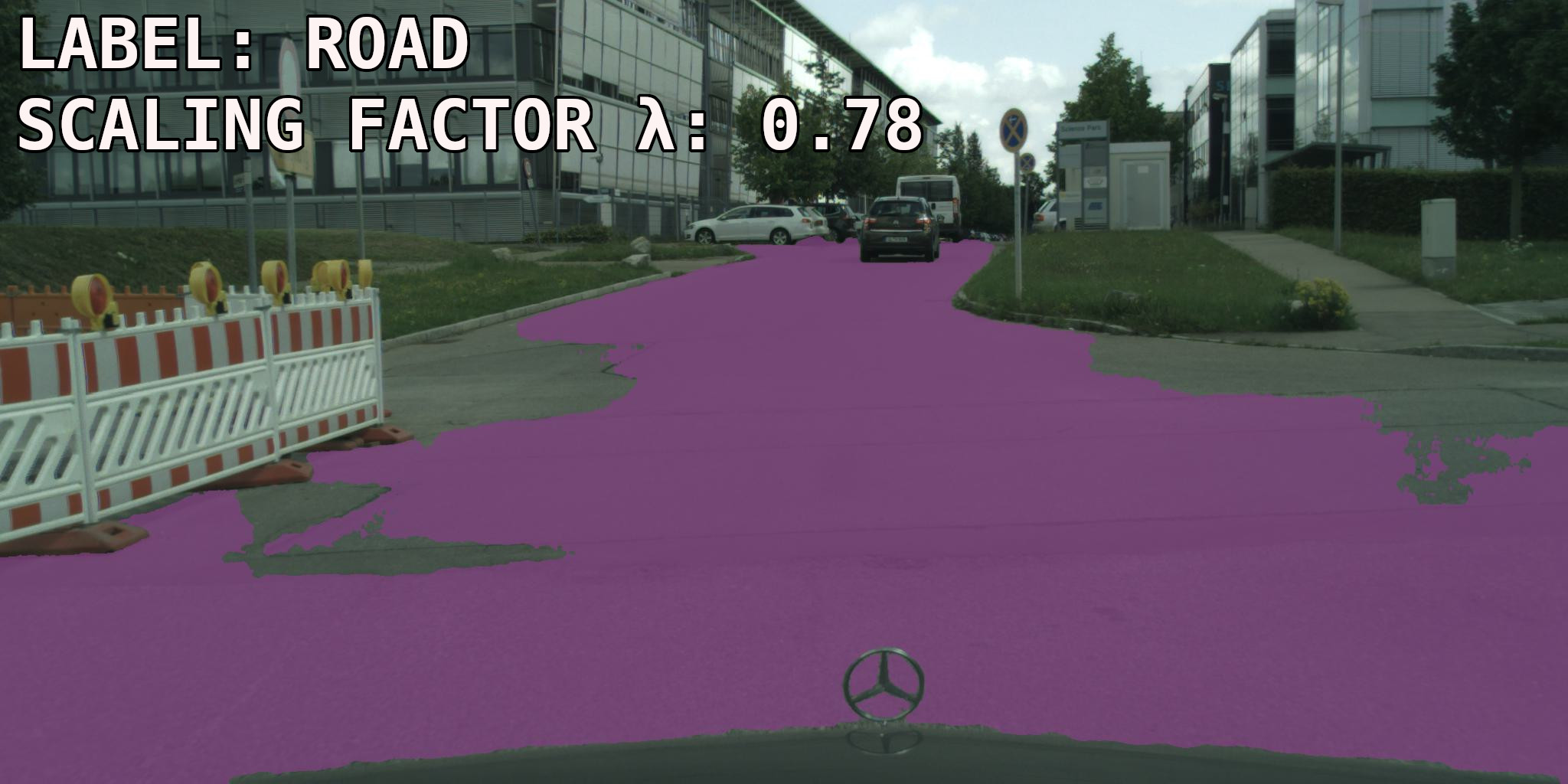} &
         \includegraphics[width=\mywt]{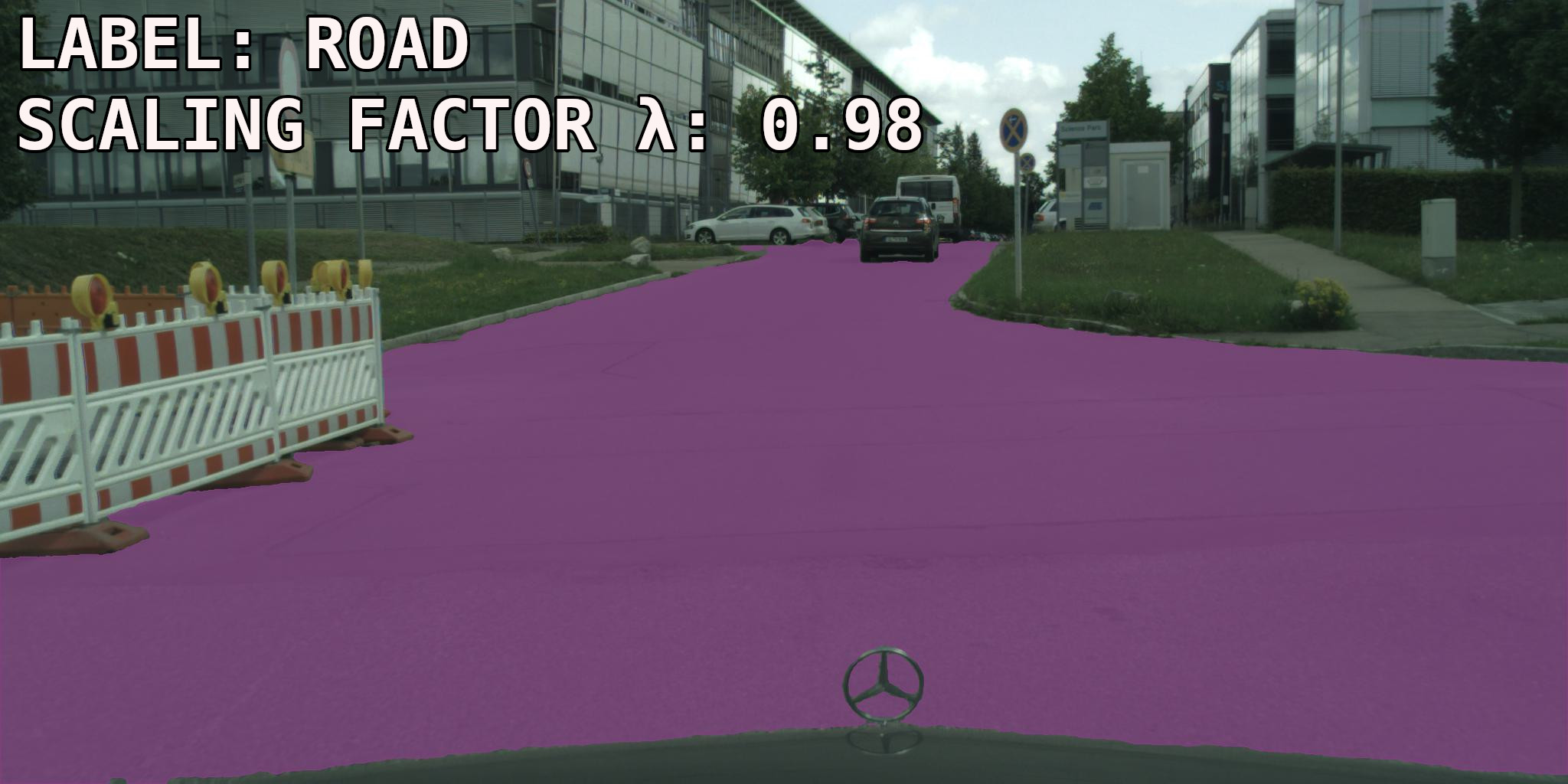}
         \\ 
         \includegraphics[width=\mywt]{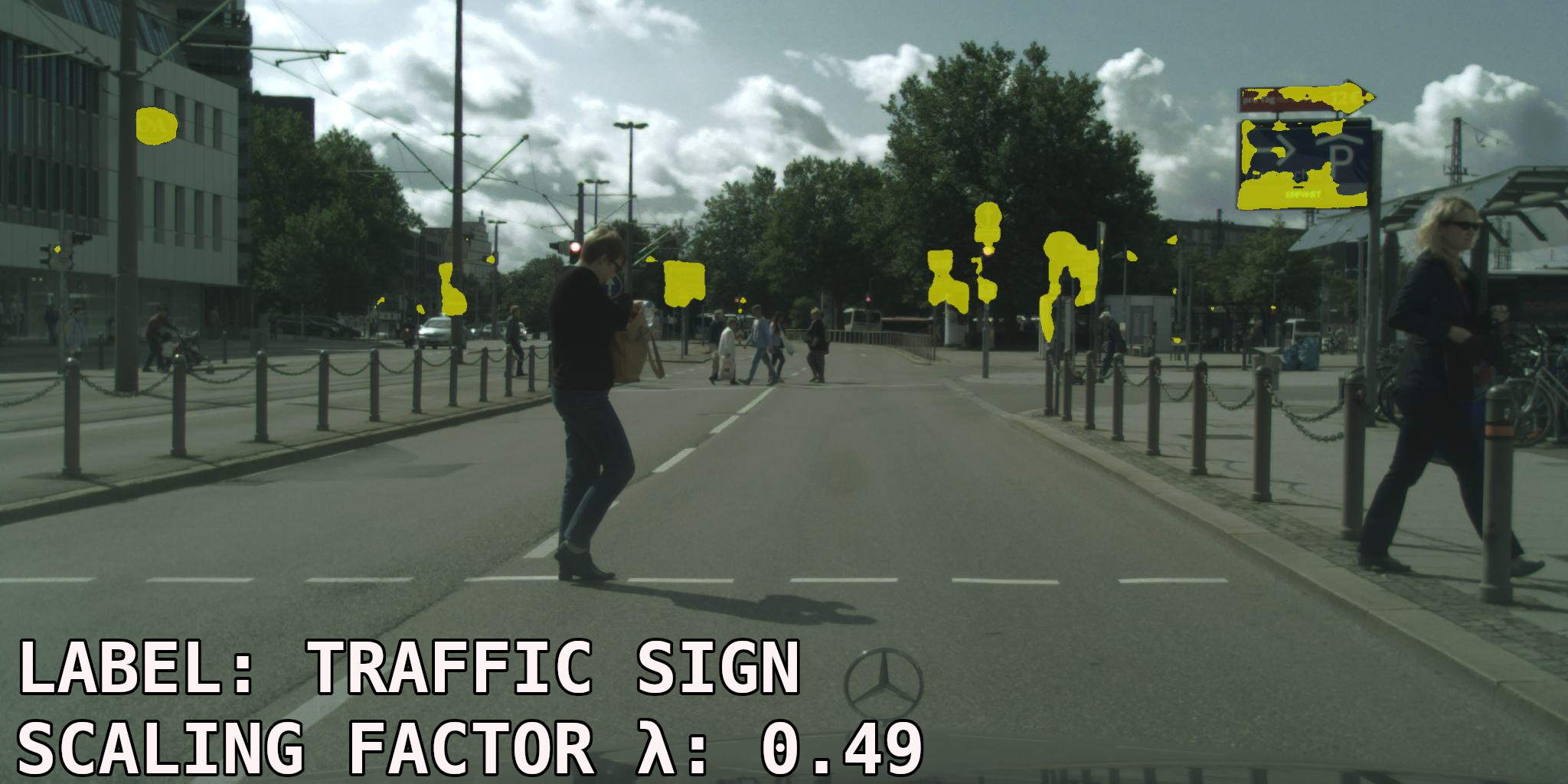} &
         \includegraphics[width=\mywt]{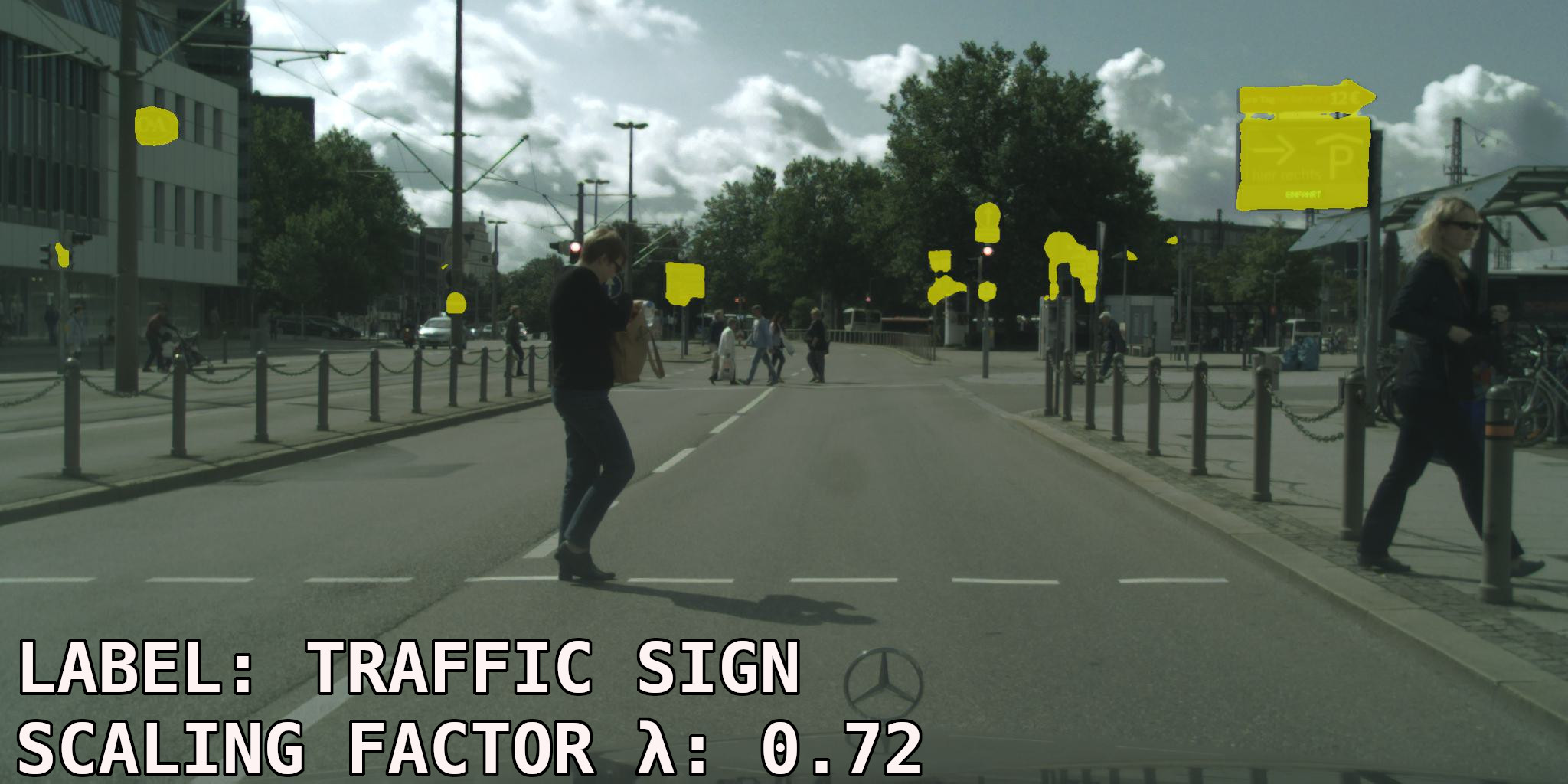} \\ 
    \end{tabular}
   \caption{\textcolor{changesframe}{
Illustration of the MLS factor $\lambda_i$ for two model checkpoints during training,
shown here for masks of classes \textcolor{csroad}{\texttt{road}} and
\textcolor{cstrafficsign}{\texttt{traffic sign}}. As training progresses, the predicted
masks become more complete and visually coherent, and the corresponding $\lambda_i$ values
increase accordingly. A strong correlation is observed between the value of $\lambda_i$ and the visual
quality of the predicted mask.
% as also illustrated in MC-PanDA~\citep{martinovic2024eccv}.
}
}
\label{fig:mls_factor_over_training}
\end{figure}

Figure~\ref{fig:complementary_contributions_mls_cbpf} clearly illustrates how our
mask-wide loss scaling (MLS) and confidence-based point filtering (CBPF)
provide complementary benefits in selecting reliable pixels for domain-adaptive 
learning. Both the \texttt{person} mask (\textcolor{csperson}{left}) and the motorcycle mask (\textcolor{csmotorcycle}{right}) fail to 
capture all of their corresponding pixels, resulting in false-negative regions. 
These regions align closely with areas of low sampling affinities (highlighted in red, 
middle). Together, MLS and CBPF prevent the student
model from learning from these erroneous 
pseudo-labels.

\newcommand{\mywf}{0.315\linewidth}
\begin{figure}[h!]
    \centering
     \begin{tabular}{c@{\,}c@{\,}c}
         \includegraphics[width=\mywf]{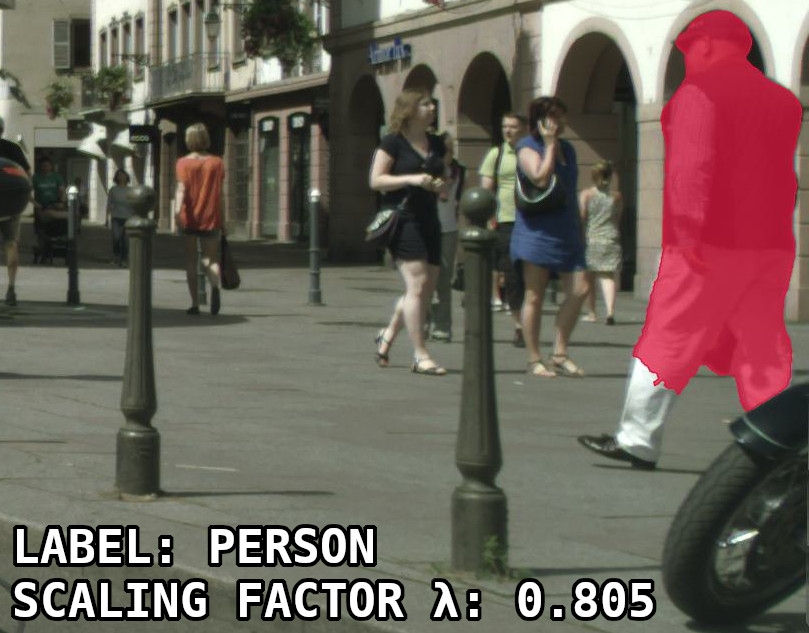} & 
         \includegraphics[width=\mywf]{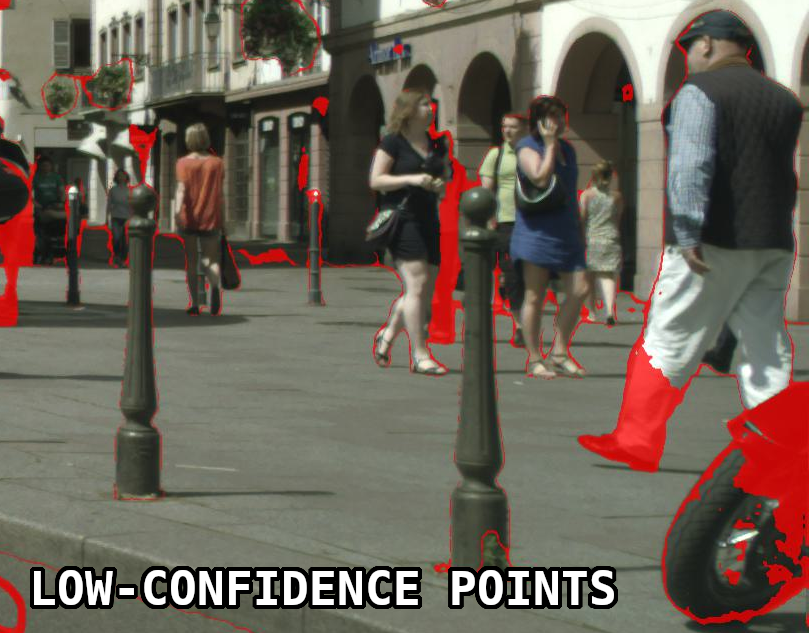} &
         \includegraphics[width=\mywf]{figures_fig11_complementary_motorcyc_conf} \\ 
    \end{tabular}
    \caption{
Visual example illustrating the complementary roles of Mask-wide Loss Scaling (MLS) 
and Confidence-based Point Filtering (CBPF) in selecting reliable pixels for 
domain-adaptive learning, as also illustrated in MC-PanDA~\citep{martinovic2024eccv}. 
The middle image highlights two regions where back-propagation is suppressed due 
to low sampling affinity. 
The left and right images show that these regions correspond to false-negative 
predictions in the \textcolor{csperson}{\texttt{person}} and \textcolor{csmotorcycle}{\texttt{motorcycle}} masks, respectively.
}
\label{fig:complementary_contributions_mls_cbpf}
\end{figure}

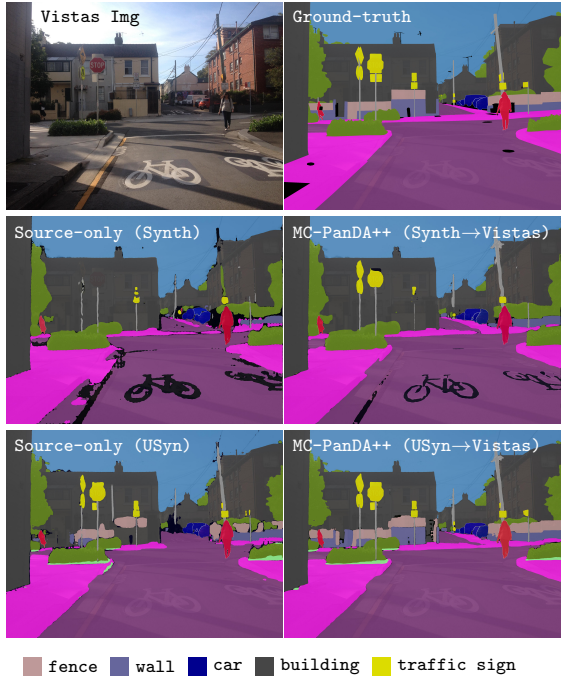
\begin{figure}[h!]
    \centering
    \begin{tabular}{c@{\hskip 1pt}c}

    \begin{tikzpicture}
        \node[inner sep=0] (img) {\includegraphics[width=0.48\linewidth]{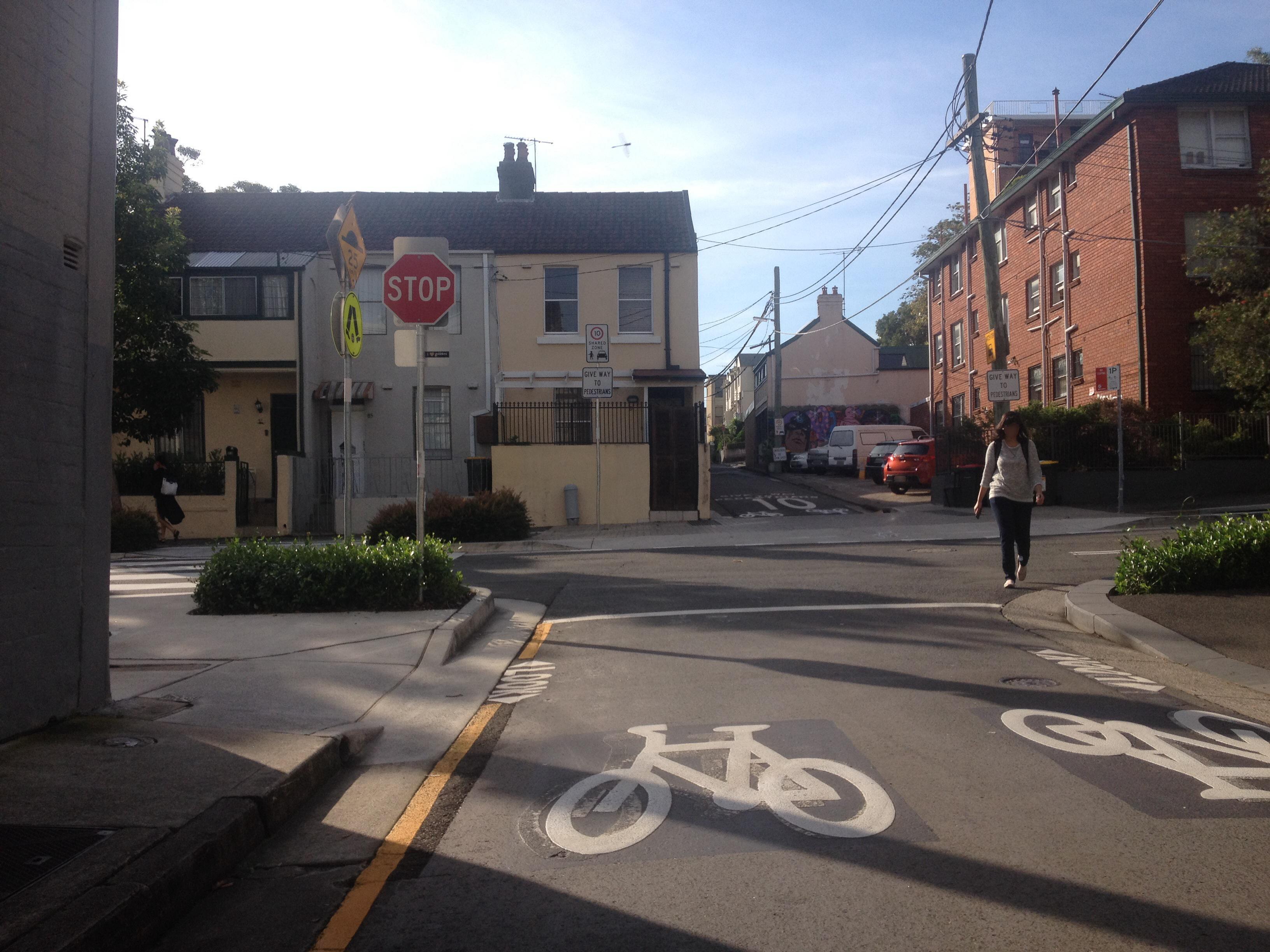}};
        % label
        \node[anchor=north west,
              fill=none,
              text=black,
              rounded corners=2pt,
              inner sep=3pt,
              font=\ttfamily\footnotesize] at ([xshift=10pt]img.north west) {Vistas Img};
    \end{tikzpicture}
    &
    \begin{tikzpicture}
        \node[inner sep=0] (img) {\includegraphics[width=0.48\linewidth]{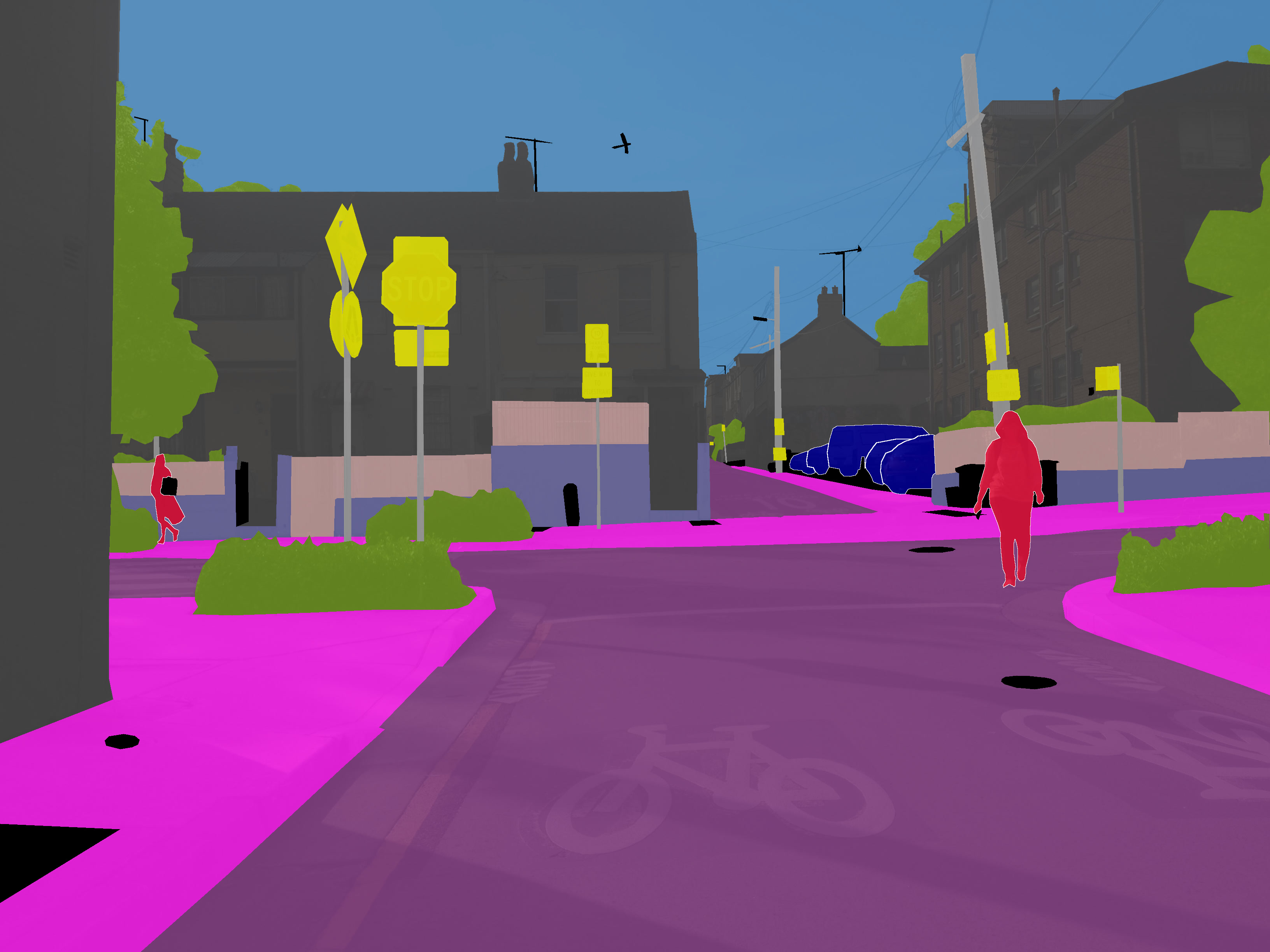}};
        \node[anchor=north west,
              fill=none,
              text=white,
              rounded corners=2pt,
              inner sep=3pt,
              font=\ttfamily\footnotesize] at (img.north west) {Ground-truth};
    \end{tikzpicture}
    \\[-1pt]

    \begin{tikzpicture}
        \node[inner sep=0] (img) {\includegraphics[width=0.48\linewidth]{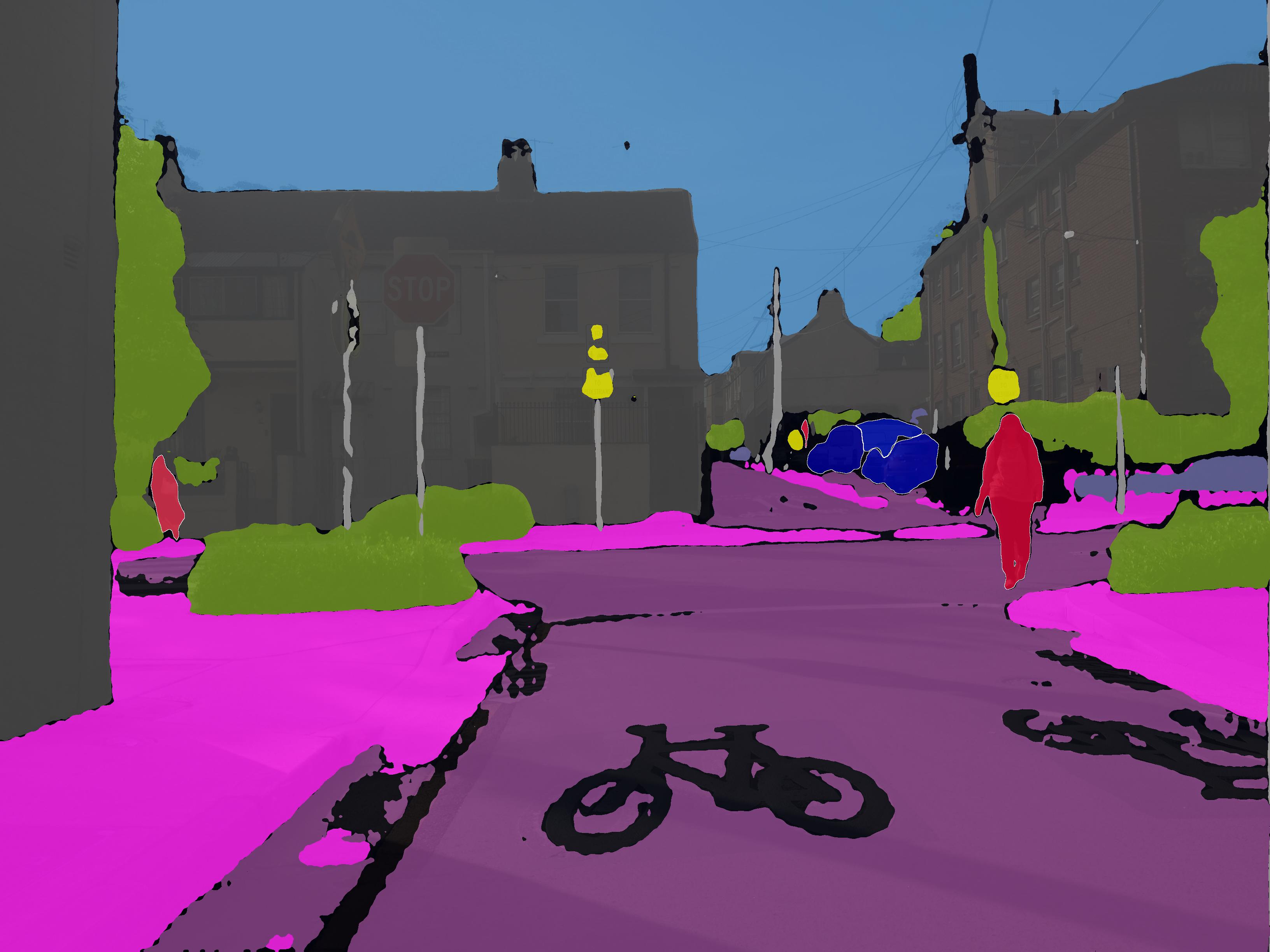}};
        \node[anchor=north west,
              fill=none,
              text=white,
              rounded corners=2pt,
              inner sep=3pt,
              font=\ttfamily\footnotesize] at (img.north west) {Source-only (Synth)};
    \end{tikzpicture}
    &
    \begin{tikzpicture}
        \node[inner sep=0] (img) {\includegraphics[width=0.48\linewidth]{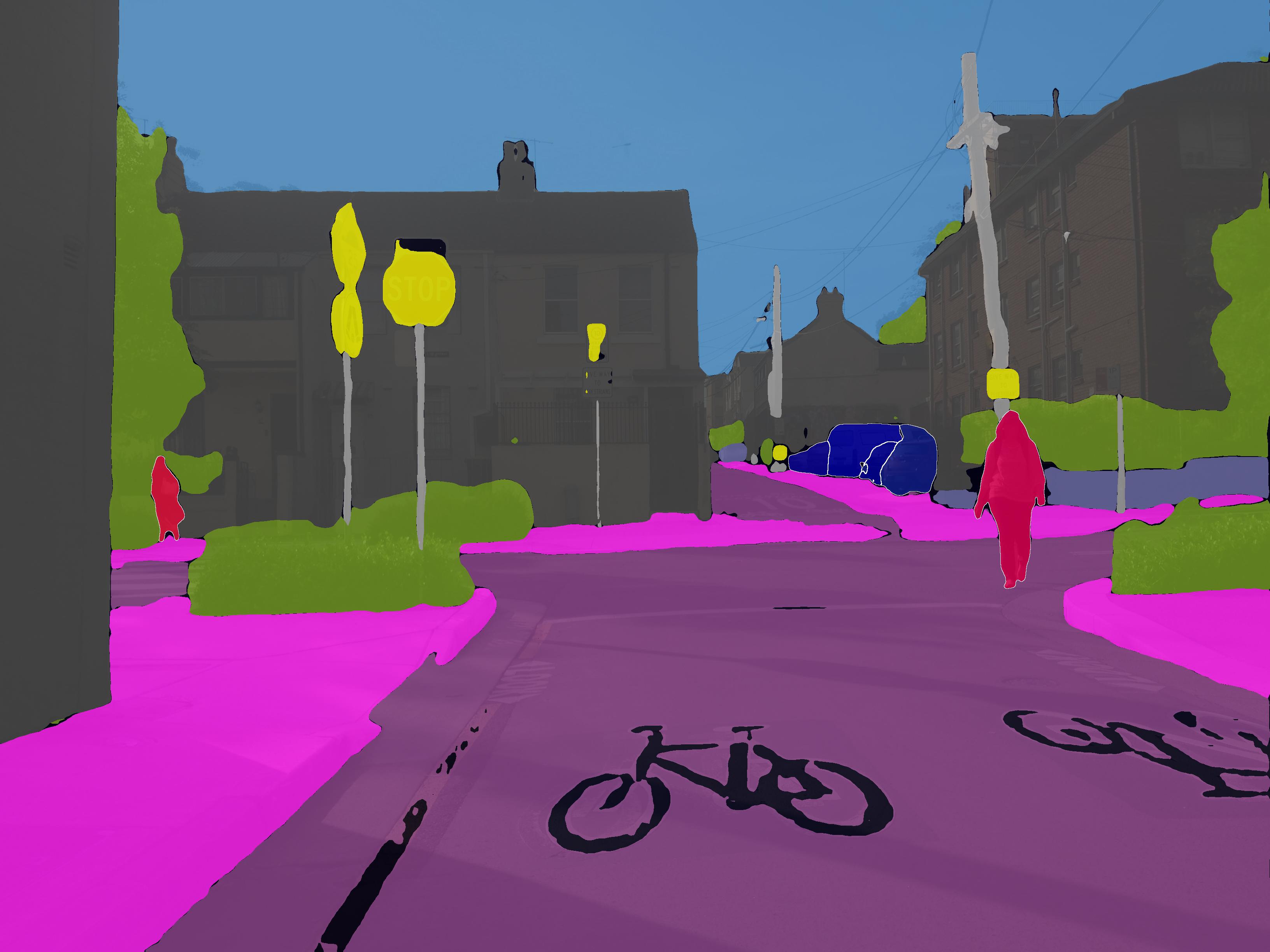}};
        \node[anchor=north west,
              fill=none,
              text=white,
              rounded corners=2pt,
              inner sep=3pt,
              font=\ttfamily\footnotesize] at (img.north west) {\ours{} (Synth$\rightarrow$Vistas)};
    \end{tikzpicture}

    \\[-1pt]

   \begin{tikzpicture}
        \node[inner sep=0] (img) {\includegraphics[width=0.48\linewidth]{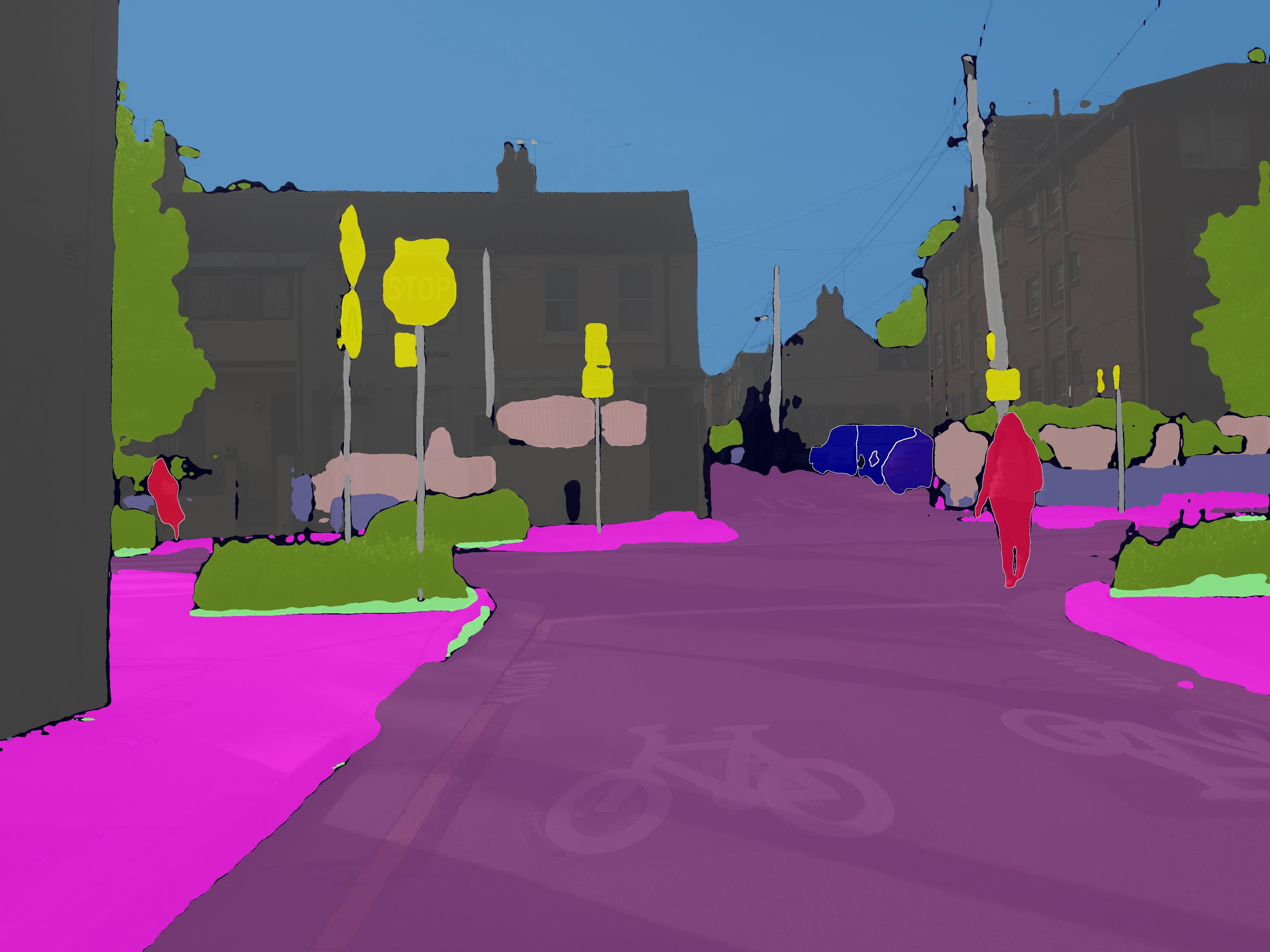}};
        \node[anchor=north west,
              fill=none,
              text=white,
              rounded corners=2pt,
              inner sep=3pt,
              font=\ttfamily\footnotesize] at (img.north west) {Source-only (USyn)};
    \end{tikzpicture}
    &
    \begin{tikzpicture}
        \node[inner sep=0] (img) {\includegraphics[width=0.48\linewidth]{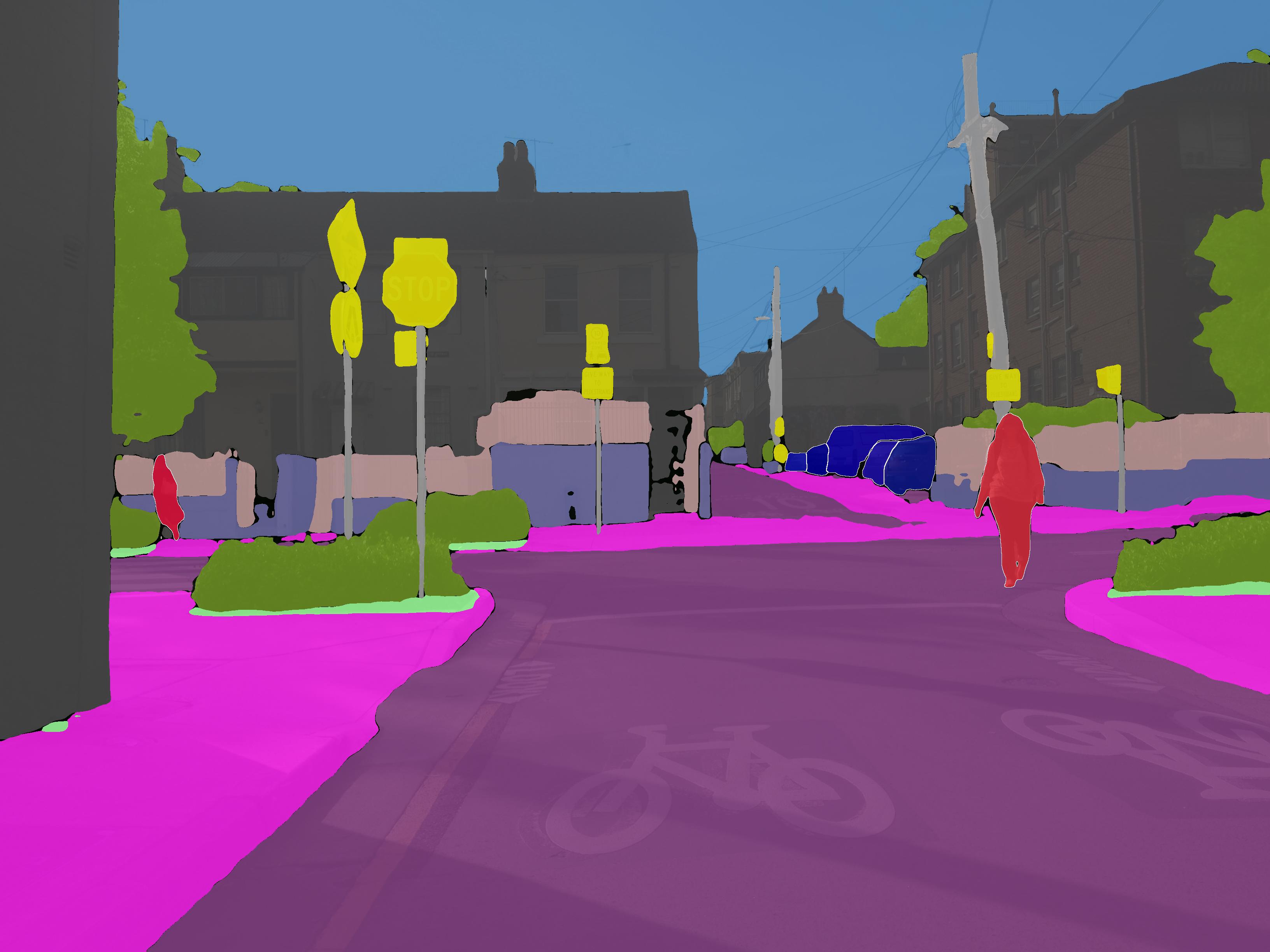}};
        \node[anchor=north west,
              fill=none,
              text=white,
              rounded corners=2pt,
              inner sep=3pt,
              font=\ttfamily\footnotesize] at (img.north west) {\ours{} (USyn$\rightarrow$Vistas)};
    \end{tikzpicture}

    \end{tabular}
    \cityscapeslegendsynth
    \vspace{4pt}
    \caption{
Predictions comparison on a Vistas image between source-only models trained on 
Synthia (middle left) and UrbanSyn (bottom left), and \ours{} (right). 
Switching the synthetic source from Synthia to UrbanSyn already yields more 
coherent predictions, even without adaptation. 
\ours{} further improves segmentation, particularly for distant 
\textcolor{cscar}{\texttt{cars}}, \textcolor{cswall}{\texttt{wall}}/ 
\textcolor{csfence}{\texttt{fence}} regions, and small 
\textcolor{cstrafficsign}{\texttt{traffic signs}}. Best viewed zoomed in.
}
    \label{fig:visualization_synthetic_real}
\end{figure}

Figure~\ref{fig:visualization_synthetic_real} shows predictions on a Vistas image 
from models trained on the synthetic source domains Synthia and UrbanSyn (without adaptation), and 
compares them with \ours{}. Several observations arise.  
First, comparing the two source-only models, switching the source domain from 
Synthia to UrbanSyn already improves segmentation quality (e.g., clearer 
\texttt{wall} and \texttt{fence} regions), consistent with our findings in 
Section~\ref{subsec:ablating_urbansyn}.  
Second, comparing source-only results with \ours{} demonstrates the substantial 
benefit of the domain adaptation, which produces markedly better predictions, such 
as correctly identifying distant \texttt{cars} and delivering more coherent 
\texttt{wall} and \texttt{fence} boundaries.  
Finally, the figure also reveals that certain biases in the synthetic datasets 
persist even after adaptation. For example, Synthia does not label road markings 
as \texttt{road}, and this misalignment propagates into the adapted model.  
This highlights the practical importance of ensuring that source- and 
target-domain taxonomies are well aligned.
%%%%%%%%%%%%%%%%%%%%%%
%% Figure City->ACDC, City->MUSES
%%%%%%%%%%%%%%%%%%%%%%
\newcommand{\myviswid}{0.24\textwidth}
\begin{figure*}[t!]
    \centering
    \begin{tabular}{c@{\hskip 0.5pt}c@{\hskip 0.5pt}c@{\hskip 0.5pt}c}

    \begin{tikzpicture}
        \node[inner sep=0] (img) {\includegraphics[width=\myviswid]{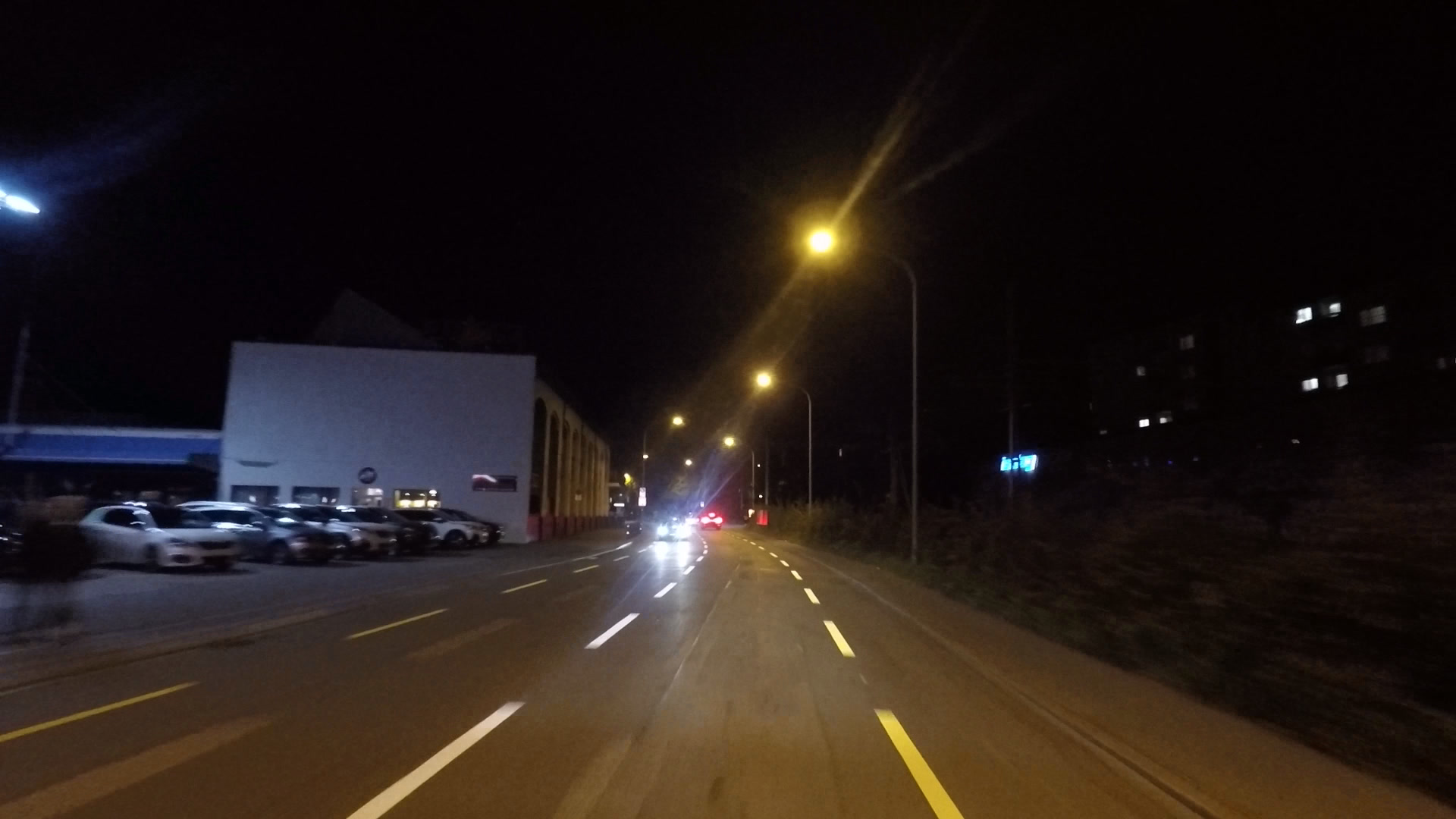}};
        \node[anchor=north west,
              fill=none,
              text=white,
              rounded corners=2pt,
              inner sep=3pt,
              font=\ttfamily\footnotesize] at (img.north west) {ACDC Img};
    \end{tikzpicture}
    &
    \begin{tikzpicture}
        \node[inner sep=0] (img) {\includegraphics[width=\myviswid]{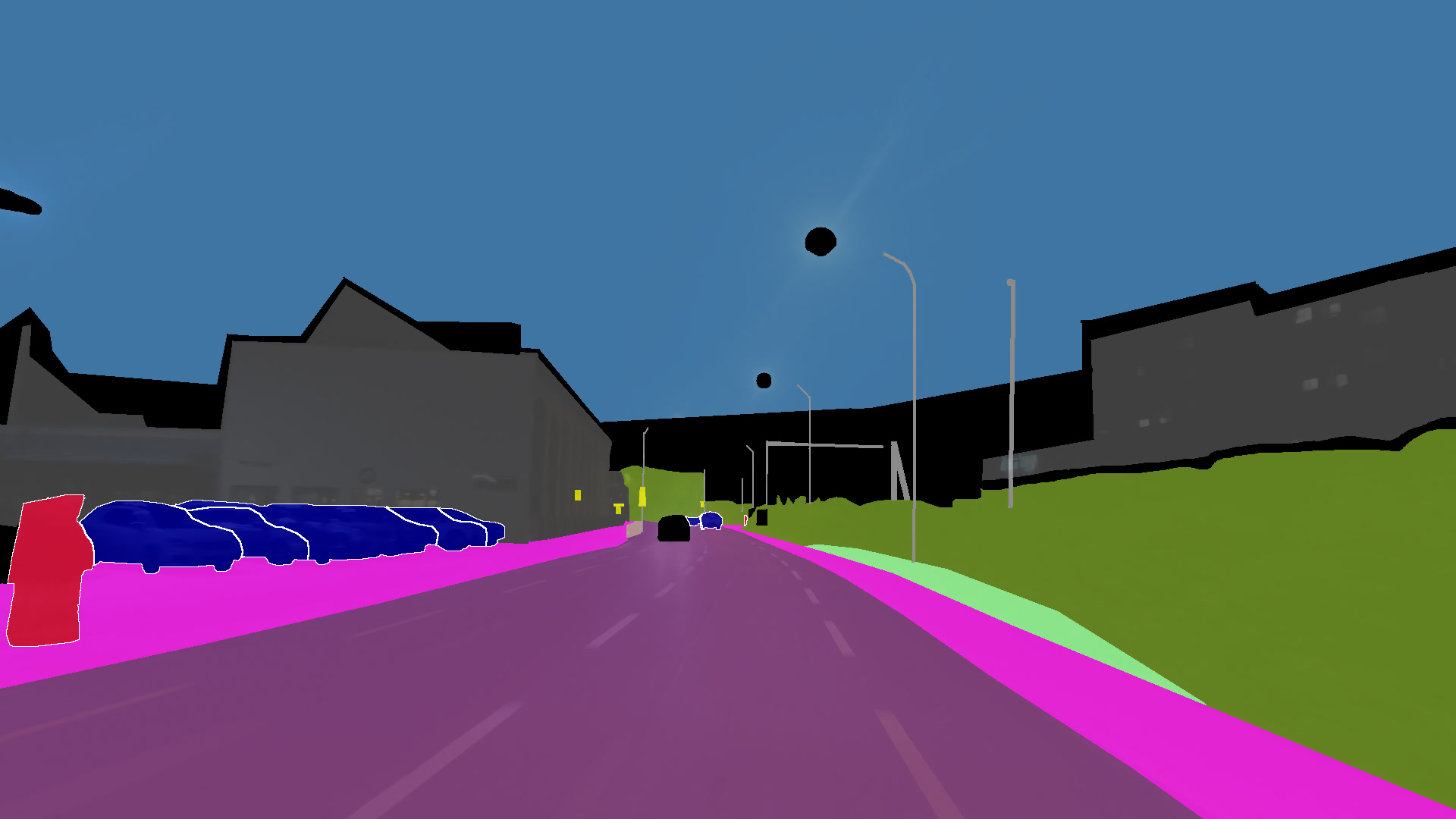}};
        \node[anchor=north west,
              fill=none,
              text=white,
              rounded corners=2pt,
              inner sep=3pt,
              font=\ttfamily\footnotesize] at (img.north west) {Ground-truth};
    \end{tikzpicture}
    & 
    \begin{tikzpicture}
        \node[inner sep=0] (img) {\includegraphics[width=\myviswid]{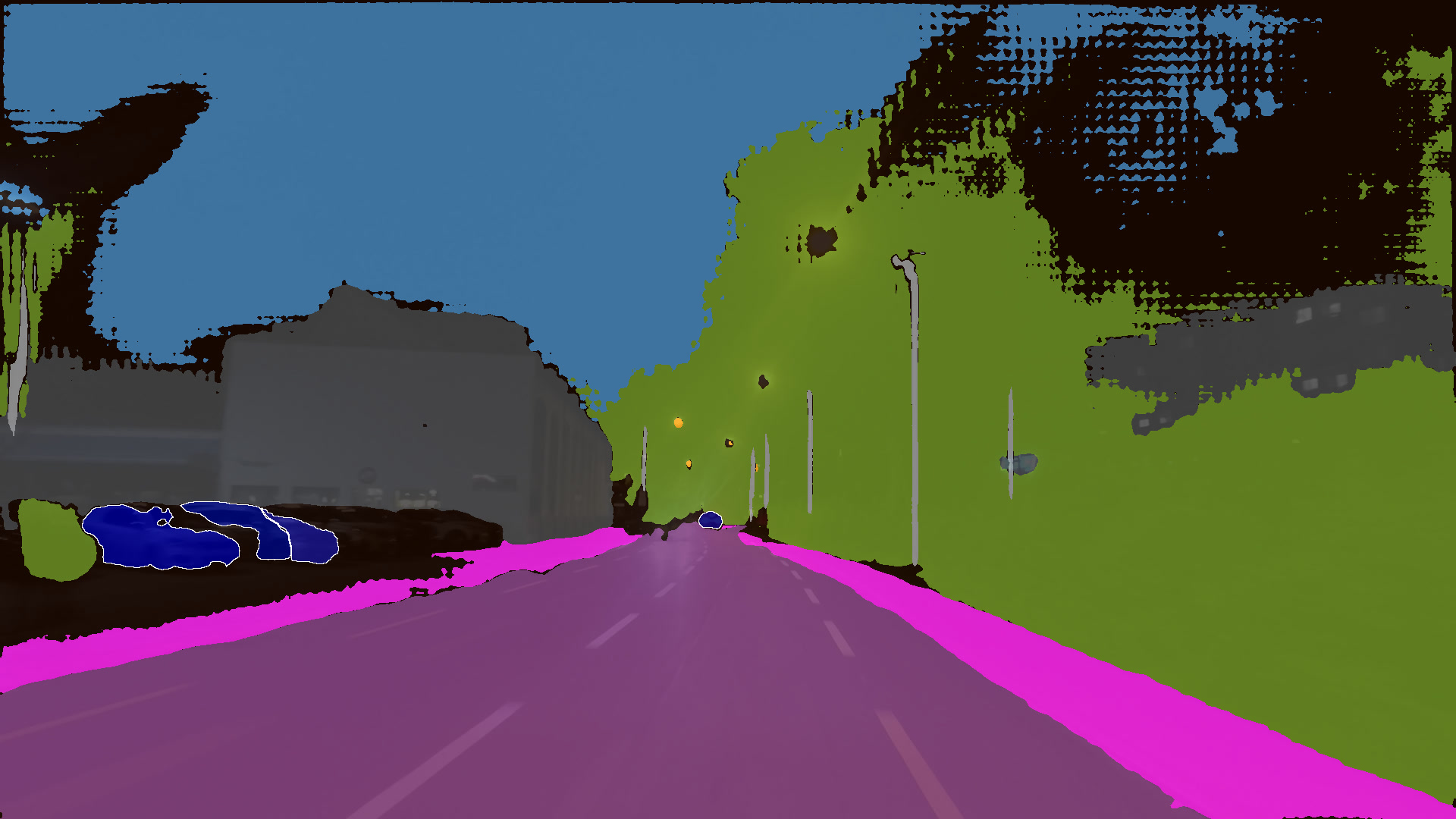}};
        \node[anchor=north west,
              fill=none,
              text=white,
              rounded corners=2pt,
              inner sep=3pt,
              font=\ttfamily\footnotesize] at (img.north west) {Source-only (City)};
    \end{tikzpicture}
    &
    \begin{tikzpicture}
        \node[inner sep=0] (img) {\includegraphics[width=\myviswid]{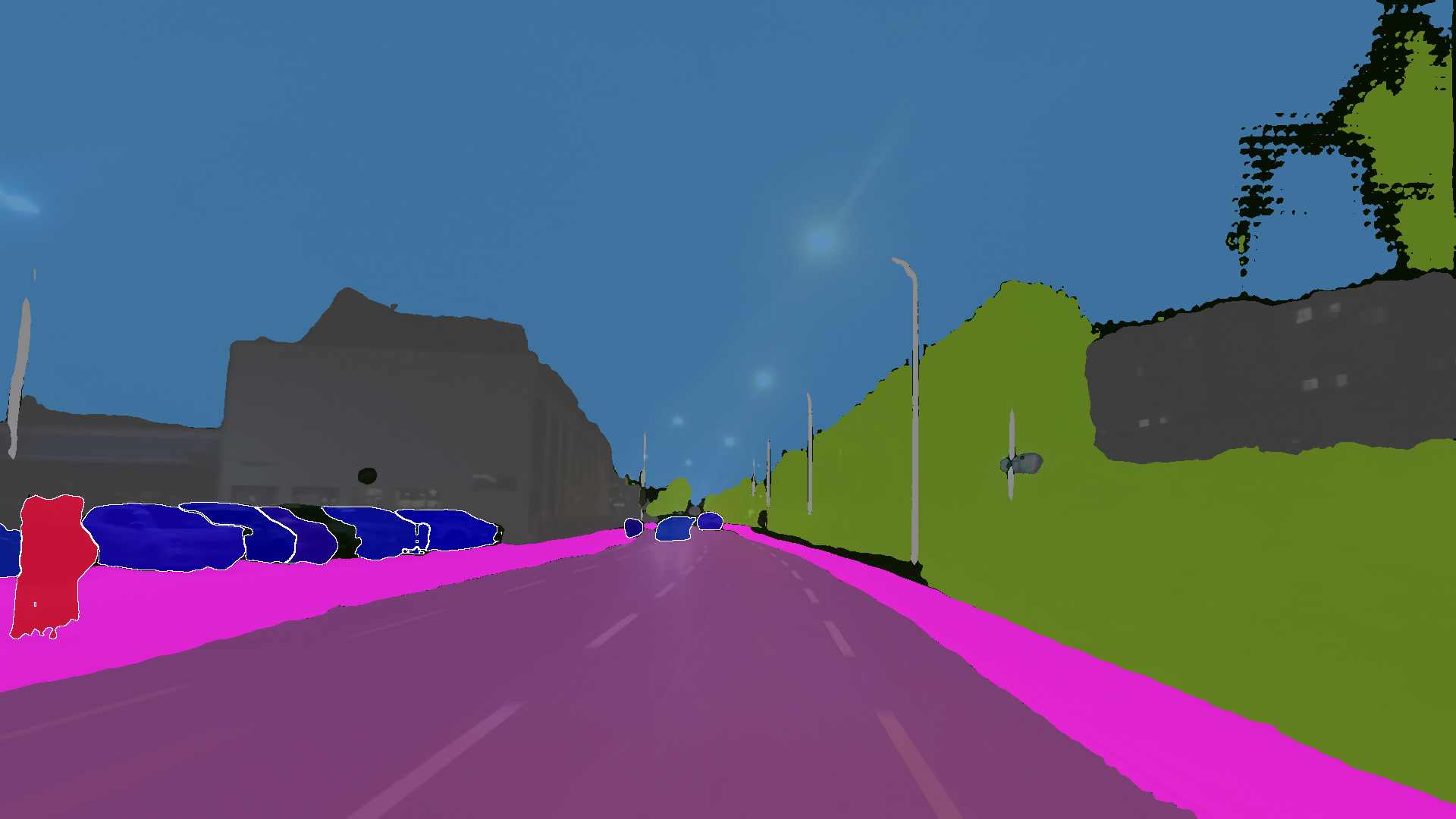}};
        \node[anchor=north west,
              fill=none,
              text=white,
              rounded corners=2pt,
              inner sep=3pt,
              font=\ttfamily\footnotesize] at (img.north west) {\ours{}};
    \end{tikzpicture} \\[-2pt]

        \begin{tikzpicture}
        \node[inner sep=0] (img) {\includegraphics[width=\myviswid]{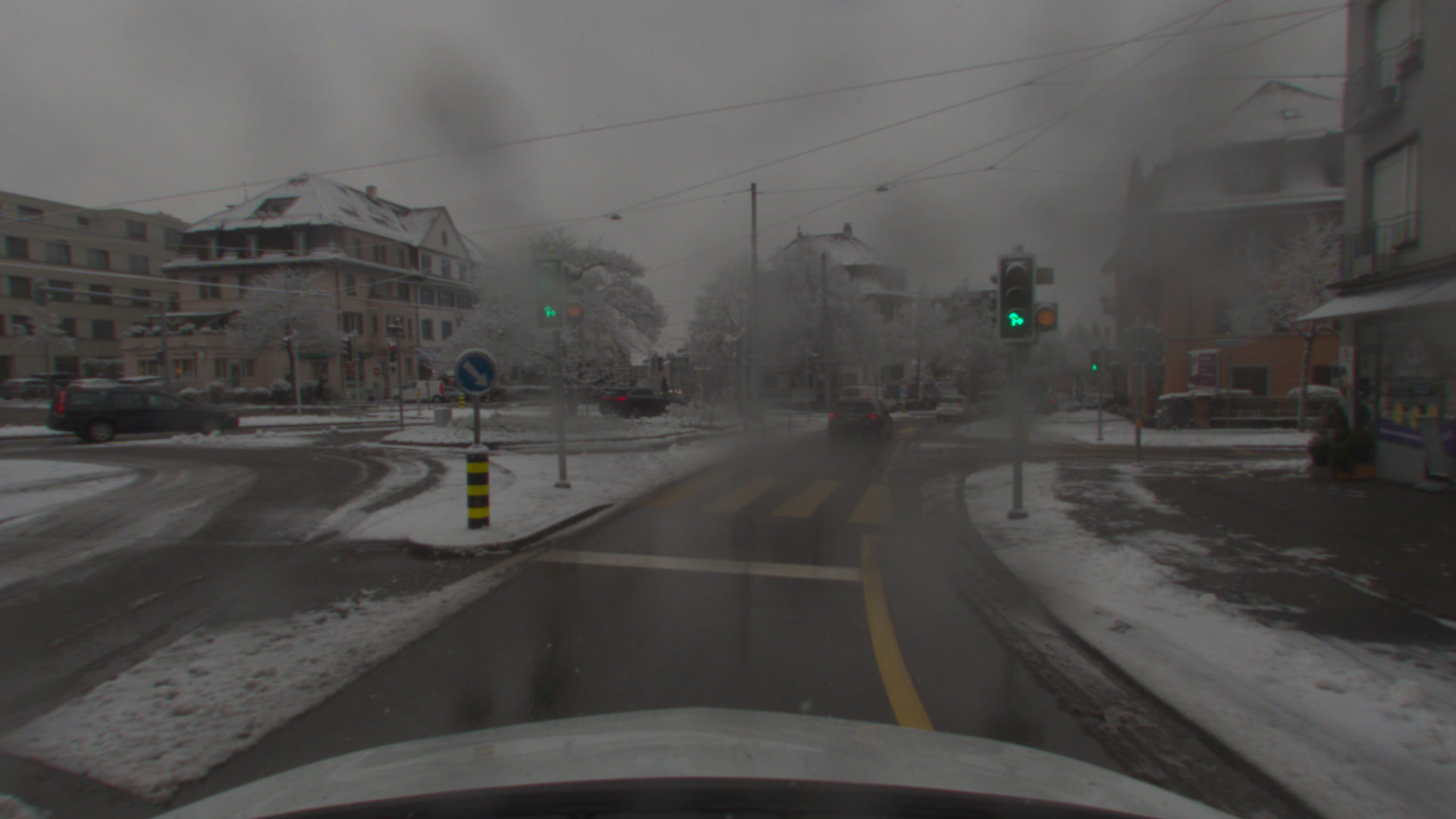}};
        % label
        \node[anchor=north west,
              fill=none,
              text=white,
              rounded corners=2pt,
              inner sep=3pt,
              font=\ttfamily\footnotesize] at (img.north west) {MUSES Img};
    \end{tikzpicture}
    &
    \begin{tikzpicture}
        \node[inner sep=0] (img) {\includegraphics[width=\myviswid]{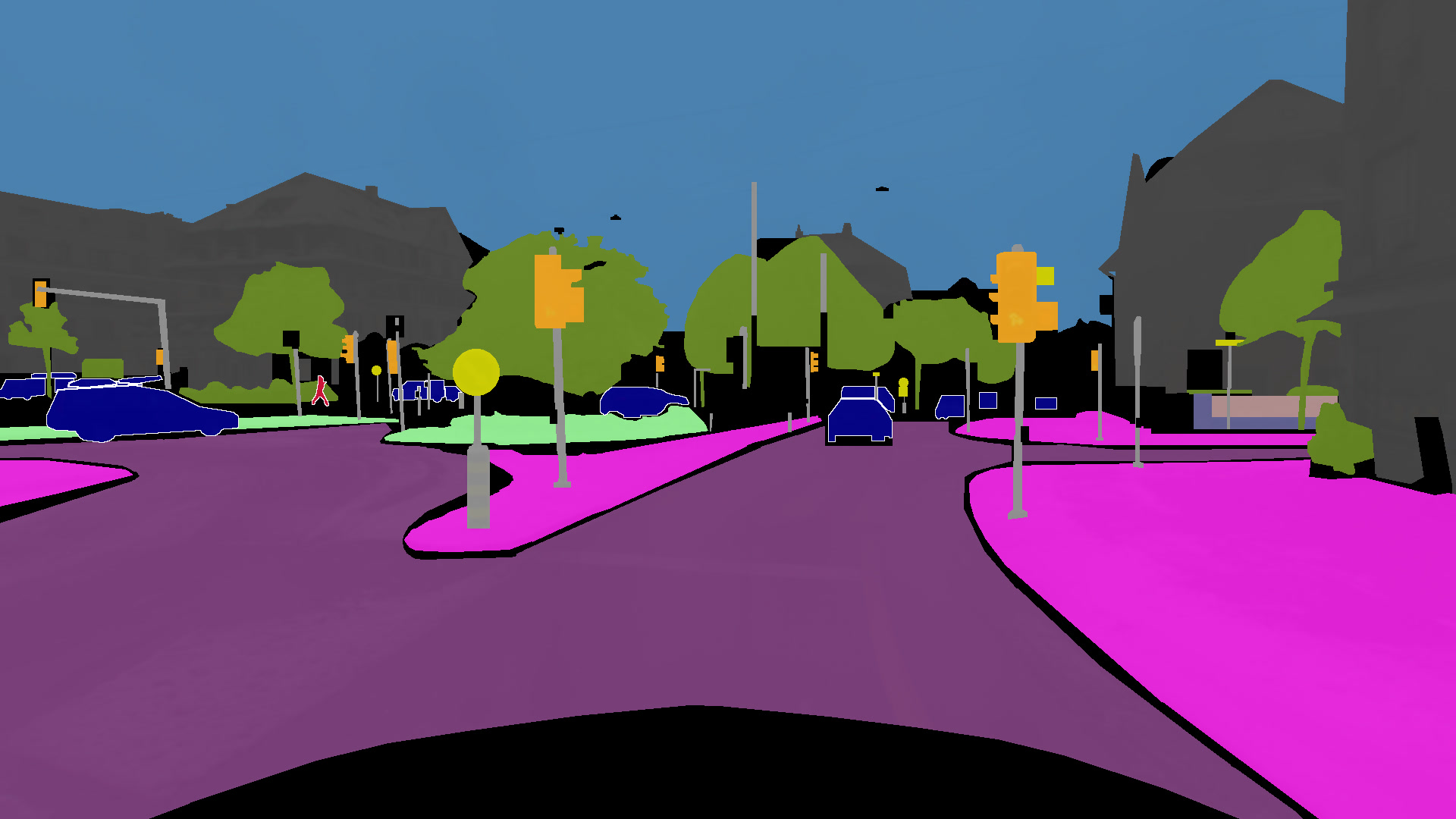}};
        \node[anchor=north west,
              fill=none,
              text=white,
              rounded corners=2pt,
              inner sep=3pt,
              font=\ttfamily\footnotesize] at (img.north west) {Ground-truth};
    \end{tikzpicture}
    &

    \begin{tikzpicture}
        \node[inner sep=0] (img) {\includegraphics[width=\myviswid]{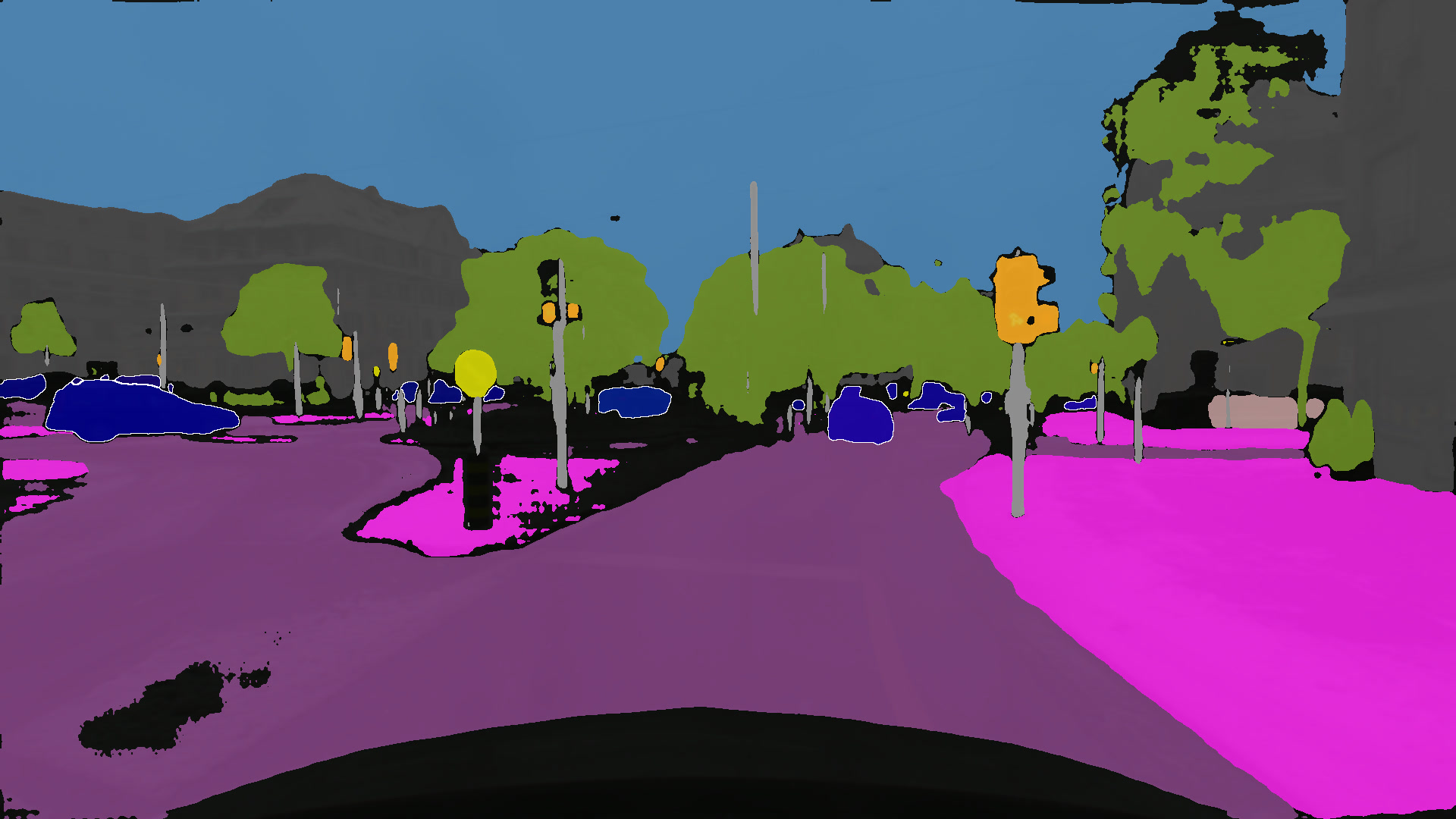}};
        \node[anchor=north west,
              fill=none,
              text=white,
              rounded corners=2pt,
              inner sep=3pt,
              font=\ttfamily\footnotesize] at (img.north west) {Source-only (City)};
    \end{tikzpicture}
    &
    \begin{tikzpicture}
        \node[inner sep=0] (img) {\includegraphics[width=\myviswid]{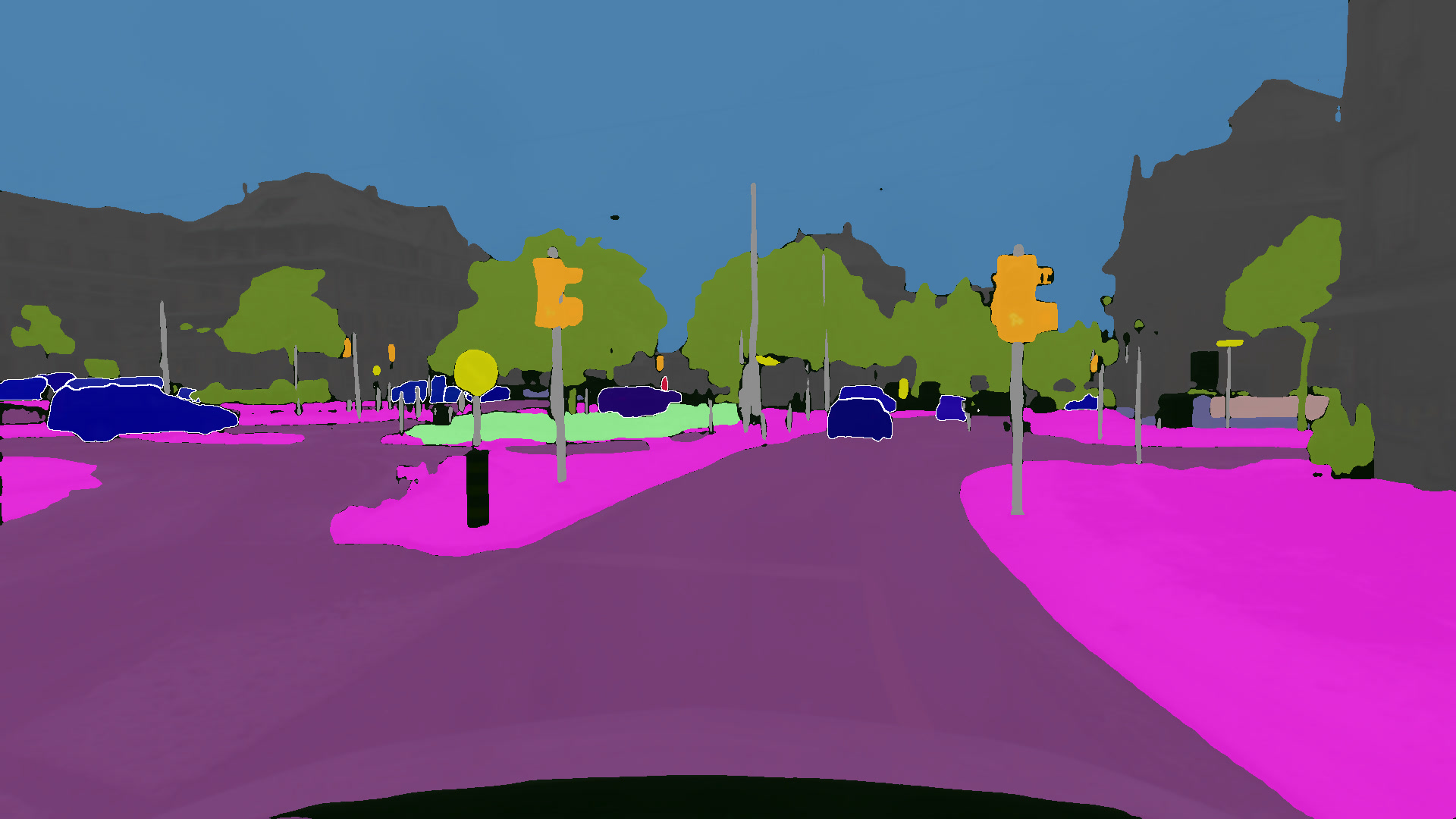}};
        \node[anchor=north west,
              fill=none,
              text=white,
              rounded corners=2pt,
              inner sep=3pt,
              font=\ttfamily\footnotesize] at (img.north west) {\ours{}};
    \end{tikzpicture} 
    \end{tabular}
   \cityscapeslegendall
    \vspace{4pt}
    \caption{
    Qualitative comparison of \ours{} against a source-only model (trained solely on 
Cityscapes) on the clear-to-adverse benchmarks City$\rightarrow$ACDC (top) and 
City$\rightarrow$MUSES (bottom). 
\ours{} yields substantially more accurate segmentations, especially in challenging 
regions such as \textcolor{csperson}{\texttt{person}} instances in top row and distant 
\textcolor{cscar}{\texttt{cars}} in bottom row, despite being trained only on clear-weather 
daytime ground-truth annotations. Best viewed zoomed in.
}
    \label{fig:visualization_acdc_muses}
\end{figure*}
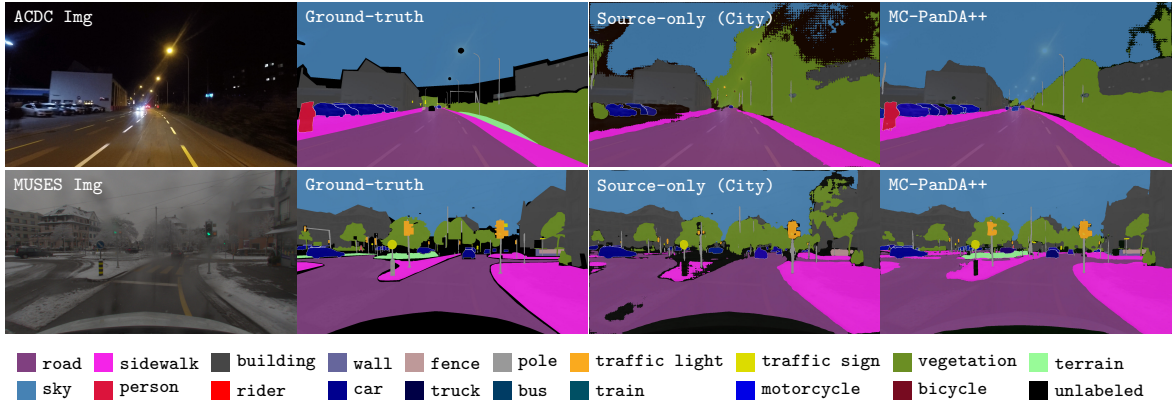

Figure~\ref{fig:visualization_acdc_muses} shows predictions of \ours{} on two 
adverse-weather examples, compared against a source-only baseline trained 
exclusively on Cityscapes. 
Despite never observing ground-truth annotations from adverse conditions, 
\ours{} successfully recovers challenging parts: in the top row it detects a 
barely visible \textcolor{black}{\texttt{person}} and produces clearer segmentation 
of \textcolor{black}{\texttt{sky}}, \textcolor{black}{\texttt{sidewalk}}, and \textcolor{black}{\texttt{car}} regions. 
In the bottom row, \ours{} yields substantially more accurate segmentation under 
snowy and foggy conditions, correctly identifying \textcolor{black}{\texttt{buildings}}, distant 
\textcolor{black}{\texttt{cars}}, and \textcolor{black}{\texttt{traffic lights}}, among others.

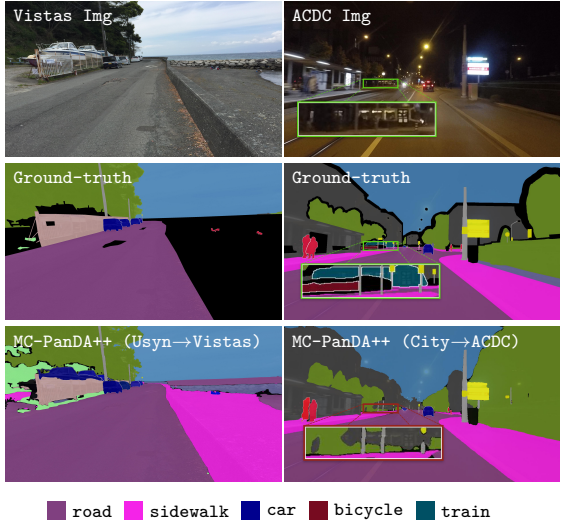
\begin{figure}[h!]
    \centering
    \begin{tabular}{c@{\hskip 1pt}c}

    \begin{tikzpicture}
        \node[inner sep=0] (img) {\includegraphics[width=0.48\linewidth, height=0.81in]{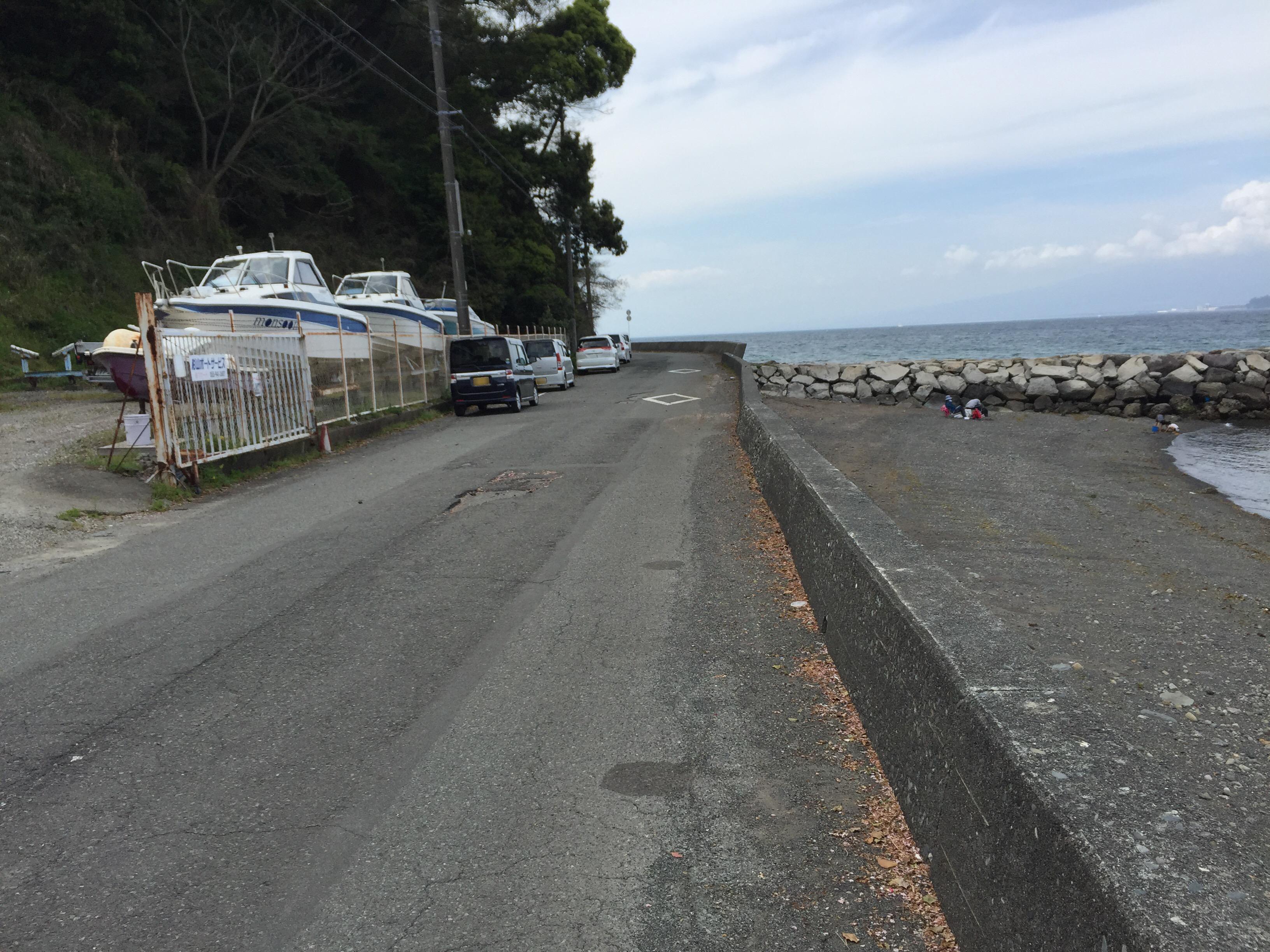}};
        % label
        \node[anchor=north west,
              fill=none,
              text=white,
              rounded corners=2pt,
              inner sep=3pt,
              font=\ttfamily\footnotesize] at (img.north west) {Vistas Img};
    \end{tikzpicture}
    &
    \begin{tikzpicture}
        \node[inner sep=0] (img) {\includegraphics[width=0.48\linewidth, height=0.81in]{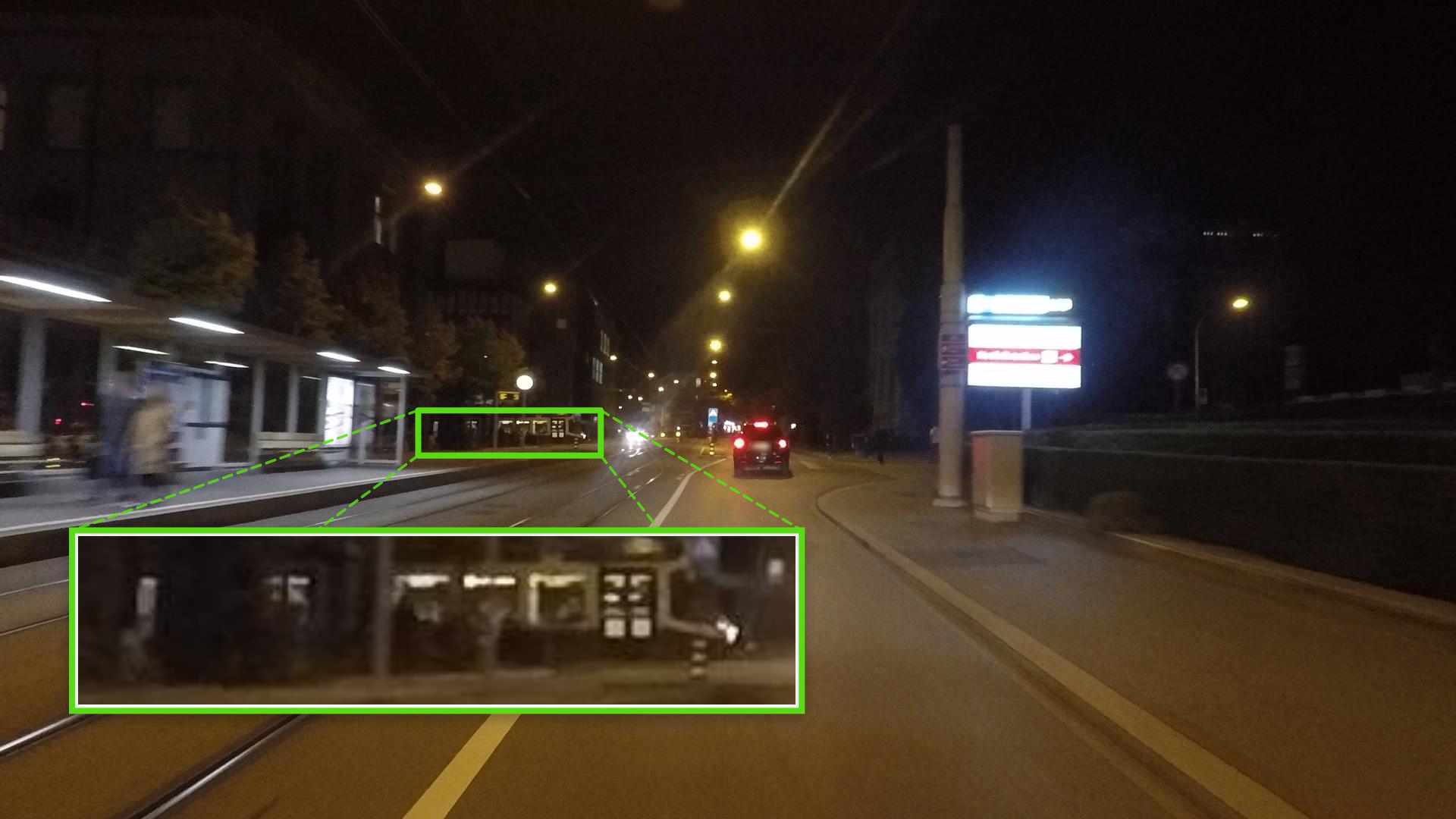}};
        \node[anchor=north west,
              fill=none,
              text=white,
              rounded corners=2pt,
              inner sep=3pt,
              font=\ttfamily\footnotesize] at (img.north west) {ACDC Img};
    \end{tikzpicture}
    \\[-1pt]

    \begin{tikzpicture}
        \node[inner sep=0] (img) {\includegraphics[width=0.48\linewidth, height=0.81in]{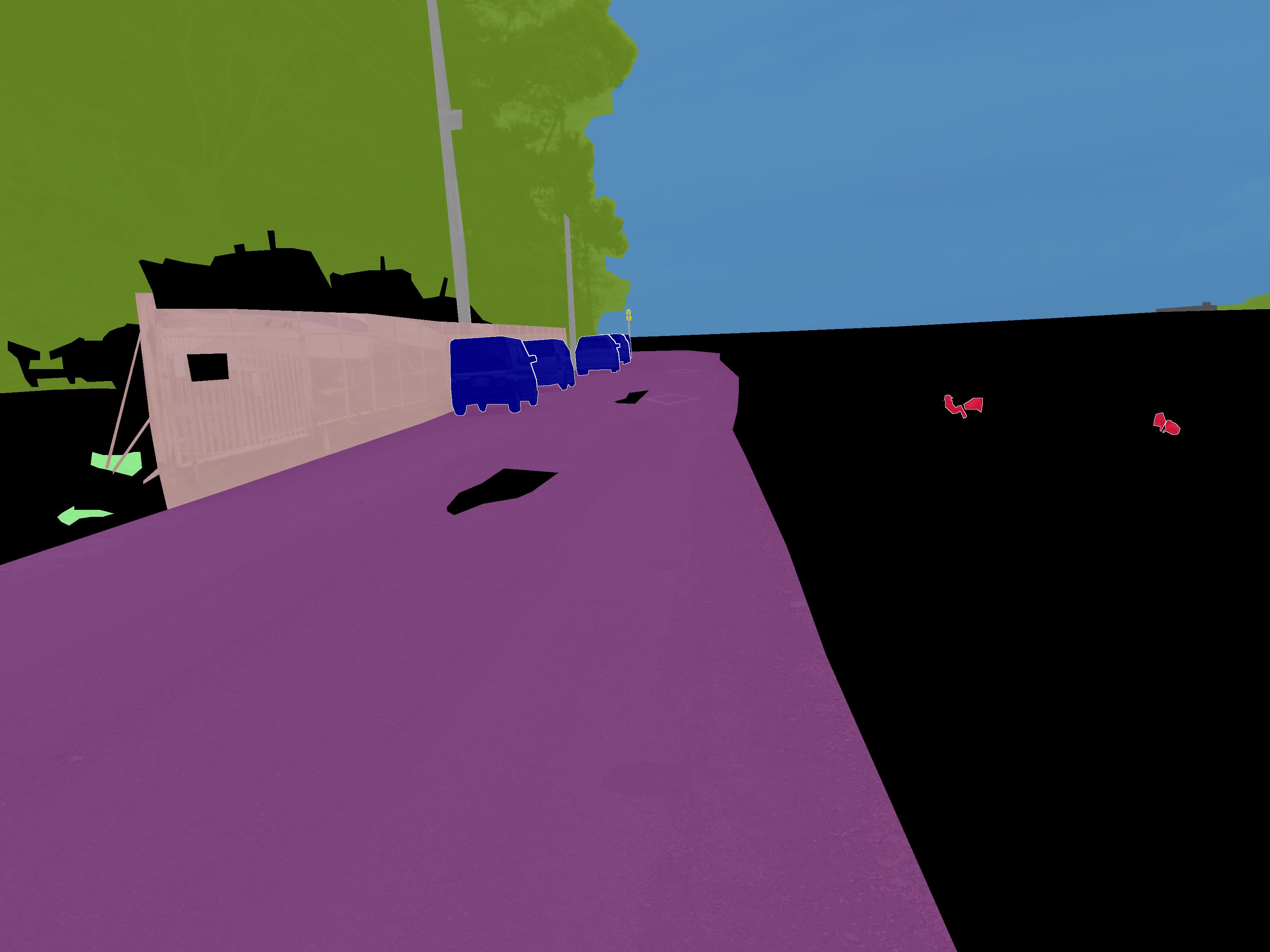}};
        \node[anchor=north west,
              fill=none,
              text=white,
              rounded corners=2pt,
              inner sep=3pt,
              font=\ttfamily\footnotesize] at (img.north west) {Ground-truth};
    \end{tikzpicture}
    &
    \begin{tikzpicture}
        \node[inner sep=0] (img) {\includegraphics[width=0.48\linewidth, height=0.81in]{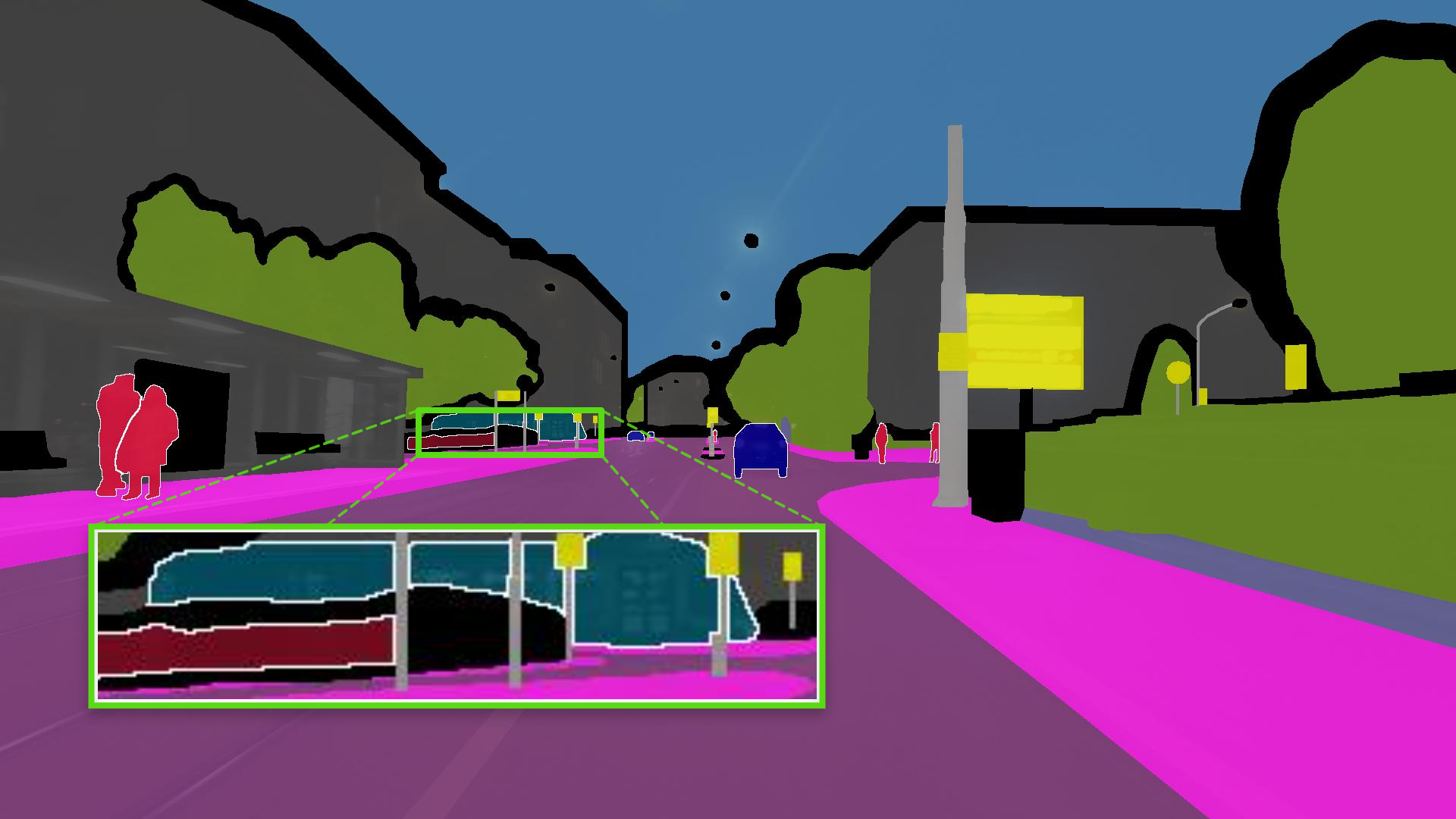}};
        \node[anchor=north west,
              fill=none,
              text=white,
              rounded corners=2pt,
              inner sep=3pt,
              font=\ttfamily\footnotesize] at (img.north west) {Ground-truth};
    \end{tikzpicture}

    \\[-1pt]

    \begin{tikzpicture}
        \node[inner sep=0] (img) {\includegraphics[width=0.48\linewidth, height=0.81in]{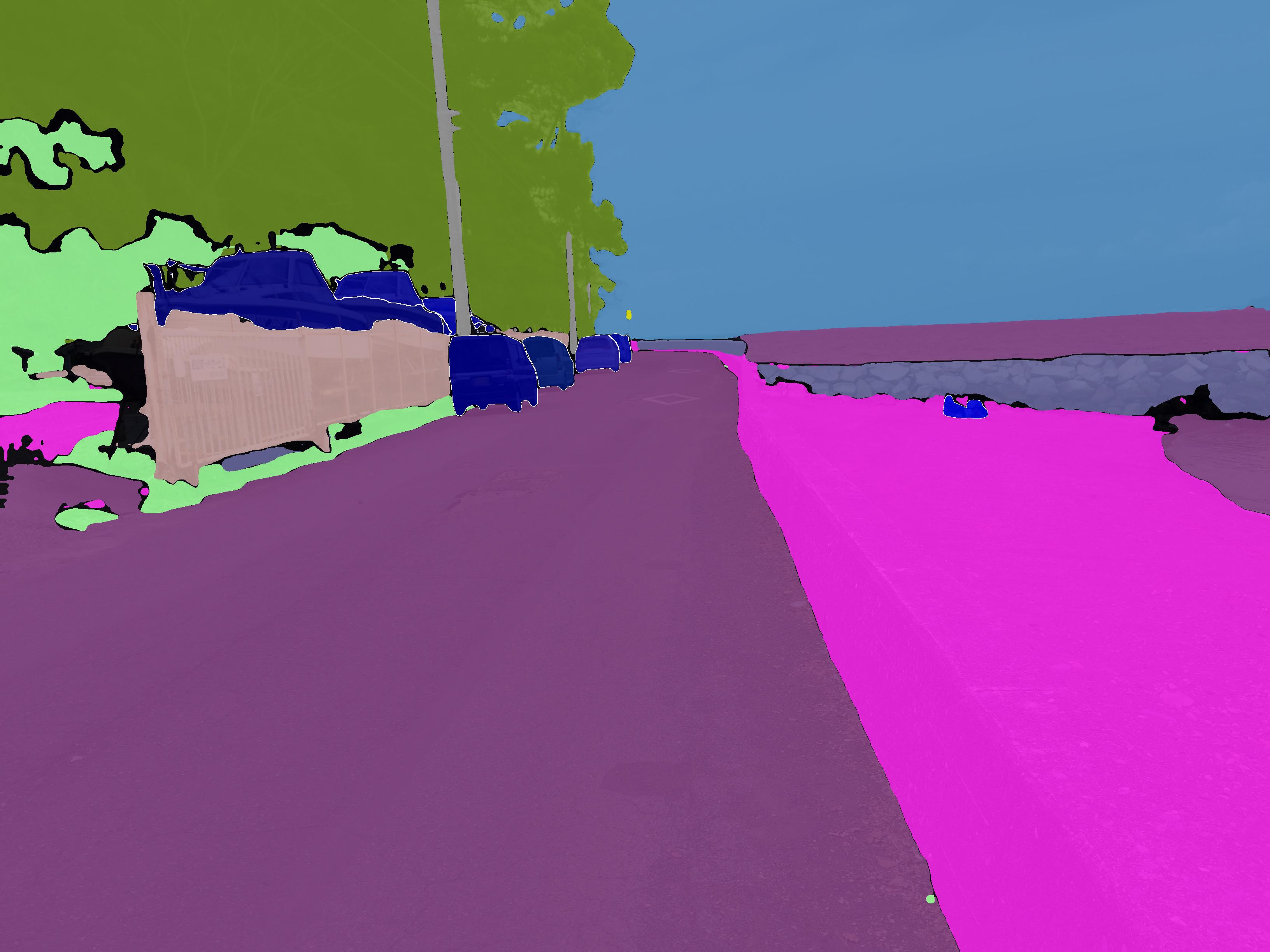}};
        \node[anchor=north west,
              fill=none,
              text=white,
              rounded corners=2pt,
              inner sep=3pt,
              font=\ttfamily\footnotesize] at (img.north west) {\ours{} (Usyn$\rightarrow$Vistas)};
    \end{tikzpicture}
    &
    \begin{tikzpicture}
        \node[inner sep=0] (img) {\includegraphics[width=0.48\linewidth, height=0.81in]{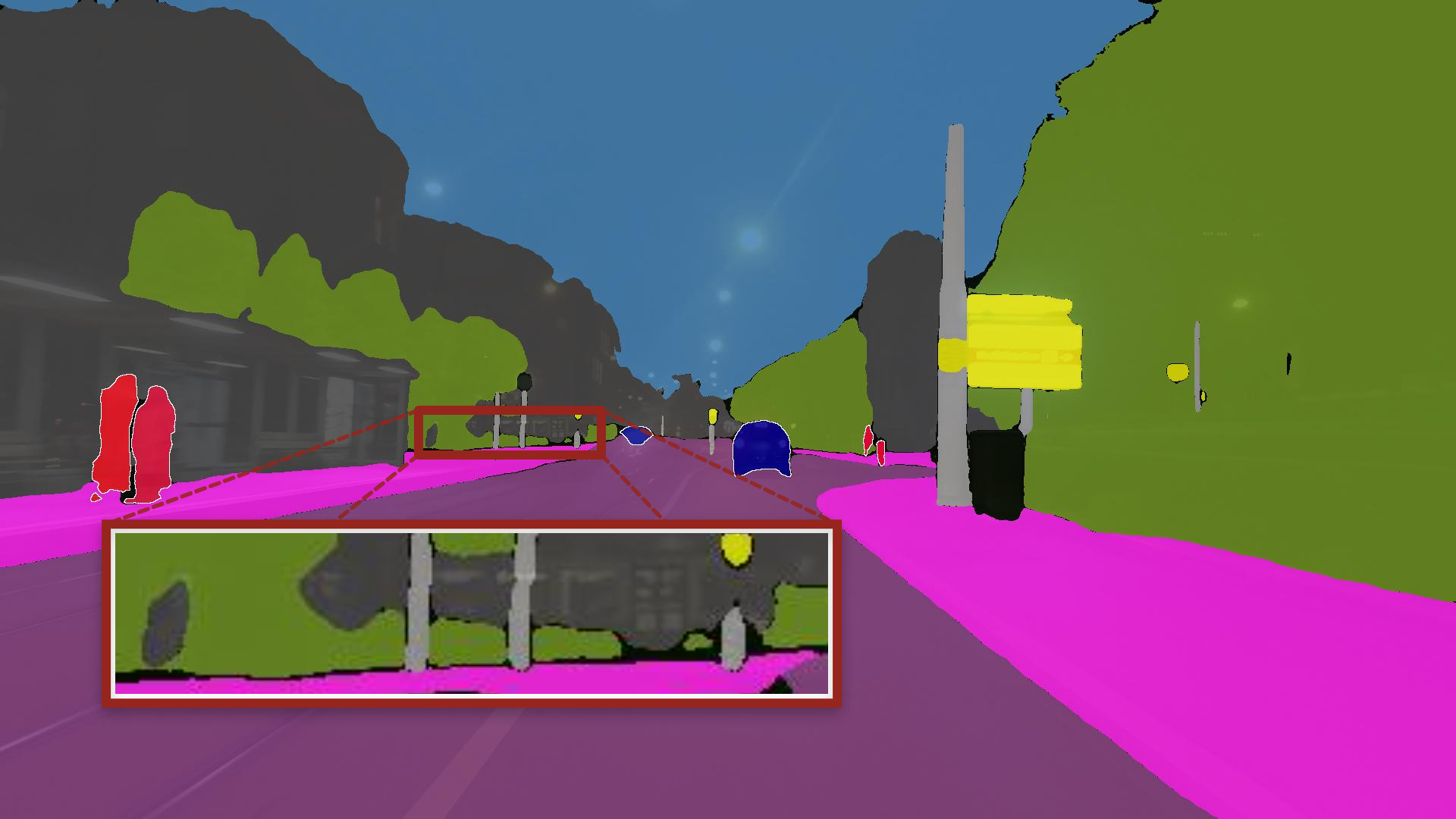}};
        \node[anchor=north west,
              fill=none,
              text=white,
              rounded corners=2pt,
              inner sep=3pt,
              font=\ttfamily\footnotesize] at (img.north west) {\ours{} (City$\rightarrow$ACDC)};
    \end{tikzpicture}
    \end{tabular}
    \cityscapeslegendfailure
    \vspace{4pt}
    \caption{
\textcolor{changesframe}{
Representative failure cases of \ours{}. Left: UrbanSyn$\rightarrow$Vistas example illustrating
source-domain coverage limitations, where target-domain regions such as \textit{sea} and
\textit{boats} are mapped to the closest available road-scene categories, such as
\texttt{\textcolor{csroad}{road}} and \texttt{\textcolor{cscar}{car}}. Right: Cityscapes$\rightarrow$ACDC nighttime example, where
distant or weakly visible objects, such as \texttt{\textcolor{cstrain}{train}} and \texttt{\textcolor{csbicycle}{bicycle}} regions, remain difficult to
segment under limited visual evidence.
}
}
    \label{fig:failure_cases}
\end{figure}

\noindent\textbf{\textcolor{changesframe}{Failure cases.}}
\textcolor{changesframe}{
Fig.~\ref{fig:failure_cases} shows representative failure cases of \ours{}. The left example
illustrates source-domain coverage limitations in UrbanSyn$\rightarrow$Vistas. When the target
domain contains visual concepts that are poorly represented in the source domain,
the model can map them to the closest available source-domain categories:
\textit{sea} regions are predicted as \texttt{\textcolor{csroad}{road}}, and \textit{boats} as \texttt{\textcolor{cscar}{cars}}. The right example shows a nighttime ACDC scene, where distant objects remain difficult to segment due to limited visual evidence. Additional failure
cases, including ambiguities such as persons on billboards
being segmented as real \texttt{\textcolor{csperson}{person}} instances, are provided in Appendix~\ref{app:failure_cases}.
}

\section{Conclusion}                    \label{sec:conclusion}

In this work, we presented \ours{}, a novel approach for 
domain-adaptive panoptic segmentation.
The proposed framework enables state-of-the-art panoptic architectures based on mask transformers to be effectively trained under domain shift within a teacher-student consistency-learning paradigm. We address the key challenge of such paradigms -- confirmation bias -- which arises from the repeated reinforcement of incorrect pseudo-labels. To mitigate this, our method prevents learning from unreliable pseudo-labels through mask-wide loss scaling based on a teacher confidence.
Furthermore, we introduce a novel point sampling mechanism that prioritizes loss computation in informative locations according to student uncertainty while simultaneously avoiding unreliable locations using teacher confidence. 

Compared to our preliminary approach~\citep{martinovic2024eccv}, \ours{} is both more accurate and more robust due to two key advancements. First, it employs class-dependent and self-adapting mask-wide loss scaling, which improves training stability, mitigates sensitivity to the initial confidence threshold, and consistently yields superior performance. Second, it leverages strong self-supervised initialization, which enables the simplification of the original three-stage training design into a streamlined single-stage pipeline, thereby reducing both conceptual complexity and the number of hyperparameters.

\ours{} sets a new state of the art in panoptic domain adaptation across standard benchmarks. While baseline consistency training suffers from collapse caused by confirmation bias, \ours{} maintains a stable learning curve through the proposed confidence-guided training strategy. Comprehensive ablation studies confirm the effectiveness of our contributions across different backbones and setups, including synthetic-to-real, real-to-real, and clear-to-adverse scenarios. 

We anticipate that \ours{} will facilitate the broader adoption of panoptic segmentation and support the more reliable deployment of mask transformers in real-world applications. Moreover, its design is inherently compatible with ongoing advancements in backbone pre-training, mask-transformers, and pseudo-label refinement, ensuring its relevance in future research.

\backmatter

\bmhead{Acknowledgements}
This work was supported by the Croatian Recovery and Resilience Fund -- NextGenerationEU (grant C1.4 R5-I2.01.0001) and the Croatian Science Foundation (grants DOK-NPOO-2023-10-2288 and MOBDOK-2023-4880). We also acknowledge the advanced computing resources provided by the University of Zagreb University Computing Centre (SRCE). 
Josip Šarić has received funding from the European Union’s Horizon Europe research and innovation program under the Marie Sklodowska-Curie COFUND Postdoctoral Programme grant agreement No.~101081355-SMASH and from the Republic of Slovenia and the European Union from the European Regional Development Fund.

We thank Jan Šnajder, Ivan Grubišić, Ryousuke Yamada, and Naomi Kombol for valuable feedback that helped improve the manuscript. 

In memory of Siniša Šegvić, who passed away before this work was published. He will be missed as a dear friend, mentor, and moral compass.
\bmhead{Disclaimer}
Co-funded by the European Union. Views and opinions expressed are however those of the author(s) only and do not necessarily reflect those of the European Union or European Research Executive Agency. Neither the European Union nor the granting authority can be held responsible for them.
\bmhead{Data availability statement}
We conduct our experiments on the
following publicly available datasets:
Cityscapes~\citep{Cordts_2016_CVPR}, 
Mapillary Vistas~\citep{Neuhold_2017_ICCV},
Foggy Cityscapes~\citep{sakaridis2018semantic},
ACDC~\citep{sakaridis21iccv},
MUSES~\citep{brodermann2024muses},
Synthia~\citep{Ros_2016_CVPR} and
UrbanSyn~\citep{gomez2025urbansyn}.

%\clearpage
\bibliography{sn-bibliography}% common bib file

\clearpage

\begin{appendices}

\appendix

\setcounter{figure}{0}
\setcounter{table}{0}

\renewcommand{\thefigure}{A\arabic{figure}}
\renewcommand{\thetable}{A\arabic{table}}

% important for hyperref anchors
\renewcommand{\theHfigure}{A.\arabic{figure}}
\renewcommand{\theHtable}{A.\arabic{table}}

\section{UrbanSyn filtering}\label{appendix:corrupted_urbansyn}

\begin{figure}[h!]
    \centering
    \begin{tabular}{c@{\,}c}
         \includegraphics[width=\mywt]{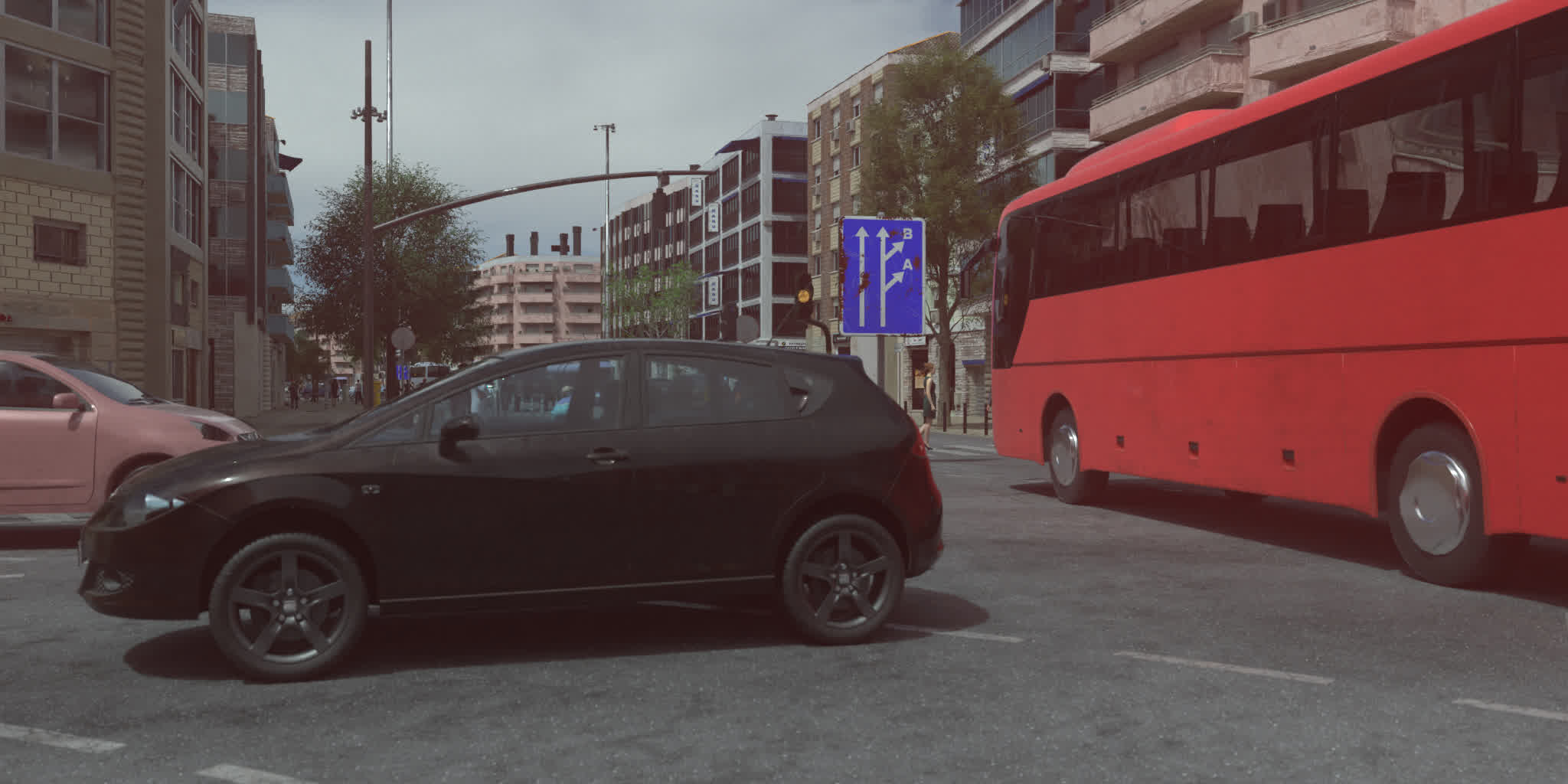} &
         \includegraphics[width=\mywt]{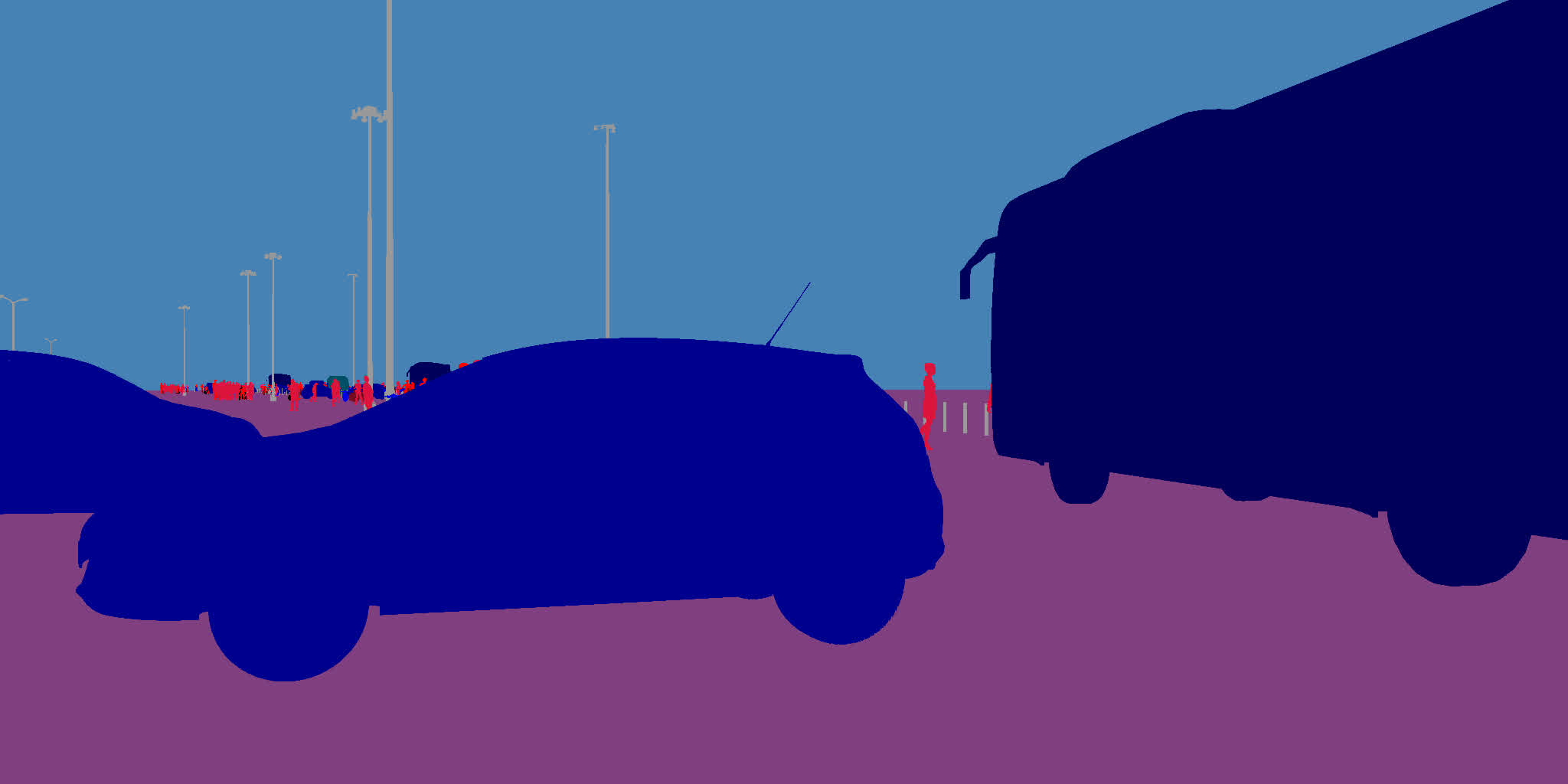} \\ 
    \end{tabular}
    \vspace{4pt}
    \cityscapeslegendmislabeled
    \caption{
Example of a mislabeled UrbanSyn image (ID \texttt{7014}) where \texttt{building}, 
\texttt{vegetation}, and \texttt{traffic sign} regions are incorrectly annotated 
as \texttt{road} or \texttt{sky}.
}
\label{fig:urbansyn_corrupted}
\end{figure}

During an initial inspection of UrbanSyn~\citep{gomez2025urbansyn},  
we identified a small subset of images with clearly corrupted semantic 
annotations (e.g., mislabeled \texttt{road} or \texttt{sky} regions;  
see Fig.~\ref{fig:urbansyn_corrupted}).

A closer inspection revealed that these errors were not isolated:  
a contiguous block of images, from ID \texttt{6973} to \texttt{7029}, 
contained consistently incorrect labels.  
Given that UrbanSyn contains approximately 7.5k labeled images, manually 
inspecting the entire dataset would be infeasible.
To systematically verify whether additional corrupted labels existed, we
designed the following screening procedure:
\begin{enumerate}
    \item Apply Mask2Former (M2F)~\citep{cheng2022masked}, trained on Cityscapes,  
          to all labeled UrbanSyn images.
    \item Compute per-image M2F losses for the entire dataset.
    \item Select images whose losses fall within the top 10\% as potential outliers.
    \item Manually inspect these high-loss images to determine whether additional
          annotation errors are present.
\end{enumerate}
This pipeline successfully re-identified 56 out of the 57 originally known 
corrupted images and revealed no further mislabeled examples.  
To avoid supervision noise in our experiments, we therefore removed all images 
with IDs \texttt{6973}--\texttt{7029} from the UrbanSyn labeled training set. \

\textcolor{changesframe}{
All experiments that use UrbanSyn as the source domain are conducted on this filtered split,
including the source-only baselines and all \ours{} variants. To quantify the effect of this
filtering step, Table~\ref{tab:app_urbansyn_filtering_impact} compares the final performance
of \ours{} with and without removing the corrupted images. The impact is minor: filtering
changes the final performance by only +0.2 PQ on UrbanSyn$\rightarrow$Cityscapes and +0.3 PQ
on UrbanSyn$\rightarrow$Vistas, which is within the observed run-to-run variation. Nevertheless,
we use the filtered split in all reported experiments to avoid known annotation noise and to
provide a clean and reproducible benchmark setup.
}

\begin{table}[t!]
\centering
\begingroup
\color{changesframe}
\caption{
Impact of UrbanSyn filtering on the final performance of \ours{}.
We report mean$\pm$std PQ over three random seeds.
}
\label{tab:app_urbansyn_filtering_impact}
\footnotesize
\setlength{\tabcolsep}{8pt}
\vspace{0.5em}
\begin{tabular}{ccc}
\toprule
UrbanSyn filtering 
& USyn$\rightarrow$City 
& USyn$\rightarrow$Vistas \\
\midrule
\no{}  & \mstd{57.0}{0.7} & \mstd{49.2}{0.5} \\
\yes{} & \mstd{57.2}{0.7} & \mstd{49.5}{0.2} \\
\bottomrule
\end{tabular}
\endgroup
\end{table}

\section{Implementation details} \label{sec:implementation_details_appendix}
\subsection{Data augmentations}
We follow the augmentation setup used in MC-PanDA~\citep{martinovic2024eccv}, repeated here for completeness.
Our baseline image transformations follow the standard Mask2Former~\citep{cheng2022masked} pipeline: random resizing with fixed aspect ratio, random cropping, horizontal flipping, and SSD-style color jittering~\citep{liu2016ssd}. All experiments use a crop size of $512\times1024$ pixels. For the shorter image side, we sample uniformly from:
\begin{itemize}
\item[$\bullet$] \texttt{[512, 2048]} for Cityscapes, Foggy Cityscapes, UrbanSyn and Vistas,
\item[$\bullet$] \texttt{[640, 1408]} for Synthia,
\item[$\bullet$] \texttt{[540, 2160]} for ACDC and MUSES.
\end{itemize}
As in MC-PanDA, color jitter is disabled for the teacher branch.
The student branch receives additional strong augmentations implemented via \texttt{torchvision}:
\begin{enumerate}
    \item 
    \texttt{ColorJitter}(\\
    \texttt{\text{brightness}=(0.2, 1.8)},\\
    \texttt{\text{contrast}=(0.2, 1.8)}, \\
    \texttt{\text{saturation}=(0.2, 1.8)}, \\ 
    \texttt{\text{hue}=(-0.2, 0.2)})
    \item \texttt{RandomGrayscale(\text{\text{p}=0.2})}
    \item \texttt{RandomApply(\\
    GaussianBlur(\text{\text{sigma}=(0.1, 2)),
    \textbf{p}=0.5}\\)}
\end{enumerate}
\newcommand{\mywa}{0.235\linewidth}
\begin{figure}[h!]
    \centering
    \begin{tabular}{l@{\,}l@{\,}l@{\,}l}
        \tiny Source Image & \tiny Target Image & \tiny Student Image & \tiny Pseudo-labels \\
         \includegraphics[width=\mywa]{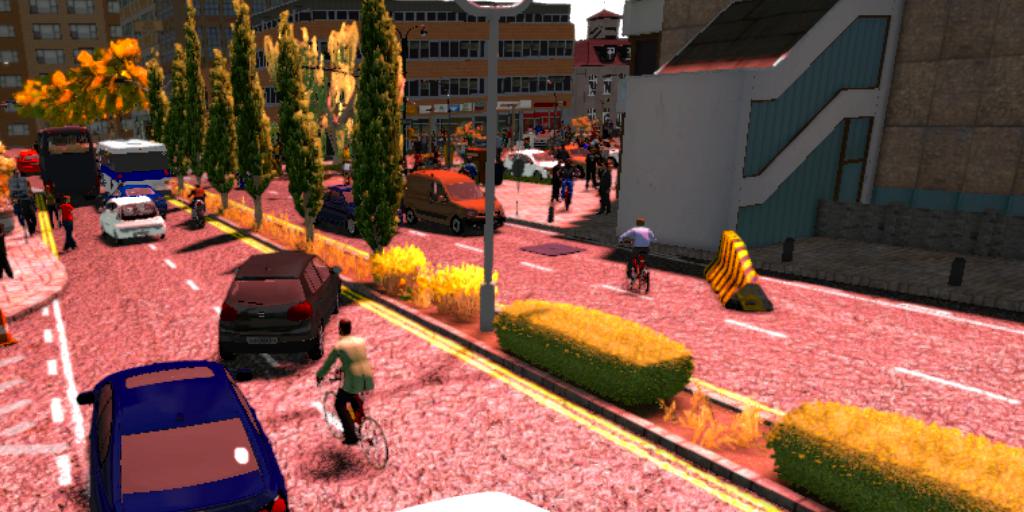} & 
         \includegraphics[width=\mywa]{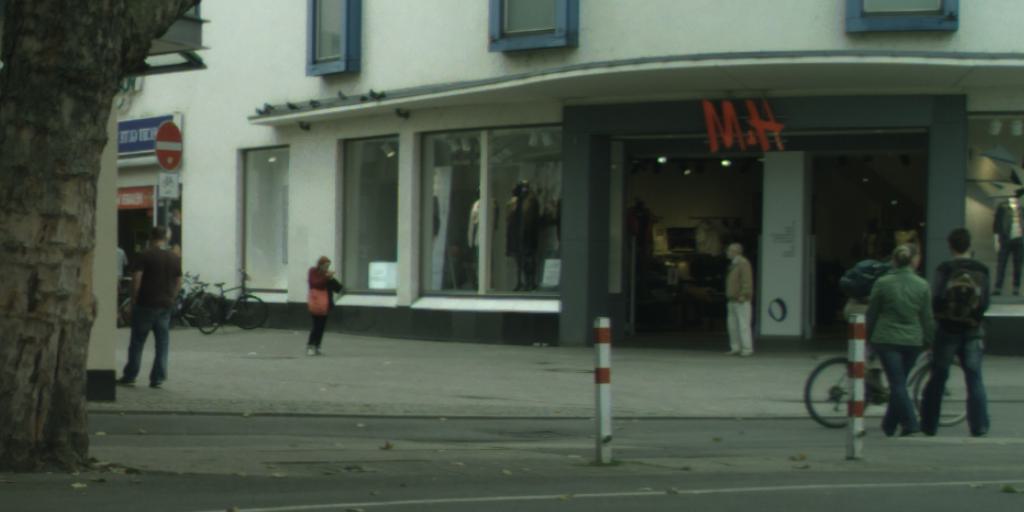} &
         \includegraphics[width=\mywa]{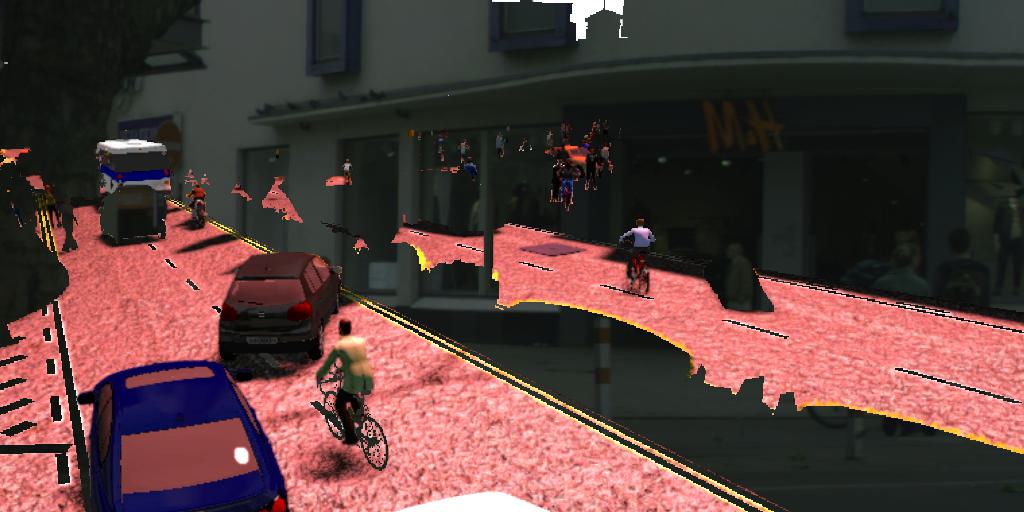} &
         \includegraphics[width=\mywa]{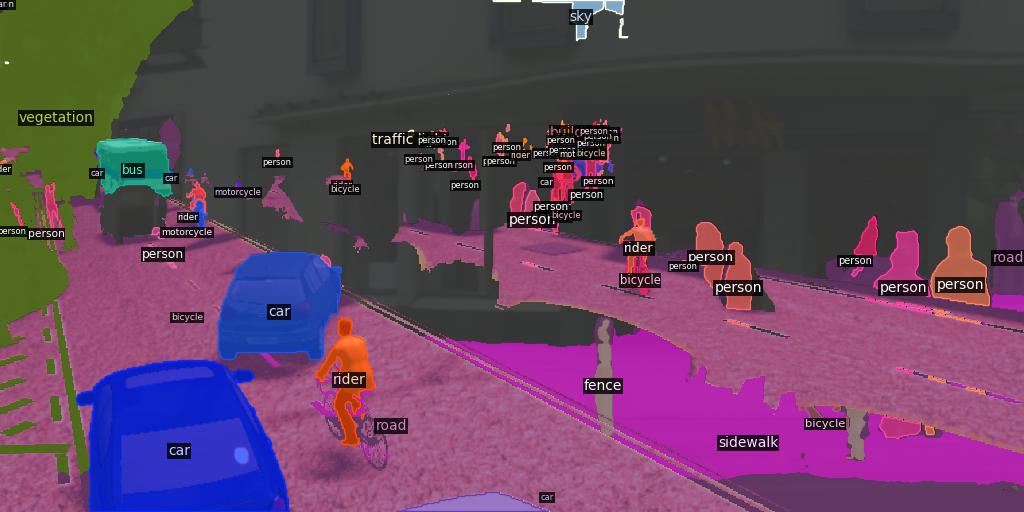} \\[-0.3em]

         \includegraphics[width=\mywa]{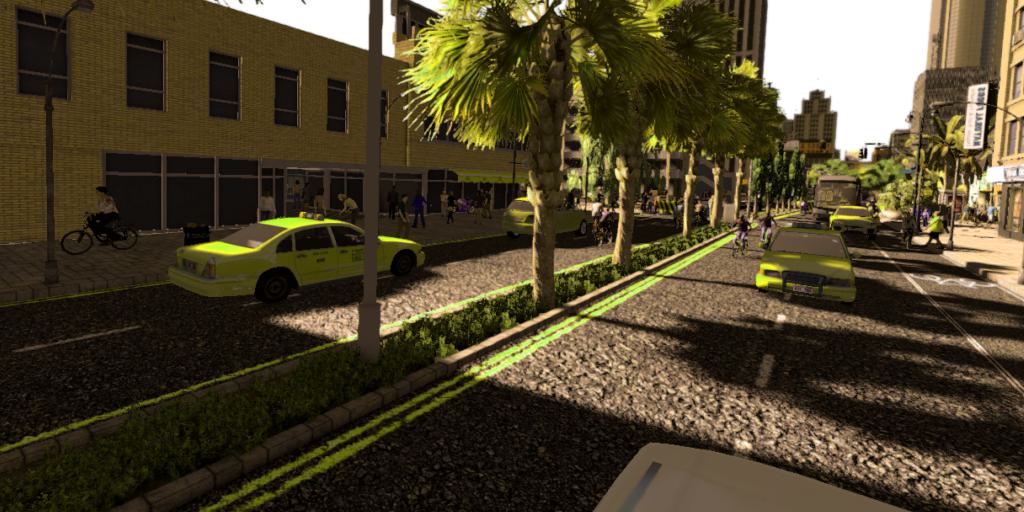} & 
         \includegraphics[width=\mywa]{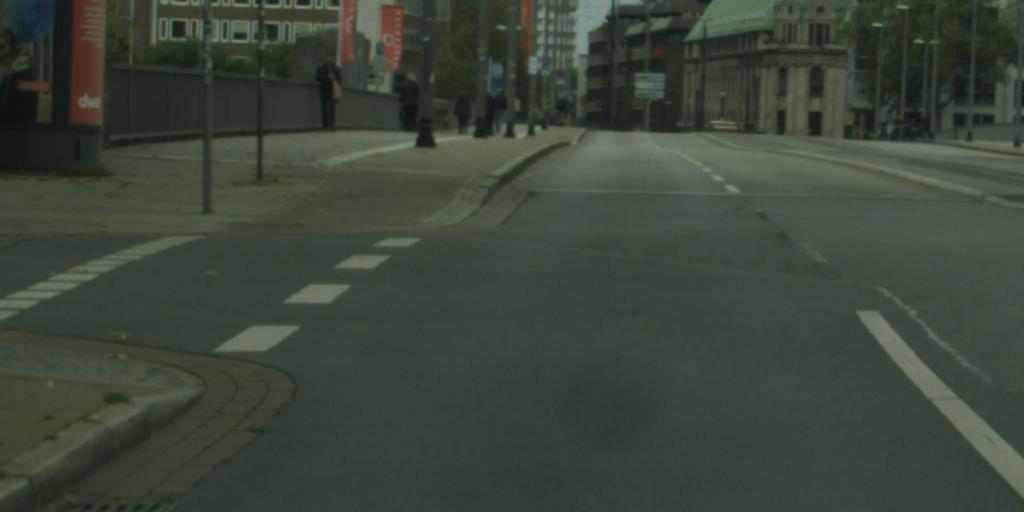} &
         \includegraphics[width=\mywa]{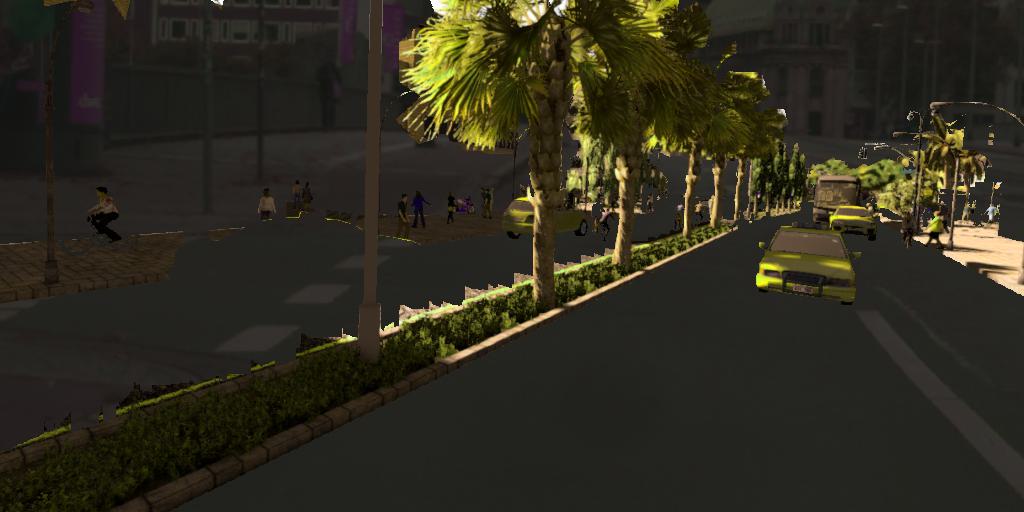} &
         \includegraphics[width=\mywa]{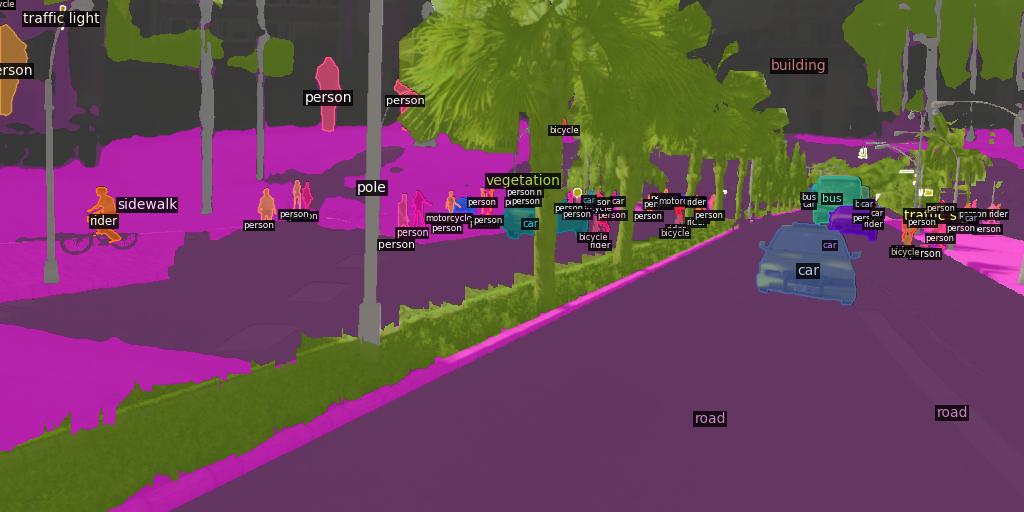} \\

    \end{tabular}
    \caption{
Illustration of SegMix during training.
For each target image, half of the \emph{segments} from a labeled source image are randomly selected and pasted onto the target image.
This differs from ClassMix, which selects half of the \emph{classes} and pastes all associated segments.
}
    \label{fig:train_ex_vis}
\end{figure}
Finally, we apply SegMix to combine labeled source-domain segments with unlabeled target images, as described in the main manuscript.
Figure~\ref{fig:train_ex_vis} shows representative training examples from the Synthia$\rightarrow$Cityscapes experiment.
The first two columns illustrate the augmented source and target images.
The last two columns show the corresponding student inputs and pseudo-labels after SegMix, which combine teacher predictions (target domain) with ground-truth annotations (source domain).
We note that segmentation-heavy source images occasionally dominate the student view, an aspect of segment-oriented mixing that may offer additional opportunities for future refinement

\subsection{Training hyperparameters}
In experiments where Cityscapes or UrbanSyn serve as the source domain, we use 100 Mask2Former queries, following~\cite{cheng2022masked}. For Synthia$\rightarrow$* experiments, we increase the number of queries to 200 due to the substantially larger number of instances per image in Synthia (approximately 152), compared to only 27 in Cityscapes. For point-based loss computation, we follow~\cite{cheng2022masked} and sample $N_p = 112 \times 112$ points, with $\beta = 75\%$ selected from pixels with the highest sampling affinity.

For the domain generalization (source-only) experiments used as supervised baselines, we train for 20k iterations (batch size 4) when the source domain is synthetic, and for 40k iterations when it is real (Cityscapes, ACDC, MUSES). We vary the number of iterations because different datasets converge at different rates, and using dataset-appropriate training durations yields stronger and fairer supervised baselines. 

\section{Per-mask vs. All-mask CBPF} \label{ref:appendix_allmask_permask}
Fig.~\ref{fig:app:per_mask_vs_all} provides a visual comparison between the \textit{all-mask} confidence-based point filtering (CBPF) used in our method and the \textit{per-mask} CBPF variant, as discussed in Tab.~\ref{tab:table_ubpf_variant_ablation} of the main manuscript.
The left and middle columns illustrate the complementary roles of mask-wide loss scaling (MLS) and all-mask CBPF in suppressing unreliable gradients.
The transition from the middle to the right column highlights the key limitation of per-mask CBPF: it fails to identify false negative regions that lie far from the predicted mask boundary.

A closer inspection of the top-right example shows that per-mask CBPF incorrectly retains a large number of low-confidence pixels on the lower part of the left leg, which would lead the student to learn from erroneous pseudo-labels.
These qualitative observations align with the quantitative results reported in Tab.~\ref{tab:table_ubpf_variant_ablation} of the main manuscript, where all-mask CBPF provides a substantial performance advantage over its per-mask counterpart.

\newcommand{\mywfa}{0.315\linewidth}
\begin{figure}[h]
    \centering
    \begin{tabular}{c@{\,}c@{\,}c}
         \includegraphics[width=\mywfa]{figures_fig11_complementary_person_conf.jpg} & 
         \includegraphics[width=\mywfa]{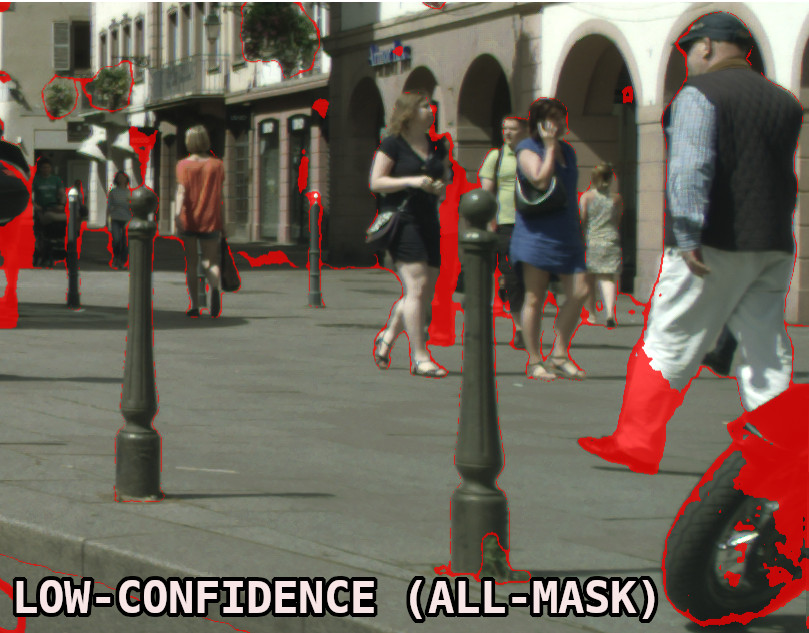} &
         \includegraphics[width=\mywfa]{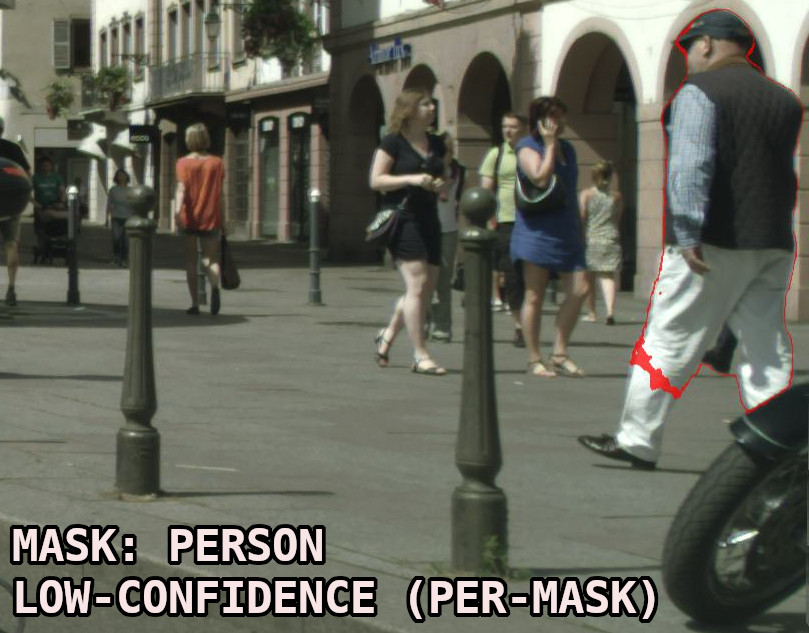} \\[0.5em]
         \includegraphics[width=\mywfa]{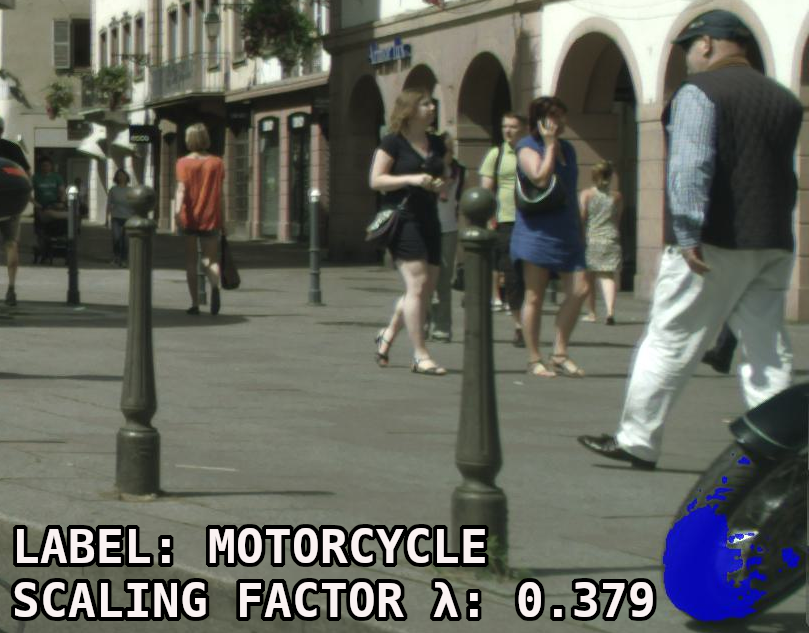} & 
         \includegraphics[width=\mywfa]{figures_figa3_uncertainty_better.jpg} &
         \includegraphics[width=\mywfa]{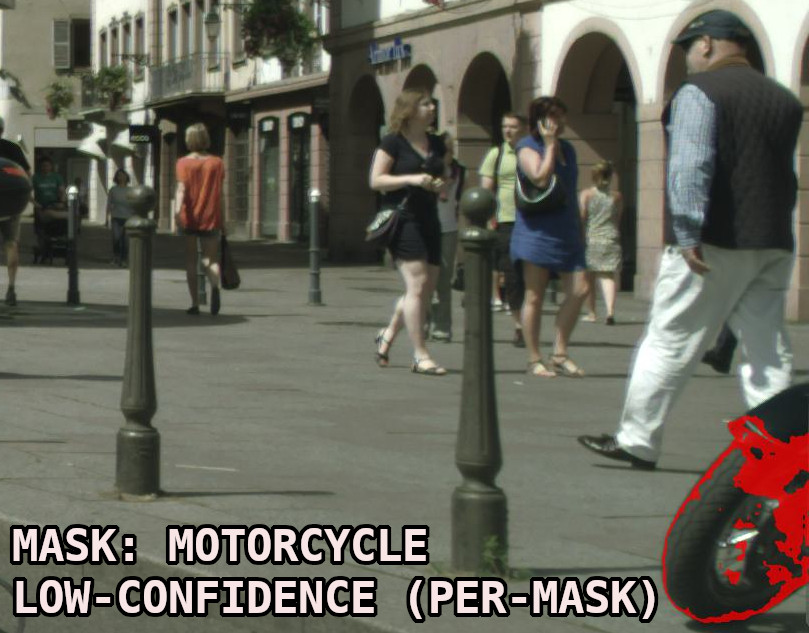} \\ 
    \end{tabular}
    \caption{
Visual comparison of \text{per-mask} and \text{all-mask} point filtering (\textit{cf.}~Table~\ref{tab:table_ubpf_variant_ablation} main manuscript).
The top row shows a \texttt{person} mask (left), uncertain pixels retained by all-mask filtering (middle), and those retained by per-mask filtering (right).
The bottom row presents the same intermediate outputs for a \texttt{motorcycle} mask.
As originally reported in the supplement of MC-PanDA~\citep{martinovic2024eccv}, all-mask filtering more reliably identifies false negative pixel assignments than per-mask filtering, a behavior that complements and explains the quantitative differences observed in Tab.~\ref{tab:table_ubpf_variant_ablation}.
}
    \label{fig:app:per_mask_vs_all}
\end{figure}

\color{changesframe}\section{Failure cases}\label{app:failure_cases}

\begin{figure}[h!]
    \centering
    \begin{tabular}{c@{\hskip 1pt}c}

    \begin{tikzpicture}
        \node[inner sep=0] (img) {\includegraphics[width=0.48\linewidth]{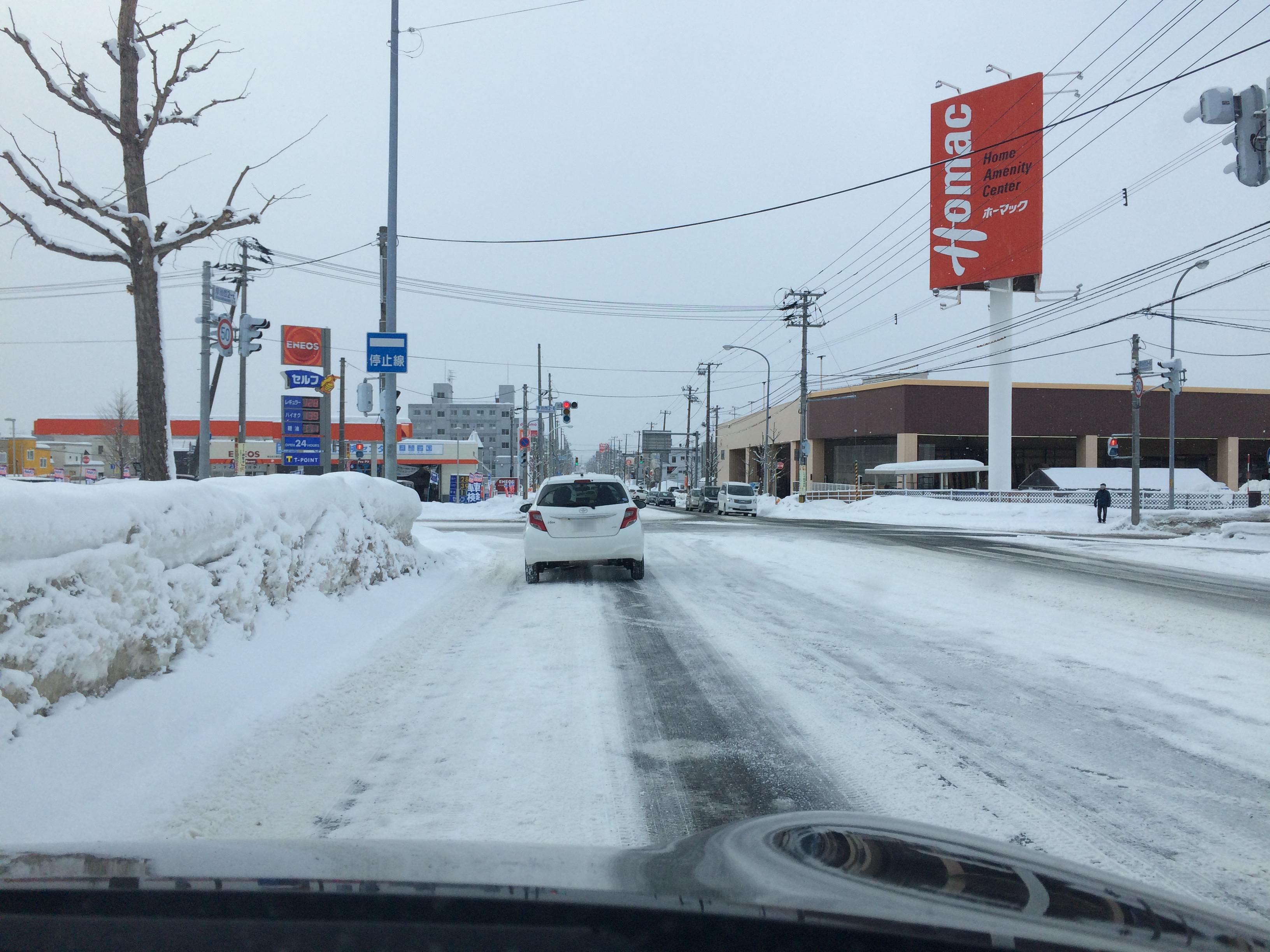}};
        % label
        \node[anchor=north west,
              fill=none,
              text=black,
              rounded corners=2pt,
              inner sep=3pt,
              font=\ttfamily\footnotesize] at (img.north west) {Vistas Img};
    \end{tikzpicture}
    &
    \begin{tikzpicture}
        \node[inner sep=0] (img) {\includegraphics[width=0.48\linewidth]{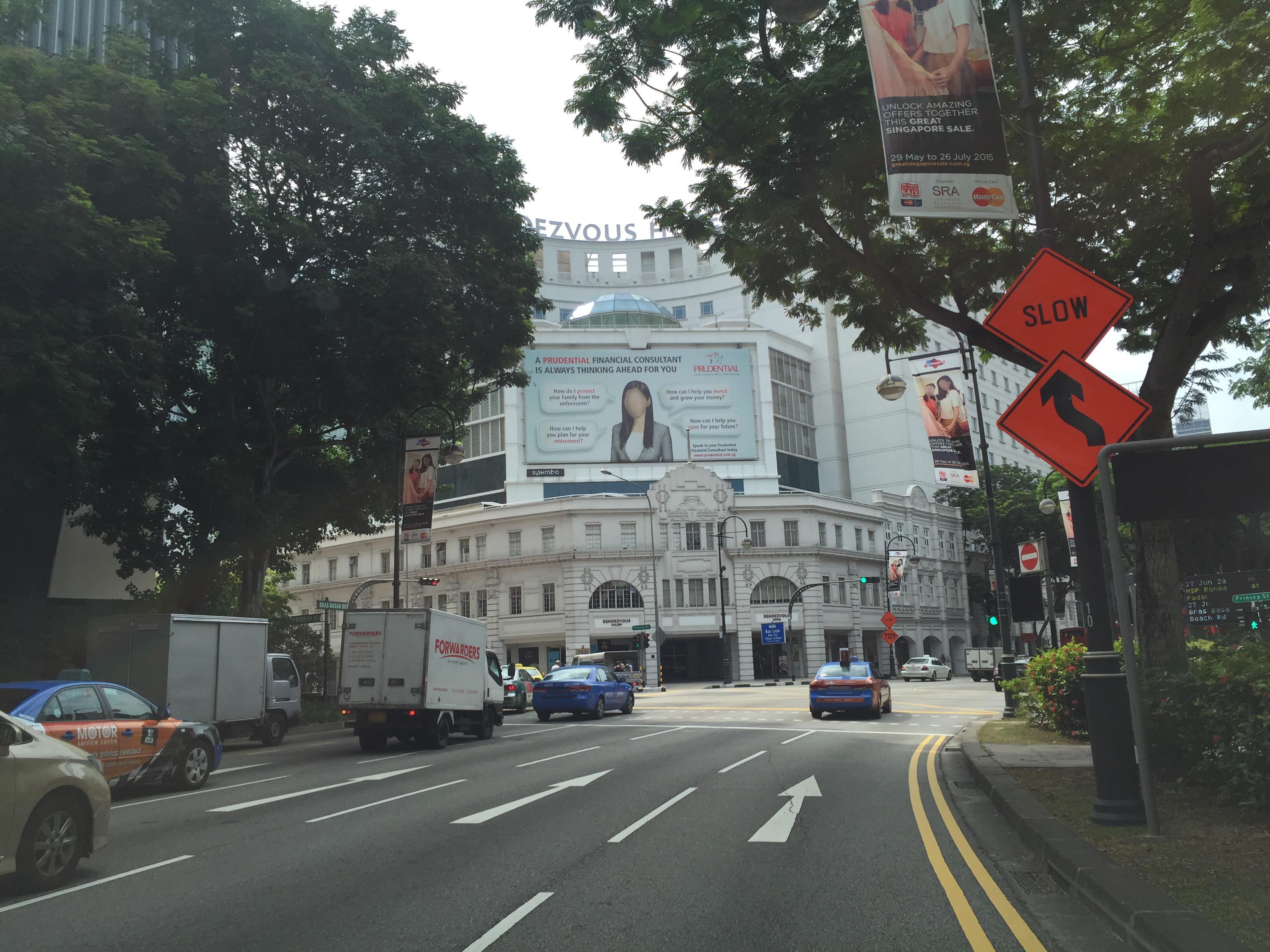}};
        \node[anchor=north west,
              fill=none,
              text=white,
              rounded corners=2pt,
              inner sep=3pt,
              font=\ttfamily\footnotesize] at (img.north west) {Vistas Img};
    \end{tikzpicture}
    \\[-1pt]

    \begin{tikzpicture}
        \node[inner sep=0] (img) {\includegraphics[width=0.48\linewidth]{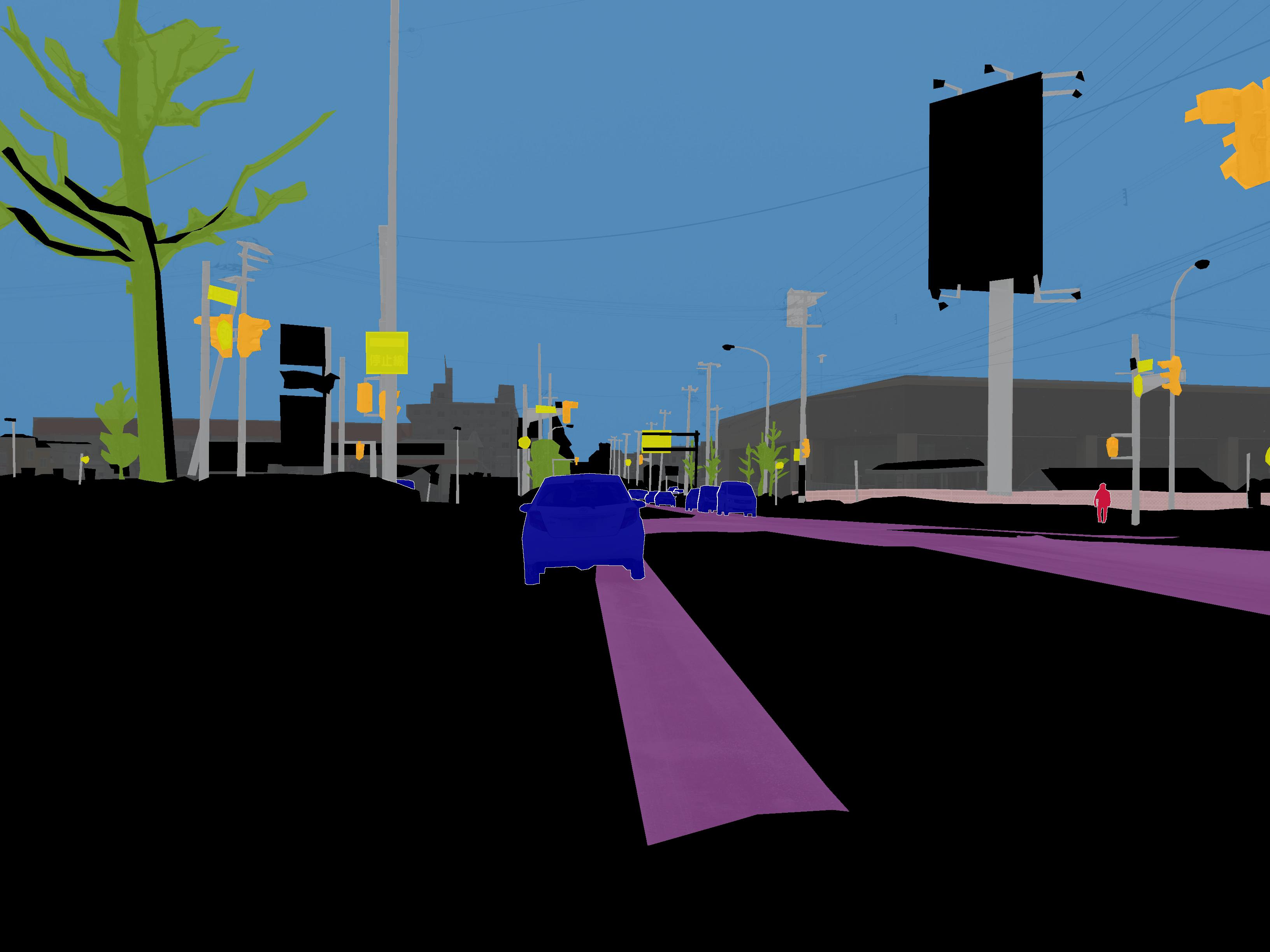}};
        \node[anchor=north west,
              fill=none,
              text=white,
              rounded corners=2pt,
              inner sep=3pt,
              font=\ttfamily\footnotesize] at (img.north west) {Ground-truth};
    \end{tikzpicture}
    &
    \begin{tikzpicture}
        \node[inner sep=0] (img) {\includegraphics[width=0.48\linewidth]{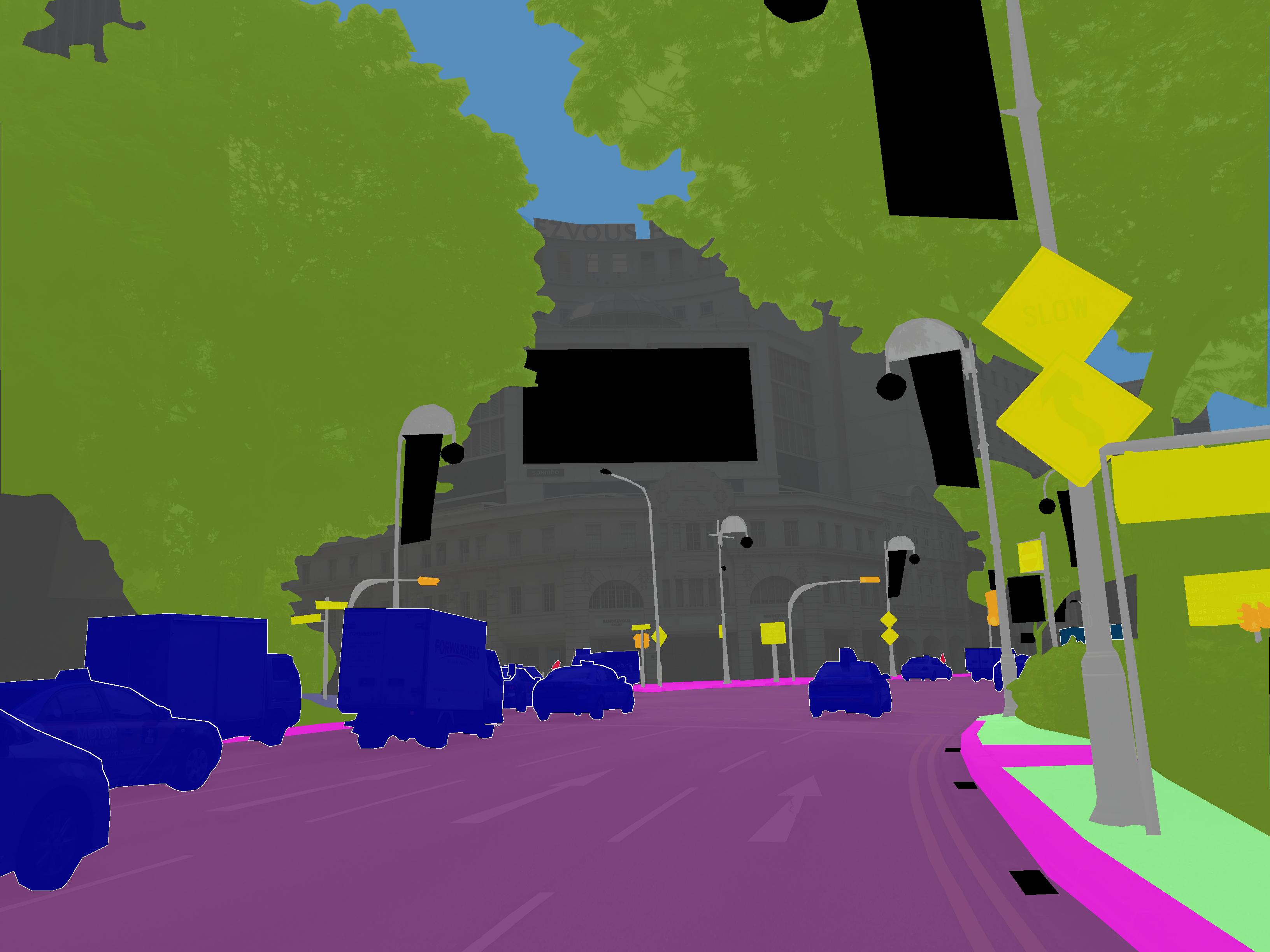}};
        \node[anchor=north west,
              fill=none,
              text=white,
              rounded corners=2pt,
              inner sep=3pt,
              font=\ttfamily\footnotesize] at (img.north west) {Ground-truth};
    \end{tikzpicture}

    \\[-1pt]

    \begin{tikzpicture}
        \node[inner sep=0] (img) {\includegraphics[width=0.48\linewidth]{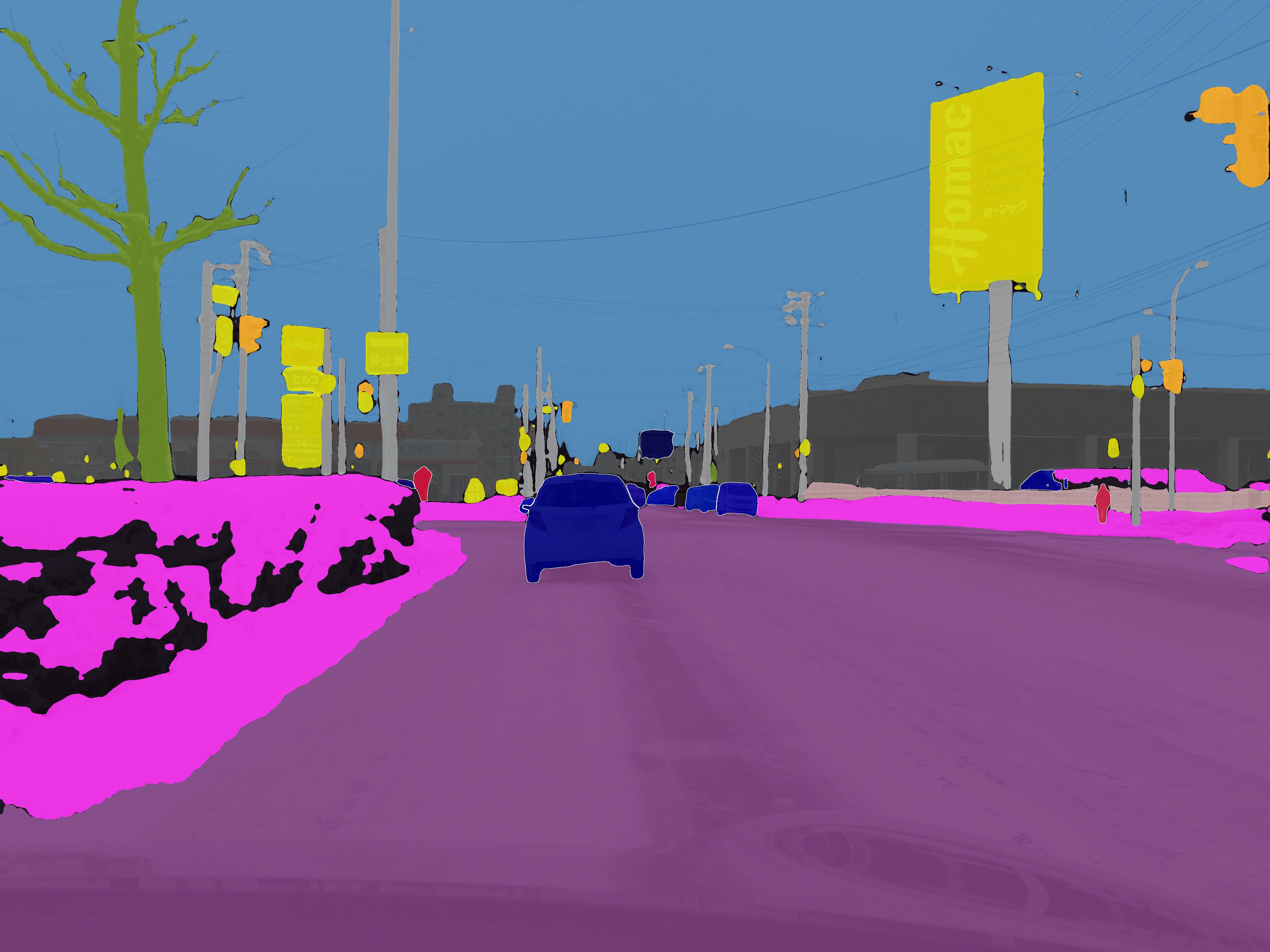}};
        \node[anchor=north west,
              fill=none,
              text=white,
              rounded corners=2pt,
              inner sep=3pt,
              font=\ttfamily\footnotesize] at (img.north west) {\ours{} (Usyn$\rightarrow$Vistas)};
    \end{tikzpicture}
    &
    \begin{tikzpicture}
        \node[inner sep=0] (img) {\includegraphics[width=0.48\linewidth]{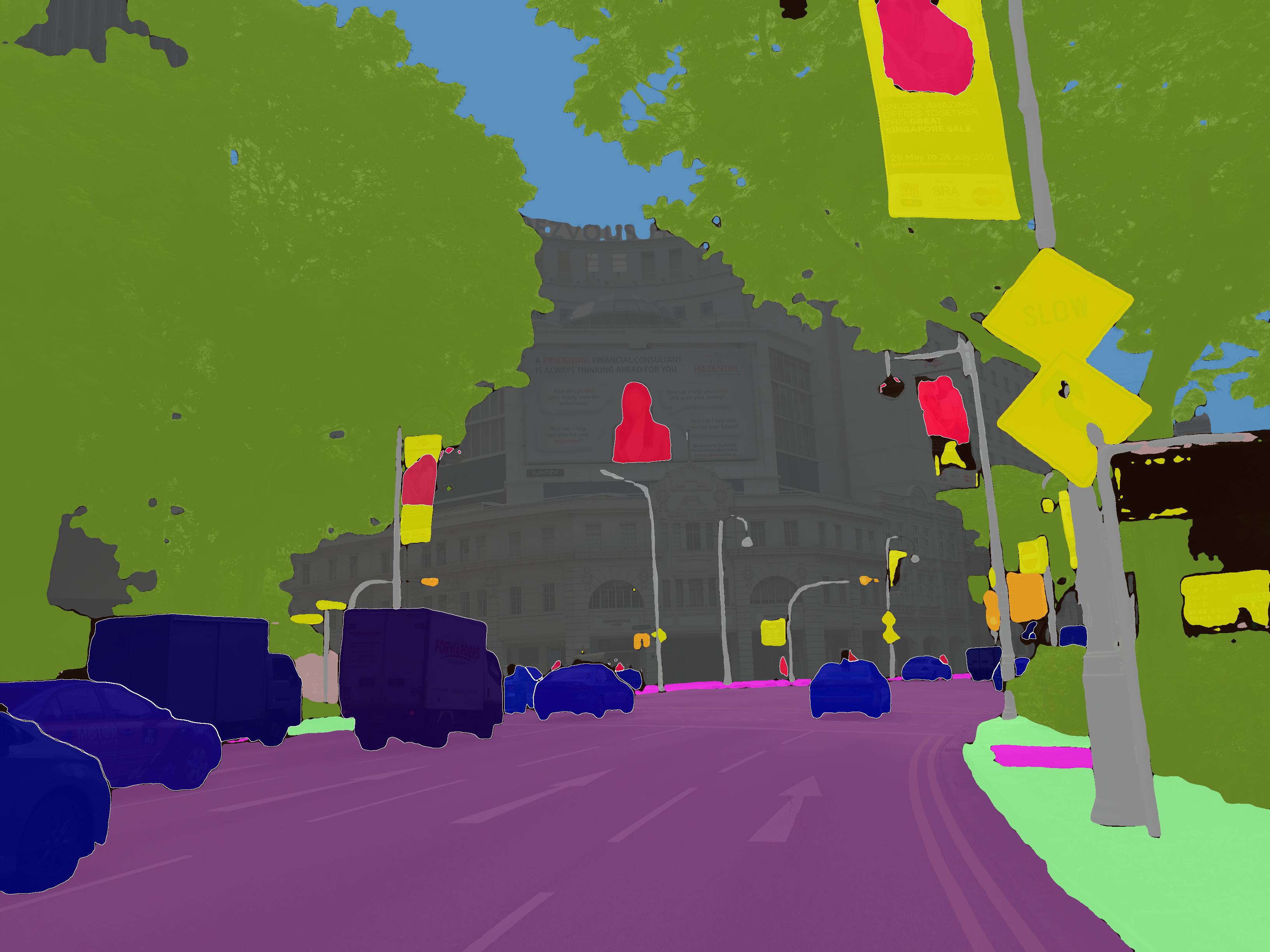}};
        \node[anchor=north west,
              fill=none,
              text=white,
              rounded corners=2pt,
              inner sep=3pt,
              font=\ttfamily\footnotesize] at (img.north west) {\ours{} (Usyn$\rightarrow$Vistas)};
    \end{tikzpicture}
    \end{tabular}
    \cityscapeslegendfailureapp
    \vspace{4pt}
\caption{
\textcolor{changesframe}{
Additional failure cases of \ours{}. Left: UrbanSyn$\rightarrow$Vistas example illustrating
source-domain coverage limitations, where target-domain \textit{snow} regions are mapped to
the closest available road-scene categories, such as
\texttt{\textcolor{csroad}{road}} and \texttt{\textcolor{cssidewalk}{sidewalk}}. Right:
UrbanSyn$\rightarrow$Vistas example illustrating an annotation-policy ambiguity, where
\textit{person} depictions on billboards are segmented as
\texttt{\textcolor{csperson}{person}} instances.
}
}

    \label{fig:failure_cases_app}
\end{figure}

\textcolor{changesframe}{
Fig.~\ref{fig:failure_cases_app} provides additional failure cases complementing
Fig.~\ref{fig:failure_cases} in the main manuscript. The left example further illustrates
source-domain coverage limitations in UrbanSyn$\rightarrow$Vistas: target-domain concepts such
as \textit{snow}, which are absent from the source supervision, are mapped to the closest
available road-scene classes, such as \texttt{\textcolor{csroad}{road}} and
\texttt{\textcolor{cssidewalk}{sidewalk}}. The right example shows an annotation-policy
ambiguity, where \textit{person} depictions on billboards are segmented as
\texttt{\textcolor{csperson}{person}} instances.
}

\color{changesframe}\section{MLS and Mask Quality}
\label{app:mls_iou_correlation}

\textcolor{changesframe}{To further analyze the behavior of MLS, we measure how the mask-wide weight $\lambda_i$
relates to the actual quality of teacher pseudo-masks. We
analyze teacher predictions from an intermediate checkpoint on all 1600 target-domain (i.e., ACDC) training
images. Specifically, for each predicted mask, we compute our mask weight $\lambda_i$ and compare it with the IoU between the prediction and its matched ground-truth mask, which we treat as the oracle mask quality. For stuff classes, the correspondence is unambiguous because each class forms
a single semantic region per image. For thing classes, we compute correlations only over
matched prediction--ground-truth instance pairs. }

\begin{table}[h!]
\centering
\begingroup
\color{changesframe}
\caption{
\textcolor{changesframe}{
Full per-class Spearman $\rho$ correlation between the MLS weight $\lambda_i$ and
pseudo-mask quality measured by IoU on Cityscapes$\rightarrow$ACDC. For thing classes,
correlations are computed over matched prediction--ground-truth instance pairs.
}
}
\label{tab:lambda_iou_correlation_per_class}
\footnotesize
\setlength{\tabcolsep}{5pt}
\vspace{0.5em}
\begin{tabular}{llcc}
\toprule
Group & Class & \# pairs & Spearman $\rho$ \\
\midrule
Stuff & \texttt{road}          & 1599 & 0.52 \\
      & \texttt{sidewalk}      & 1400 & 0.65 \\
      & \texttt{building}      & 1420 & 0.87 \\
      & \texttt{wall}          & 848  & 0.66 \\
      & \texttt{fence}         & 700  & 0.59 \\
      & \texttt{pole}          & 1587 & 0.65 \\
      & \texttt{tr.~light}     & 784  & 0.63 \\
      & \texttt{tr.~sign}      & 1463 & 0.64 \\
      & \texttt{vegetation}    & 1572 & 0.74 \\
      & \texttt{terrain}       & 715  & 0.62 \\
      & \texttt{sky}           & 1596 & 0.83 \\
\cmidrule(lr){2-4}
      & Avg. stuff             & --   & \textbf{0.67} \\
\midrule
Things & \texttt{person}       & 898  & 0.45 \\
       & \texttt{rider}        & 73   & 0.43 \\
       & \texttt{car}          & 4912 & 0.70 \\
       & \texttt{truck}        & 210  & 0.52 \\
       & \texttt{bus}          & 62   & 0.79 \\
       & \texttt{train}        & 88   & 0.69 \\
       & \texttt{motorcycle}   & 50   & 0.21 \\
       & \texttt{bicycle}      & 113  & 0.42 \\
\cmidrule(lr){2-4}
       & Avg. things           & --   & \textbf{0.52} \\
\midrule
All    & Avg. all              & --   & \textbf{0.61} \\
\bottomrule
\end{tabular}
\endgroup
\end{table}

\textcolor{changesframe}{Table~\ref{tab:lambda_iou_correlation_per_class}
reports the per-class Spearman correlation. The correlation is positive for both stuff and thing
classes, with class-averaged correlations of 0.67 and 0.52, respectively, and 0.61 over all
19 classes. This supports the intended role of $\lambda_i$ as a mask-level quality signal for
weighting pseudo-label supervision.}

\color{changesframe}\section{EDAPS/LIDAPS with DINOv2}
\label{app:dinov2_edaps_lidaps}

\begingroup
\color{changesframe}
To better separate the effect of the adaptation method from the effect of backbone
initialization, we additionally reproduce EDAPS~\citep{Saha_2023_ICCV} and
LIDAPS~\citep{mansour2025wacv} with DINOv2-B~\citep{oquab2024dinov}. We use the
official codebases of both methods and replace the original MiT-B5~\citep{NEURIPS2021_segformer}
encoder with DINOv2-B through a ViTDet-style feature pyramid~\citep{li2022exploring},
following the feature extraction strategy used in our implementation. As a sanity check, we
first verified that our environment reproduces the original EDAPS and LIDAPS results with the
MiT-B5 backbone.

We found that directly replacing MiT-B5 with DINOv2-B leads to unstable training in these
dual-branch panoptic architectures. In particular, the Mask R-CNN-style~\citep{he2017mask} instance branch often
collapsed, yielding zero PQ for \textit{things} classes, while the semantic branch for
\textit{stuff} classes remained partially functional. This observation is also consistent with
public implementation reports discussing poor detection or instance-segmentation performance
when using DINOv2 weights in Faster R-CNN~\citep{ren2016faster} / Mask R-CNN~\citep{he2017mask} or ViTDet-style pipelines.\footnote{
\url{https://github.com/facebookresearch/dinov2/issues/350},
\url{https://github.com/facebookresearch/dinov2/issues/65}
} While these reports are not controlled benchmarks, they suggest that transferring DINOv2 to
two-stage instance-detection architectures may require additional stabilization. We therefore
explored several stabilization strategies, including changes to the learning rate, backbone
learning-rate multiplier, weight decay, warmup duration, pseudo-label confidence threshold,
and normalization layers. Among these, adding GroupNorm to the instance-branch feature pyramid
was crucial for stable training. We also found that increasing the crop size from
$512{\times}512$ to $512{\times}1024$ further improved performance for EDAPS on both
benchmarks and for LIDAPS on Synthia$\rightarrow$Cityscapes.

Table~\ref{tab:app_synthetic_to_real_sota_with_dino} reports the resulting comparison. The
DINOv2-B backbone improves some EDAPS/LIDAPS results, confirming that stronger self-supervised encoders can also benefit competing methods. Nevertheless, \ours{} remains
ahead on both benchmarks. These results support the conclusion that strong initialization is
an important enabling component, but the proposed confidence-guided Mask2Former-based
adaptation strategy remains essential for robust UDA panoptic segmentation.
\endgroup

\begin{table*}[h!]
    \centering
    \begingroup
    \color{changesframe}
    \caption{
    \textcolor{changesframe}{Extended performance (PQ$_{16}$) comparison with EDAPS and LIDAPS on Synthia$\rightarrow$Cityscapes and
    Synthia$\rightarrow$Vistas. $^\ddagger$ denotes our DINOv2-B
    reimplementations using the official codebases, a ViTDet-style feature pyramid, and GroupNorm
    normalization. For \ours{}, we report the mean over three random seeds.}
    }
    \label{tab:app_synthetic_to_real_sota_with_dino}
    \footnotesize
    \setlength{\tabcolsep}{6pt}
    \vspace{0.5em}
    \begin{tabular}{llcc}
        \toprule
        Method & Encoder / setting
        & Synthia$\rightarrow$City
        & Synthia$\rightarrow$Vistas \\
        \midrule
        EDAPS~\citep{Saha_2023_ICCV}
            & MiT-B5
            & 41.2 & 36.6 \\
        LIDAPS~\citep{mansour2025wacv}
            & MiT-B5
            & 44.8 & 38.0 \\
        \midrule
        EDAPS$^\ddagger$~\citep{Saha_2023_ICCV}
            & DINOv2-B, $512{\times}512$ crop
            & 36.1 & 38.9 \\
        EDAPS$^\ddagger$~\citep{Saha_2023_ICCV}
            & DINOv2-B, $512{\times}1024$ crop
            & 42.0 & 40.5 \\
        LIDAPS$^\ddagger$~\citep{mansour2025wacv}
            & DINOv2-B, $512{\times}512$ crop
            & 40.2 & 37.0 \\
        LIDAPS$^\ddagger$~\citep{mansour2025wacv}
            & DINOv2-B, $512{\times}1024$ crop
            & 44.7 & 35.8 \\
        \midrule
        \rowcolor{gray!10}
        \ours{}
            & DINOv2-B, $512{\times}1024$ crop
            & \textbf{49.6} & \textbf{44.8} \\
        \bottomrule
    \end{tabular}
    \endgroup
\end{table*}

\color{changesframe}
\section{Compute, memory, and inference-time comparison}
\label{app:compute_memory_comparison}

\textcolor{changesframe}{
We compare the computational cost, memory usage, and inference speed of \ours{} with LIDAPS,
the strongest EDAPS-family baseline in Table~\ref{tab:compute_memory_comparison}. We report
results on Synthia$\rightarrow$Cityscapes and measure all runtimes on the same hardware setup,
using a single NVIDIA RTX 6000 Ada GPU. For LIDAPS, the training schedule consists of the
standard 40k main training stage followed by the 10k instance-mixing stage. For inference, we
evaluate 500 Cityscapes images at $1024{\times}2048$ resolution and report the mean runtime
with data loading excluded.
}

\textcolor{changesframe}{
The comparison highlights the practical trade-off. LIDAPS uses a shorter training schedule and
therefore has lower total GPU-hours than the final 110k \ours{} model, while \ours{} is faster
per training iteration. At a near matched-compute budget, the 85k checkpoint of \ours{} already
outperforms LIDAPS by +3.8 PQ. At inference, \ours{} is about $1.4$--$1.6\times$ faster,
requires substantially less memory, and achieves the best final performance. We also
experimented with extending LIDAPS to 110k iterations, but
performance even slightly degraded compared with the standard 50k schedule.
}

\begin{table*}[t]
\centering
\begingroup
\color{changesframe}
\caption{
Compute, memory, and inference-time comparison with LIDAPS on
Synthia$\rightarrow$Cityscapes. For LIDAPS, training includes the 40k main stage and the 10k
instance-mixing stage. Inference is measured on 500 Cityscapes images
($1024{\times}2048$), reporting the mean runtime with data loading excluded.
$^\ddagger$ denotes our DINOv2-B reimplementation of LIDAPS, described in
Appendix~\ref{app:dinov2_edaps_lidaps}.
}
\label{tab:compute_memory_comparison}
\footnotesize
\setlength{\tabcolsep}{4pt}
\vspace{0.5em}
\begin{tabular}{llcccccccc}
\toprule
& & \multicolumn{5}{c}{Training} & \multicolumn{2}{c}{Inference ($1024{\times}2048$)} & \\
\cmidrule(lr){3-7} \cmidrule(lr){8-9}
Method & Backbone
& Iters. & Crop & s/iter$\downarrow$ & GPU-h$\downarrow$ & Peak mem.$\downarrow$
& FPS$\uparrow$ & Peak mem.$\downarrow$
& \textbf{PQ}$\uparrow$ \\
\midrule
LIDAPS & MiT-B5
& $40k{+}10k$ & $512{\times}512$
& 1.77 & \textbf{24.5} & \textbf{17.5 GiB}
& 2.1 & 6.6 GiB & 44.8 \\
LIDAPS$^\ddagger$ & DINOv2-B
& $40k{+}10k$ & $512{\times}1024$
& 1.83 & 25.4 & 29.2 GiB
& 2.4 & 8.2 GiB & 44.7 \\
\midrule
\rowcolor{gray!5}
\ours{} (85k ckpt.) & DINOv2-B
& $85k$ & $512{\times}1024$
& \textbf{1.04} & 24.6 & 24.7 GiB
& \textbf{3.4} & \textbf{4.0 GiB} & 48.6 \\
\rowcolor{gray!10}
\ours{} (final) & DINOv2-B
& $110k$ & $512{\times}1024$
& \textbf{1.04} & 31.9 & 24.7 GiB
& \textbf{3.4} & \textbf{4.0 GiB} & \textbf{49.6} \\
\bottomrule
\end{tabular}
\endgroup
\end{table*}

\color{changesframe}
\section{Initial $\tau_1$ -- robustness}
\label{app:tau1_robustness}

\textcolor{changesframe}{
We additionally evaluate the robustness of adaptive class-dependent MLS to the initial value of
$\tau_1$ on UrbanSyn$\rightarrow$Vistas. Table~\ref{tab:app_usyn_vistas_tau1_robustness}
shows that performance remains stable across a wide range of initial thresholds. The average
performance across all tested initial values is \mstd{49.5}{0.2} PQ, confirming the robustness
trend observed on Synthia$\rightarrow$Vistas (see Table~\ref{tab:constant_vs_adaptive_synthia_vistas}), Cityscapes$\rightarrow$ACDC (see Table~\ref{tab:constant_vs_adaptive_synthia_vistas}), and
Cityscapes$\rightarrow$MUSES (see Table~\ref{tab:tau_2_robustness}).
}

\begin{table}[h!]
\centering
\begingroup
\color{changesframe}
\caption{
Robustness of adaptive class-dependent MLS to initial values of $\tau_1$ on
UrbanSyn$\rightarrow$Vistas. Each value is averaged over three random seeds. The last column
reports the mean and standard deviation across initial $\tau_1$ values.
}
\label{tab:app_usyn_vistas_tau1_robustness}
\footnotesize
\setlength{\tabcolsep}{2.5pt}
\vspace{0.5em}
\begin{tabular}{lccccccc@{\hskip 10pt}c}
\toprule
\multicolumn{1}{r}{initial $\tau_1{=}$} & 0.0 & 0.5 & 0.6 & 0.7 & 0.8 & 0.9 & 0.99 & Avg. \\
\midrule
\multirow{2}{*}{\makecell[l]{UrbanSyn \\ $\rightarrow$Vistas}}
& \multirow{2}{*}{49.5} &
\multirow{2}{*}{49.2} &
\multirow{2}{*}{49.7} &
\multirow{2}{*}{49.7} &
\multirow{2}{*}{49.3} &
\multirow{2}{*}{49.4} &
\multirow{2}{*}{49.5} &
\multirow{2}{*}{\mstd{49.5}{0.2}} \\
&&&&&&& \\
\bottomrule
\end{tabular}
\endgroup
\end{table}

\color{changesframe}
\section{Impact of adaptive thresholding on rare classes}
\label{app:rare_frequent_tau1}

\textcolor{changesframe}{
To analyze the behavior of the adaptive threshold update under class imbalance, we compare the
fixed threshold $\tau_1=0.99$ with our per-class adaptive threshold on
Cityscapes$\rightarrow$ACDC. We focus on the five most frequent and five rarest classes, where
class frequency is measured as the percentage of source-domain training images in which the
class appears at least once. Table~\ref{tab:app_rare_frequent_tau1} reports per-class PQ
averaged over three random seeds. Adaptive thresholding improves the average performance of
the five most frequent classes by 1.6 PQ and the five rarest classes by 3.1 PQ. Notably, all
five rare classes improve, suggesting that the adaptive update alleviates the difficulty of
using a single global threshold for classes with substantially different frequency and
confidence statistics.
}

\begin{table*}[h!]
\centering
\begingroup
\color{changesframe}
\caption{
Effect of per-class adaptive thresholding on the five most frequent and five rarest classes for
Cityscapes $\rightarrow$ ACDC. We compare the fixed threshold $\tau_1=0.99$ with our
per-class adaptive threshold. Class frequency, shown in parentheses, is measured as the
percentage of source-domain training images in which the class appears at least once.
Results are per-class PQ averaged over three random seeds.
}
\label{tab:app_rare_frequent_tau1}
\footnotesize
\setlength{\tabcolsep}{3pt}
\vspace{0.5em}
\resizebox{\textwidth}{!}{%
\begin{tabular}{lcccccccccccc}
\toprule
\multicolumn{13}{c}{\textbf{Five most frequent classes}} \\
\cmidrule(lr){2-11}
&
\multicolumn{2}{c}{\texttt{pole} {\scriptsize (99.1\%)}} &
\multicolumn{2}{c}{\texttt{road} {\scriptsize (98.6\%)}} &
\multicolumn{2}{c}{\texttt{building} {\scriptsize (98.6\%)}} &
\multicolumn{2}{c}{\texttt{vegetation} {\scriptsize (97.2\%)}} &
\multicolumn{2}{c}{\texttt{car} {\scriptsize (95.2\%)}} &
Avg. & \\
\cmidrule(lr){2-3}
\cmidrule(lr){4-5}
\cmidrule(lr){6-7}
\cmidrule(lr){8-9}
\cmidrule(lr){10-11}
fixed $\tau_1=0.99$
& 42.2 & \blarrow{}
& 94.9 & \blarrow{}
& 79.6 & \blarrow{}
& 74.5 & \blarrow{}
& 64.1 & \blarrow{}
& 71.0 & \blarrow{} \\
adaptive $\tau_1$
& 44.8 & \textcolor{Improved}{\textbf{+2.7}}
& 95.2 & \textcolor{Improved}{\textbf{+0.3}}
& 79.8 & \textcolor{Improved}{\textbf{+0.1}}
& 72.3 & \textcolor{Degraded}{\textbf{-2.3}}
& 71.4 & \textcolor{Improved}{\textbf{+7.3}}
& 72.7 & \textcolor{Improved}{\textbf{+1.6}} \\
\midrule
\multicolumn{13}{c}{\textbf{Five rarest classes}} \\
\cmidrule(lr){2-11}
&
\multicolumn{2}{c}{\texttt{train} {\scriptsize (4.8\%)}} &
\multicolumn{2}{c}{\texttt{bus} {\scriptsize (9.2\%)}} &
\multicolumn{2}{c}{\texttt{truck} {\scriptsize (12.1\%)}} &
\multicolumn{2}{c}{\texttt{motorcycle} {\scriptsize (17.2\%)}} &
\multicolumn{2}{c}{\texttt{wall} {\scriptsize (32.6\%)}} &
Avg. & \\
\cmidrule(lr){2-3}
\cmidrule(lr){4-5}
\cmidrule(lr){6-7}
\cmidrule(lr){8-9}
\cmidrule(lr){10-11}
fixed $\tau_1=0.99$
& 60.0 & \blarrow{}
& 62.0 & \blarrow{}
& 40.8 & \blarrow{}
& 31.0 & \blarrow{}
& 48.9 & \blarrow{}
& 48.5 & \blarrow{} \\
adaptive $\tau_1$
& 62.5 & \textcolor{Improved}{\textbf{+2.5}}
& 62.8 & \textcolor{Improved}{\textbf{+0.8}}
& 44.1 & \textcolor{Improved}{\textbf{+3.4}}
& 35.5 & \textcolor{Improved}{\textbf{+4.5}}
& 53.0 & \textcolor{Improved}{\textbf{+4.1}}
& 51.6 & \textcolor{Improved}{\textbf{+3.1}} \\
\bottomrule
\end{tabular}%
}
\endgroup
\end{table*}

\color{black}\section{Per-class results}\label{appendix:per_class_results}
%%%%%%%%%%%%%%%%%%%% TABLE %%%%%%%%%%%%%%%%%%%
% Per-class results all datasets
%%%%%%%%%%%%%%%%%%%% TABLE %%%%%%%%%%%%%%%%%%%
As a complement to Tables~\ref{tab:synthetic_to_real_mean} and~\ref{tab:city_to_other_mean} in the main manuscript, Table~\ref{tab:per_class_results} reports per-class panoptic performance of \ours{} and compares it with state-of-the-art methods across four standard domain adaptation benchmarks.
\begin{table*}[h!]
    \caption{Per-class PQ performance evaluation 
    on four standard domain adaptation setups and comparison with the state of the art.
    All experiments for MC-PanDA and \ours{} are averaged over three random seeds.}
    \label{tab:per_class_results}
    \centering
    \footnotesize
    \setlength{\tabcolsep}{3.0pt}
    \renewcommand{\arraystretch}{1}
    \newcommand{\wall}{\includegraphics[width=1em]{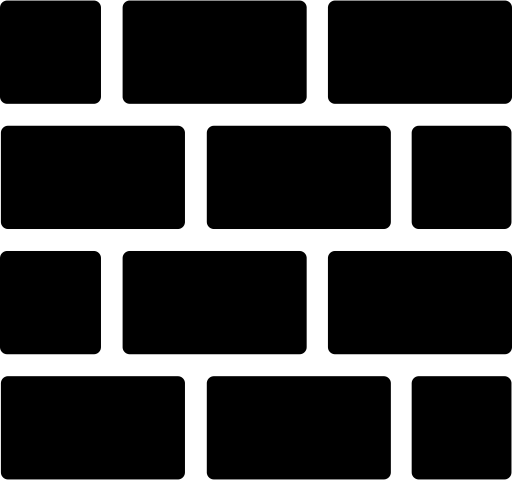}}
    \newcommand{\fence}{\includegraphics[width=1em]{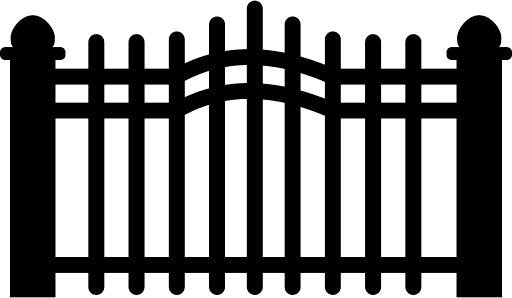}}
    \newcommand{\pole}{\includegraphics[width=0.6em]{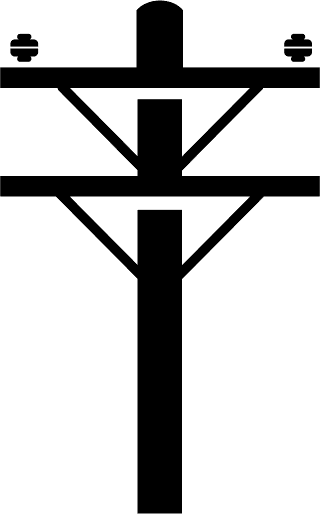}}
    \newcommand{\trsign}{\includegraphics[width=1em]{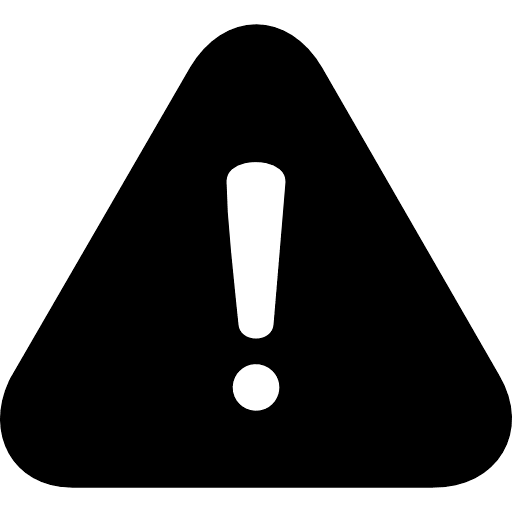}}
    \newcommand{\mybf}[1]{\text{#1}}
    \newcommand{\mybff}[1]{\textbf{#1}}
    {
    \begin{tabular}{lcccccccccccccccc@{\hskip 0.35cm}r}
        \toprule 
& \rotatebox{90}{\tiny\texttt{road}}
& \rotatebox{90}{\tiny\texttt{sidewalk}}
& \rotatebox{90}{\tiny\texttt{building}}
& \rotatebox{90}{\tiny\texttt{wall}}
& \rotatebox{90}{\tiny\texttt{fence}}
& \rotatebox{90}{\tiny\texttt{pole}}
& \rotatebox{90}{\tiny\texttt{tr.~light}}
& \rotatebox{90}{\tiny\texttt{tr.~sign}}
& \rotatebox{90}{\tiny\texttt{vegetation}}
& \rotatebox{90}{\tiny\texttt{sky}}
& \rotatebox{90}{\tiny\texttt{person}}
& \rotatebox{90}{\tiny\texttt{rider}}
& \rotatebox{90}{\tiny\texttt{car}}
& \rotatebox{90}{\tiny\texttt{bus}}
& \rotatebox{90}{\tiny\texttt{motorcycle}}
& \rotatebox{90}{\tiny\texttt{bicycle}}
& {$\text{PQ}_{16}$} \\
        \midrule
        Method & \multicolumn{17}{c}{Synthia$\rightarrow$Cityscapes}\\
        \midrule
        %\citep{huang2021cross}
        CVRN              & 86.6  & 33.8      & 74.6      & 3.4  & 0.0 & 10.0 & 5.7   & 13.5 & 80.3 & 76.3 & 26.0 & 18.0 & 34.1 & 37.4 & 7.3  & 6.2 & {32.1} \\
        UniDAF     & 73.7      & 26.5      & 71.9      & 1.0 & 0.0 & 7.6 & 9.9 & 12.4 & 81.4 & 77.4 & 27.4 & 23.1 & {47.0} & {40.9} & 12.6 & 15.4 & {33.0}\\
        UniDAF-PSN & \mybf{87.7} & 34.0 & 73.2 & 1.3 & 0.0 & 8.1 & 9.9 & 6.7 & 78.2 & 74.0 & 37.6 & 25.3 & 40.7 & 37.4 & 15.0 & {18.8} & {34.2}\\
        EDAPS              & 77.5      & {36.9}  & {80.1} & \mybf{17.2} & \mybf{1.8} & {29.2} & \mybf{33.5} & {40.9} & {82.6} & {80.4} & {43.5} & {33.8} & 45.6 & 35.6 & {18.0} & 2.8 & {41.2}\\
        [0.2em]
        LIDAPS & 80.8& 48.8& 80.8& \text{17.6}& \text{2.5}& 29.9& \textbf{34.6}& \textbf{42.9}& 82.8& 82.9& 44.4& \textbf{40.5}& 51.7& 39.2& 27.4&10.7 & 44.8\\
        MC-PanDA & {87.2} & \mybf{51.8}  & \mybf{82.5} & 16.1 & {1.7} & \mybf{36.3} & {26.1} & \mybf{54.3} & \mybff{86.3} & \mybff{86.4} & \mybff{48.3} & \mybf{37.7} & \mybf{46.9} & \mybf{45.8} & \mybf{27.4} & \mybf{23.9} & \mybf{47.4}\\
        \rowcolor{gray!10}\ours{} & \textbf{89.9} & \textbf{63.6}  & \textbf{84.8} & \textbf{24.7} & \textbf{3.6} & \textbf{43.2} & \text{34.5} & \text{42.8} & \text{82.5} & \text{81.0} & \text{46.8} & \text{36.8} & \textbf{54.2} & \textbf{51.1} & \textbf{27.9} & \textbf{25.4} & \textbf{49.6} \\
        
        \toprule
         & \multicolumn{17}{c}{Synthia$\rightarrow$Vistas}\\
        \midrule 
        CVRN       & 33.4 & 7.4 & 32.9 & 1.6 & 0.0 & 4.3 & 0.4 & 6.5 & 50.8 & 76.8 & 30.6 & 15.2 & 44.8 & 18.8 & 7.9 & {9.5} & 21.3 \\
        EDAPS       & {77.5} & {25.3} & {59.9} & \mybf{14.9} & 0.0 & {27.5} & \text{33.1} & {37.1} & \mybf{72.6} & \mybf{92.2} & {32.9} & \mybf{16.4} & {47.5} & \mybf{31.4} & {13.9} & 3.7 & 36.6 \\[0.2em]
                LIDAPS & 76.5 &25.2 &64.2 &14.0 &0.2 &29.1 &35.6 &35.3 &\textbf{72.1} &\textbf{94.4} &33.8 &18.3 &50.3 &33.9 &19.3 &5.9 &38.0 \\
        MC-PanDA & \mybf{82.7} & \mybf{26.5} &\mybf{61.0} &\text{5.5} &\text{0.0} &\mybff{39.9} &\mybf{34.4} &\mybff{51.3} &\text{62.1} &\text{85.6} &\mybff{41.9} &\text{11.4} &\mybf{50.6} &\text{25.1} &\mybf{23.0} &\mybf{18.1} & \mybf{38.7} \\
        \rowcolor{gray!10}\ours{} & \textbf{87.6} & \textbf{55.2}  & \textbf{65.3} & \textbf{20.9} & \textbf{1.2} & \textbf{39.9} & \textbf{42.1} & \text{46.6} & \text{68.4} & \text{91.7} & \text{40.8} & \textbf{19.1} & \textbf{54.7} & \textbf{35.1} & \textbf{25.2} & \textbf{23.2} & \textbf{44.8} \\
        
        \toprule
         & \multicolumn{17}{c}{Cityscapes$\rightarrow$Foggy Cityscapes}\\
        \midrule
        CVRN           
        &93.6&52.3&65.3&7.5&15.9&5.2&7.4&22.3&57.8&48.7&32.9&30.9&49.6&38.9&18.0&25.2 & 35.7 \\
        UniDAF  
        &93.9&53.1&63.9&8.7&14.0&3.8&10.0&26.0&53.5&49.6&38.0&35.4&57.5&44.2&28.9&29.8& 37.6  \\
        EDAPS  
        & {91.0} & {68.5} & {80.9} & {24.1} & 29.0 & {50.1} & {47.2} & {67.0} & {85.3} & {71.8} & {50.9} & \mybf{51.2} & {64.7} & {47.7} & {36.9} & 41.5 & 56.7 \\[0.2em]
        LIDAPS & 92.3 &70.0 &83.2 &23.8 &31.9 &56.4 &47.7 &68.8& 86.6 &72.5 &53.2 &\mybff{53.6} &\mybff{68.0} &56.6 &42.8& \mybff{45.9} & 59.6 \\
        MC-PanDA & \mybff{98.0} & \mybf{80.6} & \mybf{85.8} & \mybf{45.7} & \mybf{43.4} & \mybf{60.3} & \mybf{49.4} & \mybff{73.8} & \mybf{87.9} & \mybf{81.7} & \mybff{53.5} & 47.8 & \mybf{65.2} & \mybf{61.6} & \mybf{40.2} & \mybf{44.3} & \mybf{63.7}\\ 
        \rowcolor{gray!10}\ours{} & \text{97.9} & \textbf{82.2}  & \textbf{87.9} & \textbf{51.7} & \textbf{53.4} & \textbf{63.4} & \textbf{51.3} & \text{72.5} & \textbf{89.3} & \textbf{83.0} & \text{53.4} & \text{49.3} & \text{66.3} & \textbf{72.1} & \textbf{45.6} & \text{45.7} & \textbf{66.6} \\
        \toprule
         & \multicolumn{17}{c}{Cityscapes$\rightarrow$Vistas}\\
        \midrule
        CVRN & 77.3& 21.0& 47.8& 10.5& 13.4& 7.5& 14.1& 25.1& 62.1& 86.4& 37.7& 20.4& 55.0& 21.7& 14.3& 21.4 & 33.5 \\
        EDAPS  & {58.8} & {43.4} & {57.1} & {25.6} & 29.1 & {34.3} & {35.5} & {41.2} & {77.8} & {59.1} & {35.0} & {23.8} & {56.7} & {36.0} & {24.3} & 25.5 & 41.2  \\[0.2em]
        LIDAPS & 49.1 &44.3 &70.1 &26.5 &29.9 &37.4 &37.2 &43.2 &80.0 &46.1 &35.9 &25.0 &57.1 &41.6 &29.6 &28.4 & 42.6 \\
        MC-PanDA & \mybf{88.4} & \mybf{49.1} & \mybf{75.2} & \mybf{35.2} & \mybf{39.7} & \mybff{50.3} & \mybf{45.2} & \mybff{54.0} & \mybf{81.1} & \mybff{96.2} & \mybf{46.1} & \mybf{30.1} & \mybf{57.2} & \mybf{42.0} & \mybf{33.9} & \mybff{37.3} & \mybf{53.8} \\ 
        \rowcolor{gray!10}\ours{} & \textbf{92.9} & \textbf{58.0}  & \textbf{75.9} & \textbf{41.6} & \textbf{45.9} & \text{48.4} & \textbf{48.6} & \text{49.7} & \textbf{81.2} & \text{96.0} & \textbf{46.6} & \textbf{31.1} & \textbf{63.0} & \textbf{44.4} & \textbf{36.0} & \text{34.9} & \textbf{55.9} \\
        \bottomrule
    \end{tabular}
    }
\end{table*}
\end{appendices}
\end{document}